\documentclass{article}

\PassOptionsToPackage{numbers}{natbib}

\usepackage[preprint]{neurips_2024}

\usepackage[utf8]{inputenc} 
\usepackage[T1]{fontenc}    
\usepackage{hyperref}       
\usepackage{url}            
\usepackage{booktabs}       
\usepackage{amsfonts}       
\usepackage{nicefrac}       
\usepackage{microtype}      
\usepackage{xcolor}         
\usepackage{graphicx}
\usepackage{subcaption}
\usepackage{array}
\usepackage{multirow}
\usepackage{xcolor} 
\usepackage{wrapfig}
\usepackage{svg}
\usepackage{comment}
\usepackage{lineno} 

\definecolor{darkblue}{rgb}{0,0,0.75}

\hypersetup{
    hidelinks,  
    colorlinks=true,  
    linkcolor=red,  
    citecolor=darkblue   
}

\title{\textit{LTX-Video}: Realtime Video Latent Diffusion}

\author{%
  \renewcommand{\arraystretch}{1.8}
  \begin{tabular}{c c c c}
    Yoav HaCohen & Nisan Chiprut & Benny Brazowski & Daniel Shalem \\
    Dudu Moshe & Eitan Richardson & Eran Levin & Guy Shiran \\
    Nir Zabari & Ori Gordon & Poriya Panet & Sapir Weissbuch \\
    Victor Kulikov & Yaki Bitterman & Zeev Melumian & Ofir Bibi\thanks{Authors are listed with project leads first, followed by the team in alphabetical order, and concluding with senior management.} \\
  \end{tabular}\\[1ex]\\
  Lightricks\\
  \texttt{ltx-video@lightricks.com}
}

\newcommand{\citeme}[1]{\citetalias{#1}~\citep{#1}}

\newcommand{\successprompt}[1]{\textcolor{darkgreen}{#1}}
\definecolor{darkgreen}{rgb}{0.0, 0.4, 0.0}

\defcitealias{ho2020denoising}{DDPM}
\defcitealias{brooks2024video}{Sora}
\defcitealias{chen2023pixart}{Pixart-$\alpha$}
\defcitealias{xie2024sana}{SANA}
\defcitealias{blattmann2023stable}{SVD}
\defcitealias{guo2023animatediff}{AnimateDiff}
\defcitealias{bar2024lumiere}{Lumiere}
\defcitealias{girdhar2024factorizing}{EmuVideo}
\defcitealias{polyak2024movie}{MovieGen}
\defcitealias{yang2024cogvideox}{CogVideoX}
\defcitealias{opensoraplan}{Open-Sora Plan}
\defcitealias{opensora}{Open-Sora}
\defcitealias{jin2024pyramidal}{PyramidFlow}
\defcitealias{chen2024deepcompressionautoencoderefficient}{DC-VAE}
\defcitealias{peebles2023scalable}{DiT}
\defcitealias{lu2024fit}{FiT}
\defcitealias{largedit}{LargeDiT}
\defcitealias{hoogeboom2023simple}{SimpleDiffusion}
\defcitealias{crowson2024scalable}{H-DiT}
\defcitealias{esser2024scaling}{SD3}
\defcitealias{chen2022scaling}{ScalingViT}
\defcitealias{su2024roformer}{RoPE}
\defcitealias{lei2016layer}{LayerNorm}
\defcitealias{nichol2021glide}{GLIDE}
\defcitealias{saharia2022photorealistic}{Imagen}
\defcitealias{ho2022imagen}{ImagenVideo}
\defcitealias{betker2023improving}{DALL-E-3}
\defcitealias{zhang2018unreasonable}{LPIPS}
\defcitealias{podell2023sdxl}{SD-XL}
\defcitealias{karras2019style}{StyleGAN}
\defcitealias{raffel2020exploring}{T5-XXL}
\defcitealias{flux}{FLUX.1}
\defcitealias{auraflow}{AuraFlow}
\defcitealias{ni2023conditional}{LFDM}
\defcitealias{zhang2023i2vgen}{I2VGen-XL}
\defcitealias{tian2024reducio}{REDUCIO!}
\defcitealias{siamese}{Siamese Network}
\defcitealias{jolliffe2002principal}{PCA}
\defcitealias{zhang2024pixel}{Pixel-loss}
\defcitealias{kong2024hunyuanvideo}{HunyuanVideo}

\begin{document}

\maketitle

\begin{abstract}
We introduce \textit{LTX-Video}, a transformer-based latent diffusion model that adopts a holistic approach to video generation by seamlessly integrating the responsibilities of the Video-VAE and the denoising transformer. Unlike existing methods, which treat these components as independent, \textit{LTX-Video} aims to optimize their interaction for improved efficiency and quality. At its core is a carefully designed Video-VAE that achieves a high compression ratio of 1:192, with spatiotemporal downscaling of $32 \times 32 \times 8$ pixels per token, enabled by relocating the patchifying operation from the transformer’s input to the VAE’s input.
Operating in this highly compressed latent space enables the transformer to efficiently perform full spatiotemporal self-attention, which is essential for generating high-resolution videos with temporal consistency. However, the high compression inherently limits the representation of fine details. To address this, our VAE decoder is tasked with both latent-to-pixel conversion and the final denoising step, producing the clean result directly in pixel space. This approach preserves the ability to generate fine details without incurring the runtime cost of a separate upsampling module.
Our model supports diverse use cases, including text-to-video and image-to-video generation, with both capabilities trained simultaneously. It achieves faster-than-real-time generation, producing 5 seconds of 24 fps video at 768×512 resolution in just 2 seconds on an Nvidia H100 GPU, outperforming all existing models of similar scale. The source code and pre-trained models are publicly available\footnote{\href{https://github.com/Lightricks/LTX-Video}{https://github.com/Lightricks/LTX-Video}.}, setting a new benchmark for accessible and scalable video generation.
\end{abstract}

\section{Introduction}

\begin{figure}[htbp]
\scriptsize
\centering
\setlength\tabcolsep{1pt} 
\renewcommand\arraystretch{0.8} 

\begin{tabular}{@{}cccc@{}}
    \includegraphics[width=0.24\textwidth]{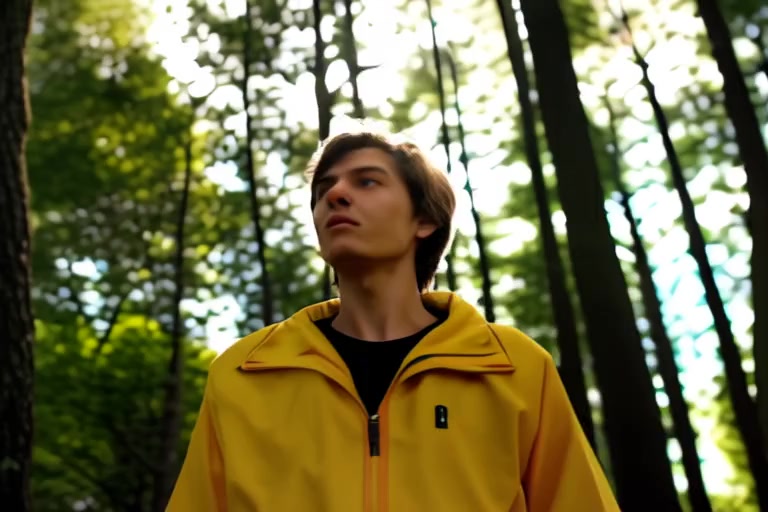} &
    \includegraphics[width=0.24\textwidth]{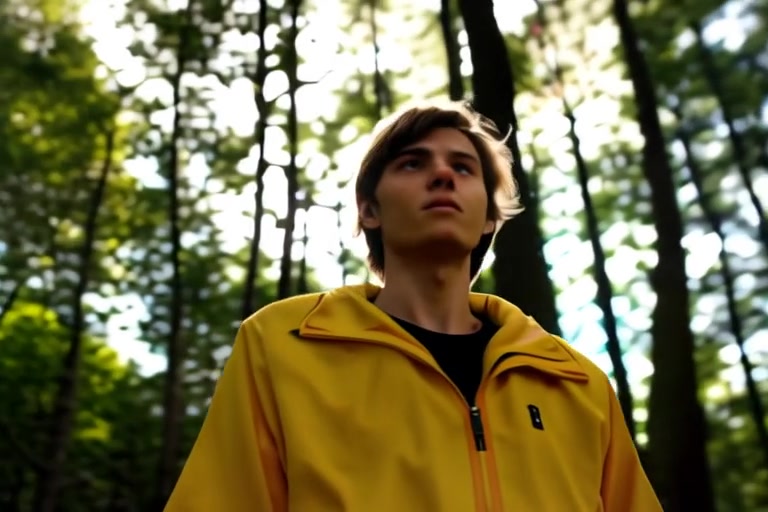} &
    \includegraphics[width=0.24\textwidth]{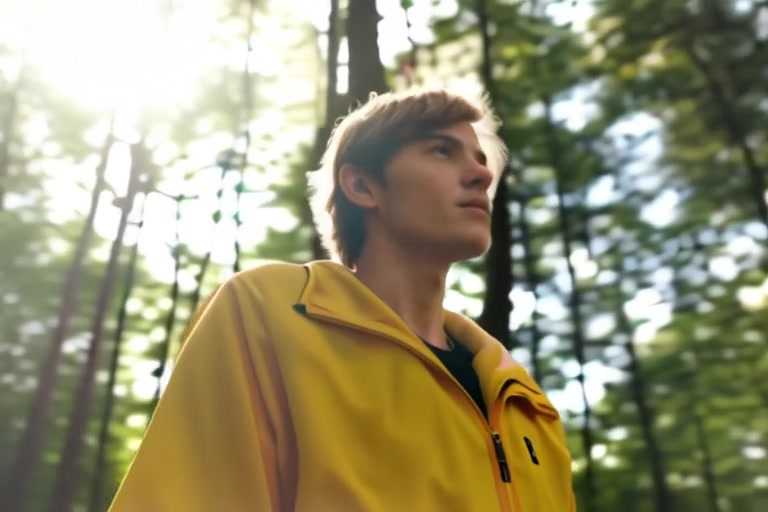} &
    \includegraphics[width=0.24\textwidth]{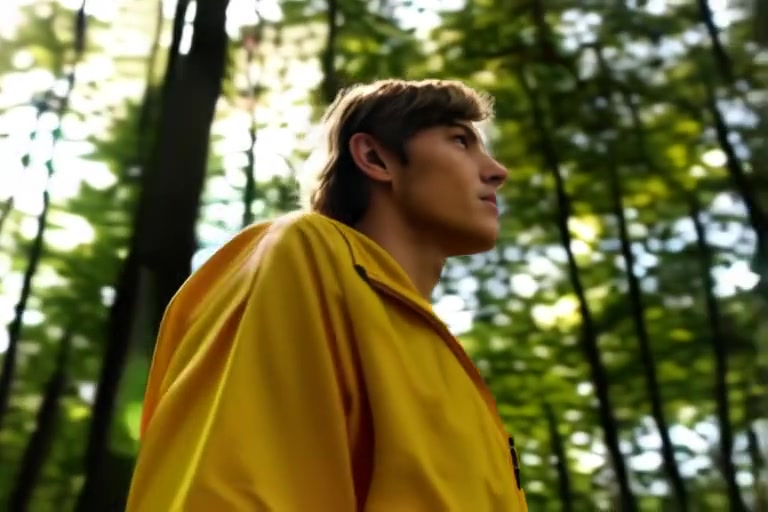} \\
\end{tabular}

``A young man with blond hair wearing a yellow jacket stands in a forest and looks around. He has light skin and his hair is styled with a middle part. He looks to the left and then to the right, his gaze lingering in each direction. The camera angle is low, looking up at the man, and remains stationary throughout the video. The background is slightly out of focus, with green trees and the sun shining brightly behind the man. The lighting is natural and warm, with the sun creating a lens flare that moves across the man’s face. The scene is captured in real-life footage.''

\vspace{0.2cm}

\begin{tabular}{@{}ccccc@{}}

\fbox{\includegraphics[width=0.24\textwidth]{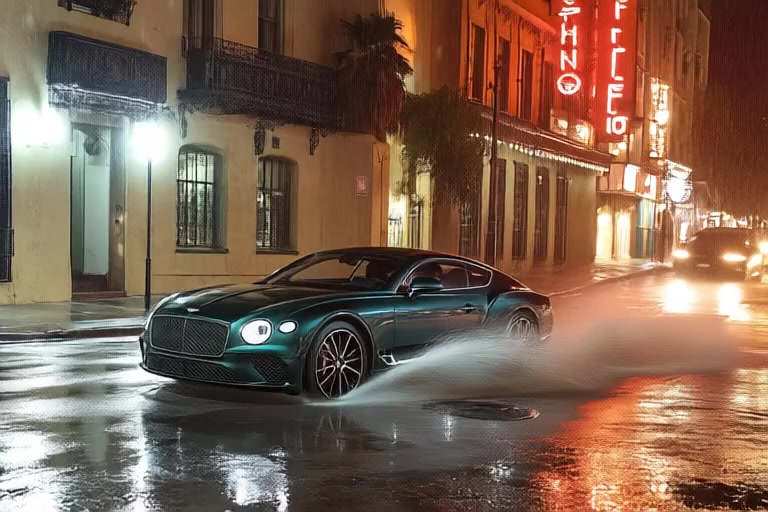}} & 
\includegraphics[width=0.24\textwidth]{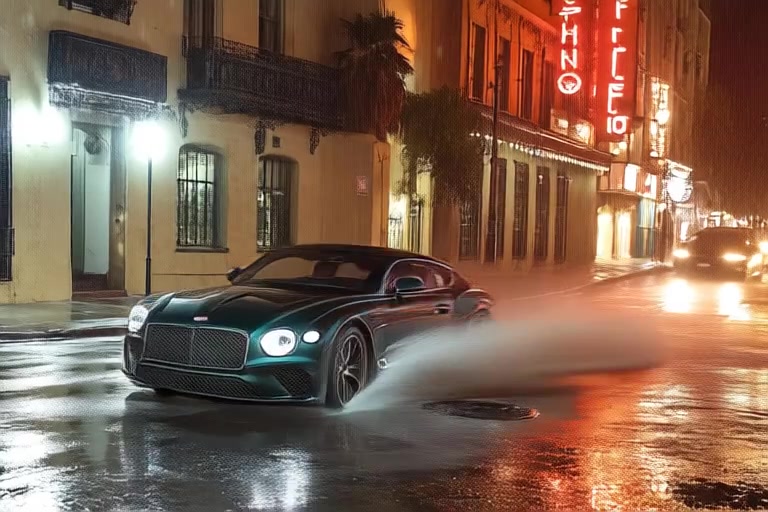} & 
\includegraphics[width=0.24\textwidth]{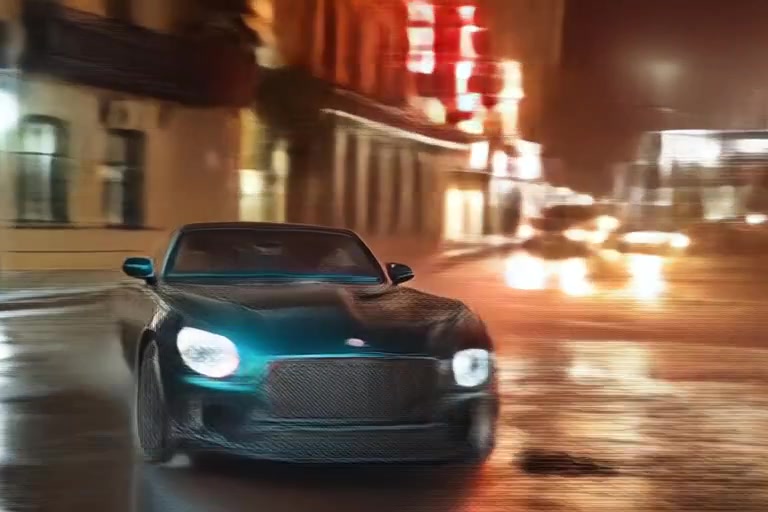} & 
\includegraphics[width=0.24\textwidth]{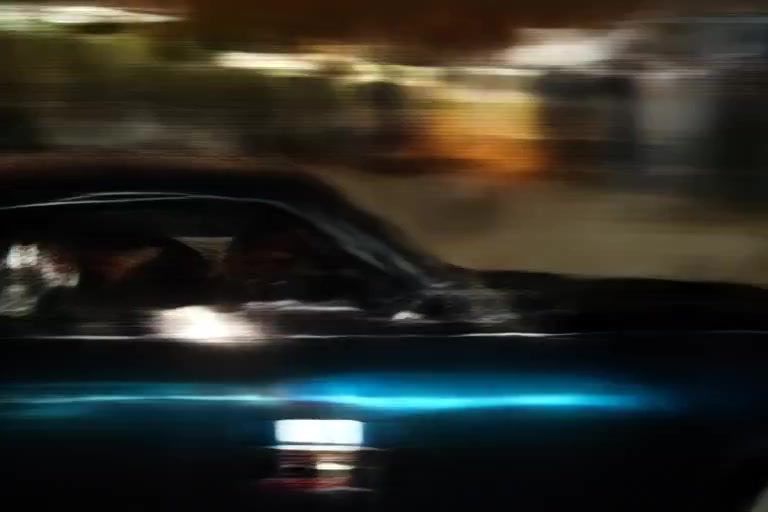} \\

\fbox{\includegraphics[width=0.24\textwidth]{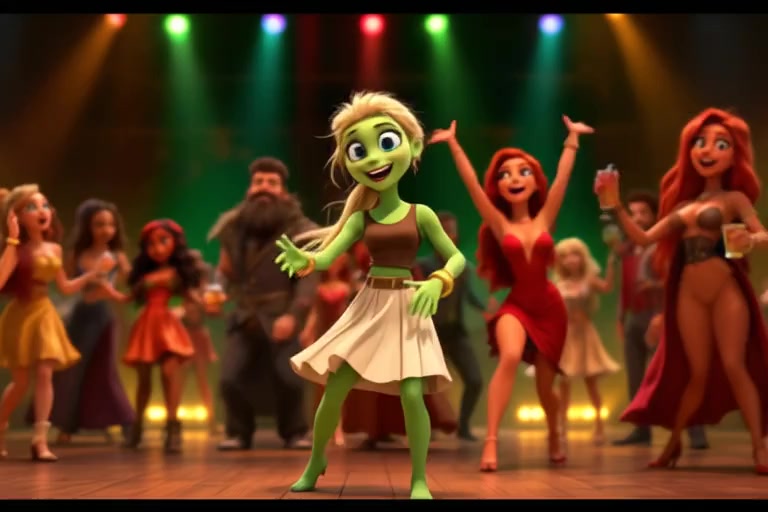}} &
\includegraphics[width=0.24\textwidth]{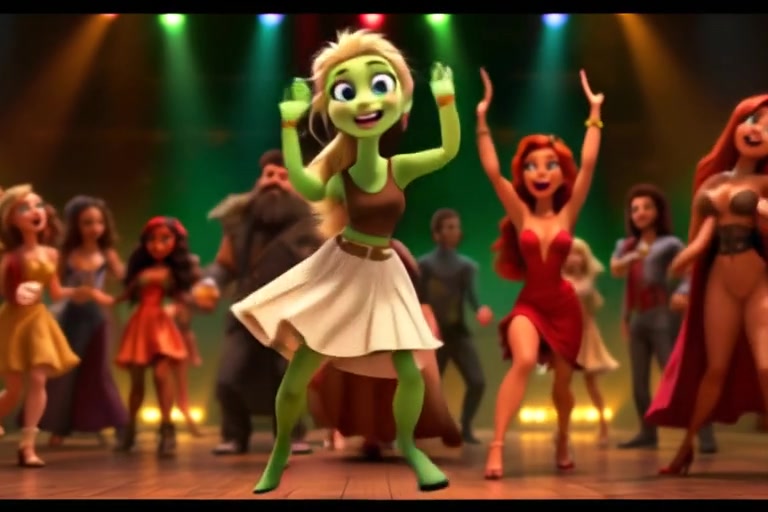} &
\includegraphics[width=0.24\textwidth]{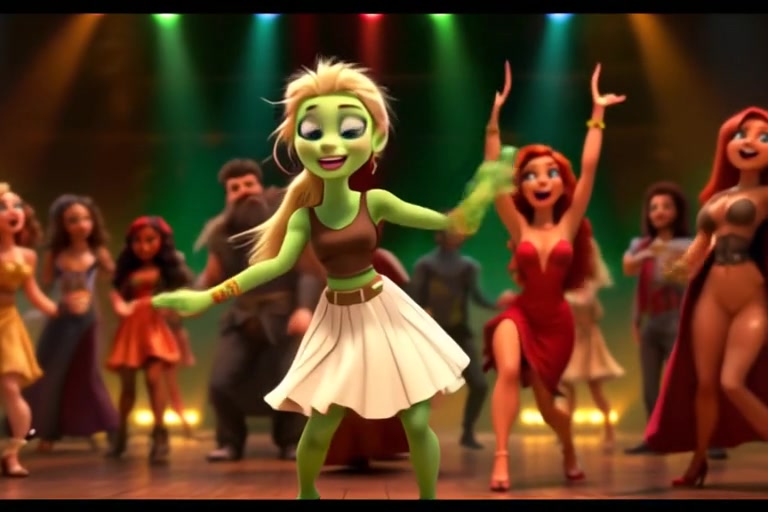} &
\includegraphics[width=0.24\textwidth]{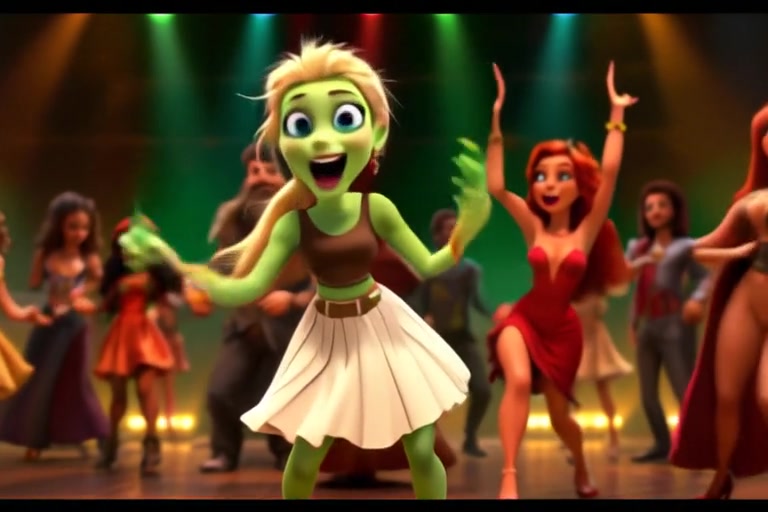} \\

\end{tabular}
\caption{Text-to-video (first row) and image-to-video samples (last 2 rows, conditioned on the left frame) generated by \textit{LTX-Video}, highlighting our model's high level of prompt adherence, visual quality and motion fidelity. Each row shows evenly-spaced frames from a generated 5-second video.
}
\label{fig:results_1}
\end{figure}

The rise of text-to-video models such as \citeme{brooks2024video}, \citeme{polyak2024movie}, \citeme{yang2024cogvideox}, \citeme{opensoraplan} and \citeme{jin2024pyramidal} has demonstrated the effectiveness of spatiotemporal transformers with self-attention and a global receptive field, coupled with 3D VAEs for spatiotemporal compression. While these approaches validate the fundamental architectural choices, they often rely on conventional VAE designs that may not optimally balance spatial and temporal compression.

Concurrently with our work, \citeme{chen2024deepcompressionautoencoderefficient} demonstrated that text-to-image transformer-based diffusion models perform more effectively when paired with VAEs that employ higher spatial compression factors and a high dimensional latent spaces with up to 64 channels. However, extending this approach to video presents significant challenges.

Inspired by these developments and the success of diffusion models in generating high-resolution images and video, we propose \textit{LTX-Video}, a transformer-based latent diffusion model that equally prioritizes both spatial and temporal dimensions. Our approach features a carefully designed VAE architecture that achieves higher spatial compression while maintaining video quality through an increased latent depth of 128 channels. This design choice not only enables more efficient processing of video data but also results in a highly performant 3D VAE implementation.

Latent diffusion models trade the ability to apply pixel-level training loss for improved training efficiency, often at the expense of generating plausible high-frequency details. \citeme{brooks2024video} and \citeme{polyak2024movie} mitigate this limitation by applying a second-stage diffusion model for generating the high-resolution output. \citeme{zhang2024pixel} attempt to address this issue by incorporating pixel-level loss on VAE-decoded noisy latents, but retained the entire generation process within the limits of the compressed latent space. In contrast, we propose tasking the VAE decoder with performing the \textit{last denoising step} in conjunction with converting latents to pixels. This modification is particularly impactful at high latent compression rates, where not all high-frequency details can be reconstructed and must instead be generated.

We adopt the scalable and flexible transformer architecture, known for its effectiveness in various applications, enabling our model to generate images and videos across a range of sizes and durations. Building upon \citeme{chen2023pixart}'s architecture, which extends the \citeme{peebles2023scalable} framework to be conditioned on open text inputs rather than constrained to ImageNet class labels, we introduce several key enhancements. Specifically, we replace traditional absolute positional embeddings with Rotary Positional Embeddings (\citeme{su2024roformer}) enhanced by normalized fractional coordinates, which improve spatial and temporal coherence in video generation. Additionally, we normalize the \textit{key} and \textit{query} tensors to stabilize attention computations, enhancing robustness and increasing the entropy of attention weights.
Our approach addresses limitations in existing models, offering a more integrated and efficient solution for robust video generation.

Our model is the fastest video generation model of its kind, capable of generating videos faster than the time it takes to watch them (2 seconds to generate 121 frames at $768 \times 512$ pixels and 20 diffusion steps on an Nvidia H100 GPU), while outperforming all available models of similar scale (2B parameters, before distillation).

In addition to text-to-video generation, we extend our model’s functionality to handle image-to-video, a practical application in content creation. Through a simple timestep-based conditioning mechanism, the model can be conditioned on any part of the input video without requiring additional parameters or special tokens.

See Fig~\ref{fig:results_1} for text-to-video and image-to-video samples generated by \textit{LTX-Video}. Additional samples are provided in figures \ref{fig:more_results} and \ref{fig:more_results_img_to_video}.

Our main contributions are:

\begin{itemize}
    \item \textbf{A holistic approach to latent diffusion}: \textit{LTX-Video} seamlessly integrates the Video-VAE and the denoising transformer, optimizing their interaction within a compressed latent space and sharing the denoising objective between the transformer and the VAE's decoder.
    
    \item \textbf{High-compression Video-VAE leveraging novel loss functions}: By relocating the patchifying operation to the VAE and introducing novel loss functions, we achieve a 1:192 compression ratio with spatiotemporal downsampling of $32 \times 32 \times 8$, enabling the generation of high-quality videos at unprecedented speed.
    
    \item \textbf{\textit{LTX-Video} -- A fast, accessible, and high-quality video generation model}: We train and evaluate our enhanced diffusion-transformer architecture and publicly release \textit{LTX-Video}, a faster-than-real-time text-to-video and image-to-video model with fewer than 2B parameters.

\end{itemize}

\section{Method}

\begin{figure}[htbp]
    \centering
    \includegraphics[width=\textwidth]{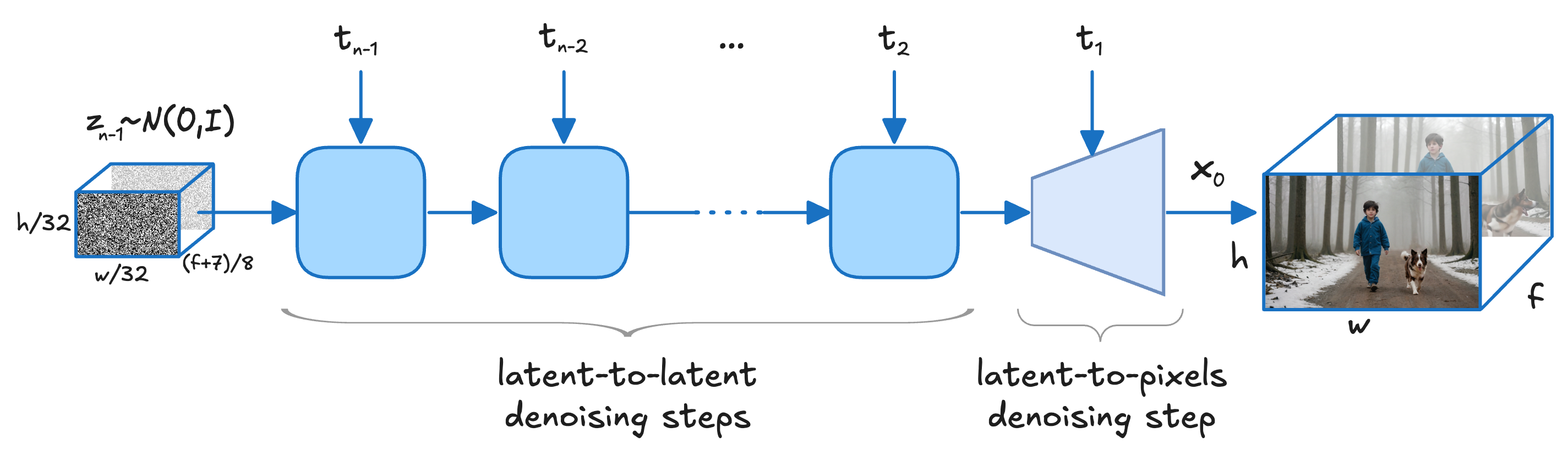}
    \caption{\textit{LTX-Video} holistic denoising strategy -- latent-to-latent diffusion denoising steps + final latent-to-pixels denoising step.}
    \label{fig:model}
\end{figure}

To facilitate the faster-than-realtime operation of \textit{LTX-Video} while maintaining high visual quality, motion fidelity, and prompt adherence, we employ a holistic approach to latent diffusion, optimizing the interaction between the Video VAE and the diffusion-transformer. We utilize a high-dimensional latent space with a high compression rate of 1:192 and spatiotemporal downsampling of $32 \times 32 \times 8$. To support the generation of high-frequency details, we assign the VAE decoder the task of performing the last denoising step alongside converting the latents to pixels, as illustrated in Fig~\ref{fig:model}.

Our holistic approach, along with key design changes in the Video VAE architecture, loss functions, and in the diffusion-transformer architecture, enable generating high quality videos despite the high pixels-to-tokens ratio. These enhancements, which were crucial to the success of our approach, are highlighted in the following sections.

\subsection{Video VAE}
\label{sec:vae}

Operating in a compressed latent space is key to the success of text-to-video diffusion-transformer models: training and inference time for these models is dominated by the number of tokens (the attention operation is quadratic in the number of tokens), while the diffusion process benefits from the compressed latent representation, as it decreases the inherent information redundancy of the original signal (\citeme{hoogeboom2023simple}, \citeme{esser2024scaling}).

Recent text-to-video models (\citeme{yang2024cogvideox}, \citeme{polyak2024movie}, 
\citeme{jin2024pyramidal}, \citeme{opensoraplan}, \citeme{kong2024hunyuanvideo}) employ VAEs that downscale the spatio-temporal dimensions by either $8 \times 8 \times 4$ or $8 \times 8 \times 8$,  while increasing the number of channels from 3 to 16. These configurations result in a total compression factor of 1:48 or 1:96.
Subsequently, a \textit{patchifier} collects latent patches of size $2 \times 2 \times 1$ into tokens, achieving an effective pixels-to-tokens ratio of 1:1024 or 1:2048, at the input of the transformer.

In contrast, our Video-VAE applies a spatio-temporal compression of $32 \times 32 \times 8$ with 128 channels, resulting in a total compression of 1:192 (twice the typical compression) and a pixels-to-tokens ratio of 1:8192 (four times the typical ratio), without requiring a patchifier. See Table~\ref{tab:model_comparison} for additional details.

The challenge of information redundancy in pixel space at high resolutions has been highlighted by \citeme{hoogeboom2023simple}, who mitigated it by increasing the amount of noise added at each diffusion step. This challenge also applies to higher-resolution latents (\citeme{esser2024scaling}), as well as to redundancy in the time dimension. Therefore, spatial and temporal compression is crucial. 

\begin{figure}[htbp]
    \centering
    \begin{subfigure}[t]{0.8\textwidth} 
        \includegraphics[width=\textwidth]{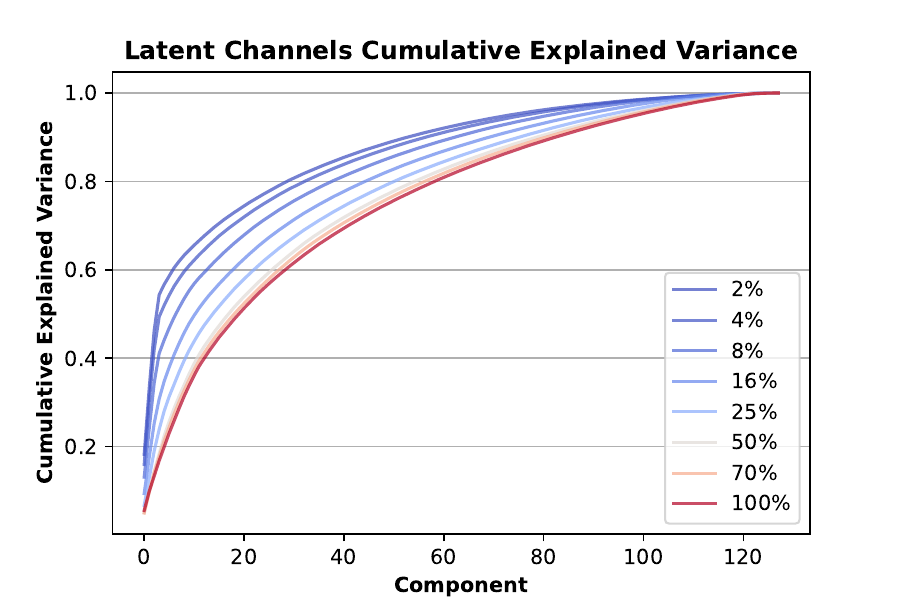}
        \caption{Latent channels cumulative explained variance at different training steps.}
        \label{fig:explained_varaince}
    \end{subfigure}%
    \hfill
    \begin{subfigure}[t]{0.2\textwidth} 
        \vbox{ 
            \includegraphics[width=\textwidth]{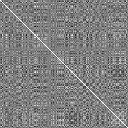}
            \caption{Correlation at 4\%}
            \label{fig:latent_corr_init}
            \vspace{5pt} 
            \includegraphics[width=\textwidth]{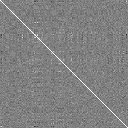}
            \caption{Final Correlation}
            \label{fig:latent_corr_final}
        }
    \end{subfigure}
    \caption{Latent-space redundancy. (a) Cumulative explained-variance of latent channels at different training steps (2\% - 100\% of training). As training progresses, the redundancy reduces and components contribute more evenly to the variance. (b, c) Latent channels auto-correlation matrices: high off-diagonal values early (at 4\% of total training steps) and near-zero at training completion.}
    \label{fig:latent_redundancy}
\end{figure}

We analyze the redundancy in our latent space using \citeme{jolliffe2002principal} over the latent pixels of 128 video samples (see Fig~\ref{fig:latent_redundancy}). As training progresses, our VAE learns to utilize the available channels and reduces their redundancy. Note that naive patchification of the latents before passing them to the transformer, as done by recent models, does not contribute to reducing the redundancy. 

This pixels-to-latents compression is a critical process in our approach and the main enabler of its unprecedented speed.

\begin{figure}[htbp]
    \centering
    \begin{subfigure}[t]{0.49\textwidth}
        \centering
        \includegraphics[scale=0.115]{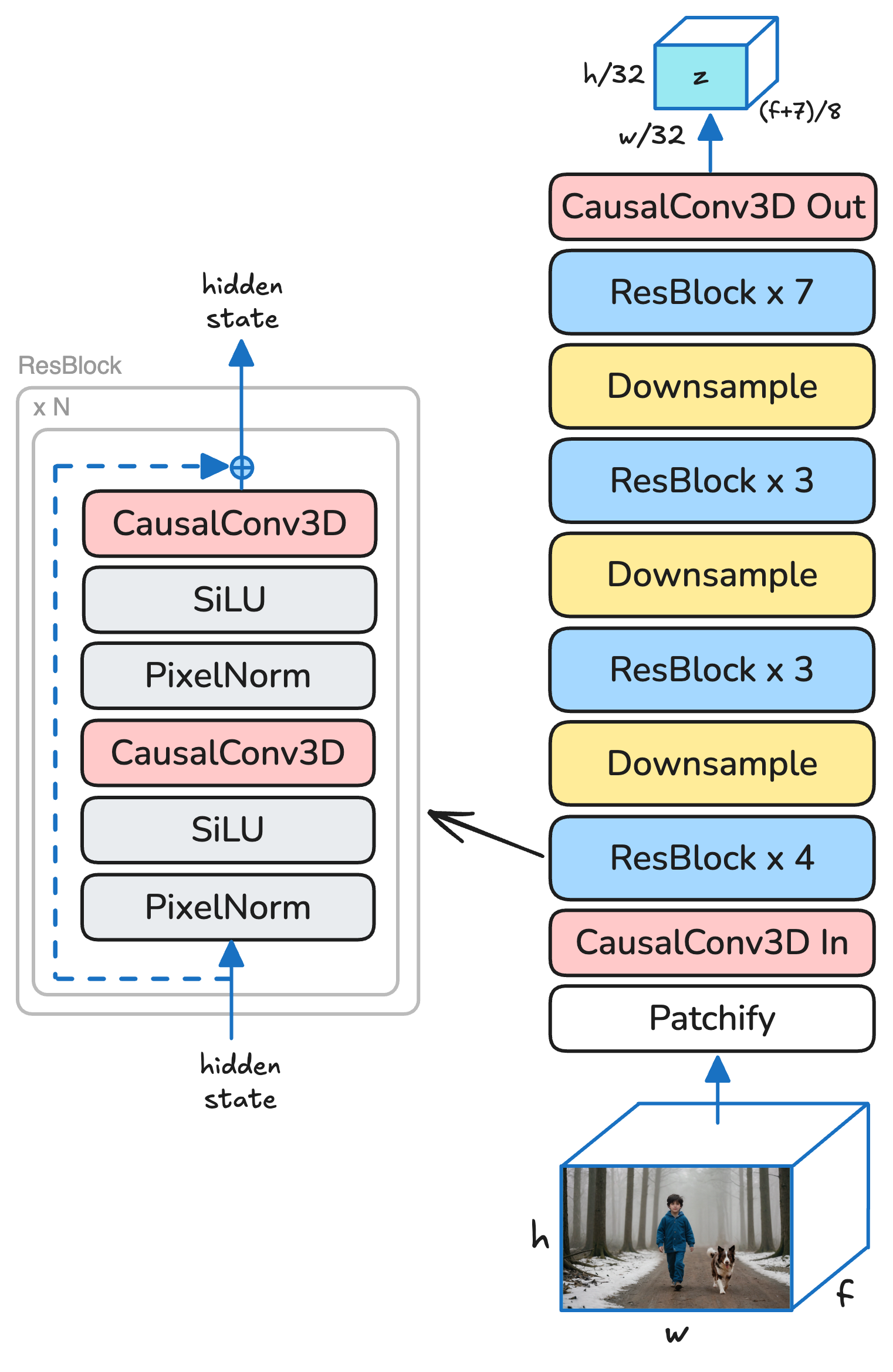}
        \caption{Causal Encoder}
        \label{fig:vae_dec}
    \end{subfigure}
    \begin{subfigure}[t]{0.49\textwidth}
        \centering
        \includegraphics[scale=0.115]{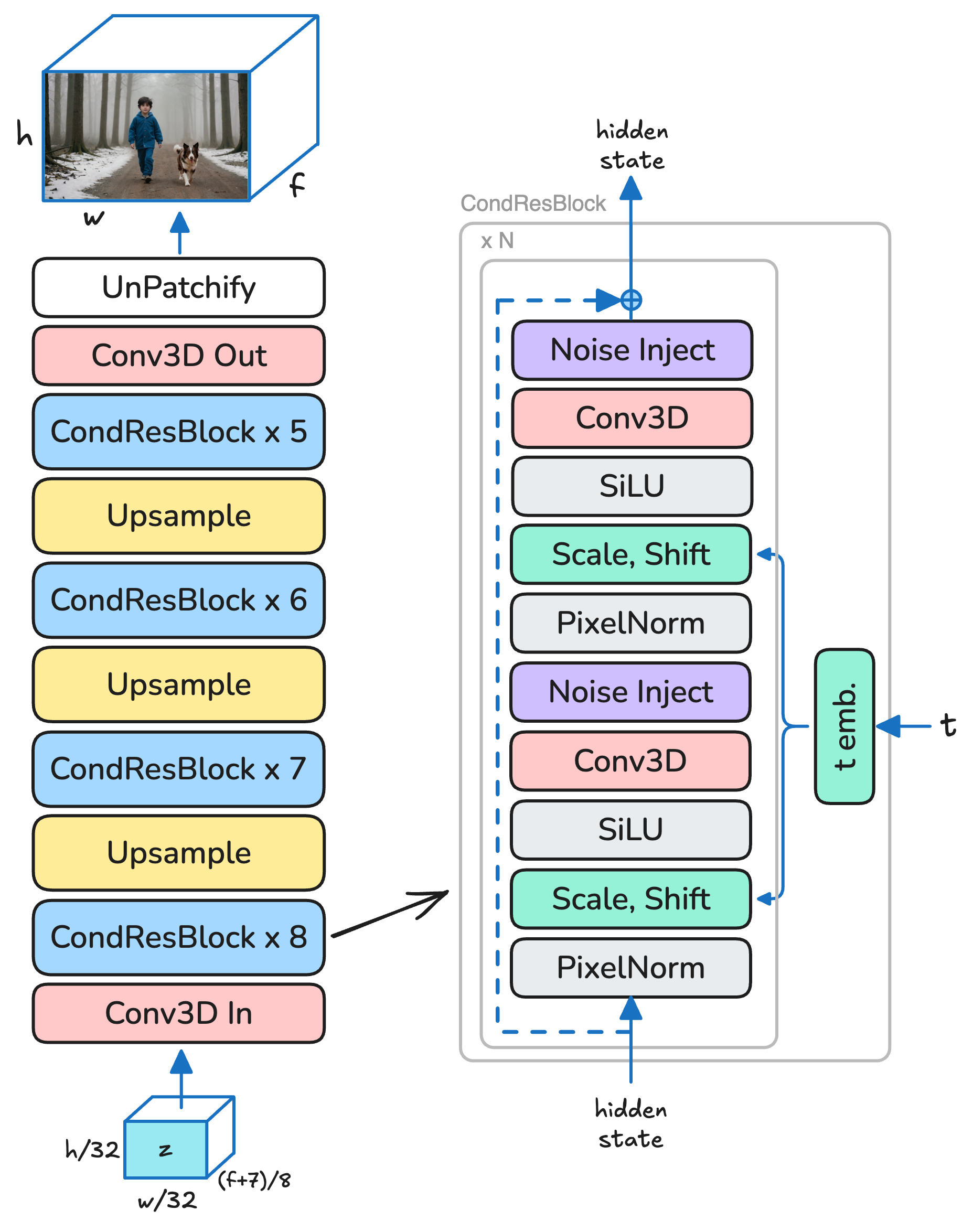}
        \caption{Denoising Decoder}
        \label{fig:vae_enc}
    \end{subfigure}
     \caption{The \textit{LTX-Video} Video-VAE architecture: (a) \textit{Causal Encoder} utilizing 3D Causal Convolutions, applying $32 \times 32 \times 8$ compression (except the first frame, which is encoded as a separate latent frame). (b) \textit{Denoising Decoder} with diffusion-timestep conditioning and multi-layer noise injection.}
     \label{fig:vae_arch}
\end{figure}

To facilitate the high-compression rate without loss of quality, we introduce several key enhancements over current VAEs, as described in the following sections. We trained and compared several VAEs designed to efficiently map both videos and images into a unified latent space. These VAEs represent a key component of our model. See Fig~\ref{fig:vae_arch} for our VAE architecture. Notably, compared to standard DiT diffusion models, we move the patchifying layer from the beginning of the transformer to the beginning of the VAE encoder and task the VAE decoder with performing the last denoising step in conjunction with decoding the latents into pixels.

\subsubsection{Shared Diffusion Objective}

Rectified-flow models are designed to map noisy latents, $z_{t_i} = (1 - t_i) z_0 + t_i \epsilon$, to expected clean latents, $z_0$, such that $f^\theta\left(z_{t_i}, t_i\right)=z_0$\footnote{In most implementations, the model $f^\theta$ predicts an intermediate term from which $z_0$ can be inferred in closed form.}.
By initializing $z_0$ as pure noise ($\epsilon$) and iteratively updating $z_0$, with decreasing noise levels, $\left\{t_N, t_{N-1}, t_{N-2}, \ldots, t_2, t_1\right\}$, the predicted $z_0$ becomes progressively cleaner until it closely matches the distribution of the training set.

In practice, however, the number of iterations is limited, and true convergence is rarely achieved. Consequently, residual uncertainty persists.

In \textit{latent} diffusion, where $z_0 = \mathcal{E}(x_0)$ is a compressed representation of the data sample $x_0$, this residual uncertainty often manifests as out-of-distribution inputs to the decoder $\mathcal{D}$, resulting in artifacts in the reconstructed pixel space, $x_0=\mathcal{D}(z_0)$.
Our experiments show that this effect is exacerbated when the encoder $\mathcal{E}$ operates at high compression rates, particularly in regions with high-frequency signals that are poorly represented in the compressed latent space.

As operating within a compressed latent space is critical for the practical usability of video diffusion transformers, various strategies have been proposed to address these artifacts. For instance, \citeme{polyak2024movie} introduced a diffusion-based upsampler operating in a less aggressively compressed latent space, while \citeme{brooks2024video}  proposed an upsampler working directly in pixel space, conditioned on the latent outputs of the base model. While these approaches effectively mitigate artifacts, they incur significant computational and runtime costs.

To generate precise fine details while retaining faster generation times, we propose a novel approach that fuses the decoding and denoising steps. Specifically, we train the decoder as a diffusion model that maps noisy latents to clean pixels at varying noise levels: $x_0=\mathcal{D}(z_{t_i}, t_i)=\mathcal{D}\left((1 - t_i) z_0 + t_i \epsilon, \, t_i\right)$.
Since $\mathcal{D}$ maps between spaces of different dimensionality, it cannot be applied iteratively like a standard diffusion model. However, it can execute the final denoising step, $x_0=\mathcal{D}(z_{t_1}, \, t_1)$, in a manner inaccessible to the base model. Unlike the latent-to-latent denoising base model, constrained by the limited expressiveness of the latent space, our denoising-decoder directly outputs in pixel space and is trained with pixel-space losses.

Our implementation of $\mathcal{D}(z_{t_i}, t_i)$ follows a standard coarse-to-fine latent-to-pixel decoding architecture. To condition on the timestep $t_i$, we employ adaptive normalization layers, as commonly used in U-Net-based diffusion models (e.g. \citeme{ho2020denoising}). The denoising-decoder is trained with noise levels in the range $[0, 0.2]$ corresponding to the final diffusion timestep in common noise schedulers.

\subsubsection{Reconstruction GAN (rGAN)}
A common approach in VAE training is to balance pixel-wise L2 loss, perceptual loss (\citeme{zhang2018unreasonable}), and a GAN discriminator. At high compression rates, L2 loss often produces blurry outputs. The addition of perceptual loss reduces blurriness but can introduce texture artifacts, particularly in high-motion scenarios. However, the adversarial training approach typically relies on discriminators designed for tasks unrelated to reconstruction. These discriminators are tasked with distinguishing between real and fake samples without additional context, making their job unnecessarily challenging for reconstruction-specific tasks. This challenge is particularly pronounced for Patch-GAN discriminators, which have a spatially limited context. For instance, it may be difficult for the discriminator to determine whether a blurry patch is due to depth-of-field effects or if it originates from a fake sample.

\begin{figure}[htbp]
    \centering
    \begin{subfigure}[b]{0.425\textwidth}
        \centering
        \includegraphics[width=\textwidth]{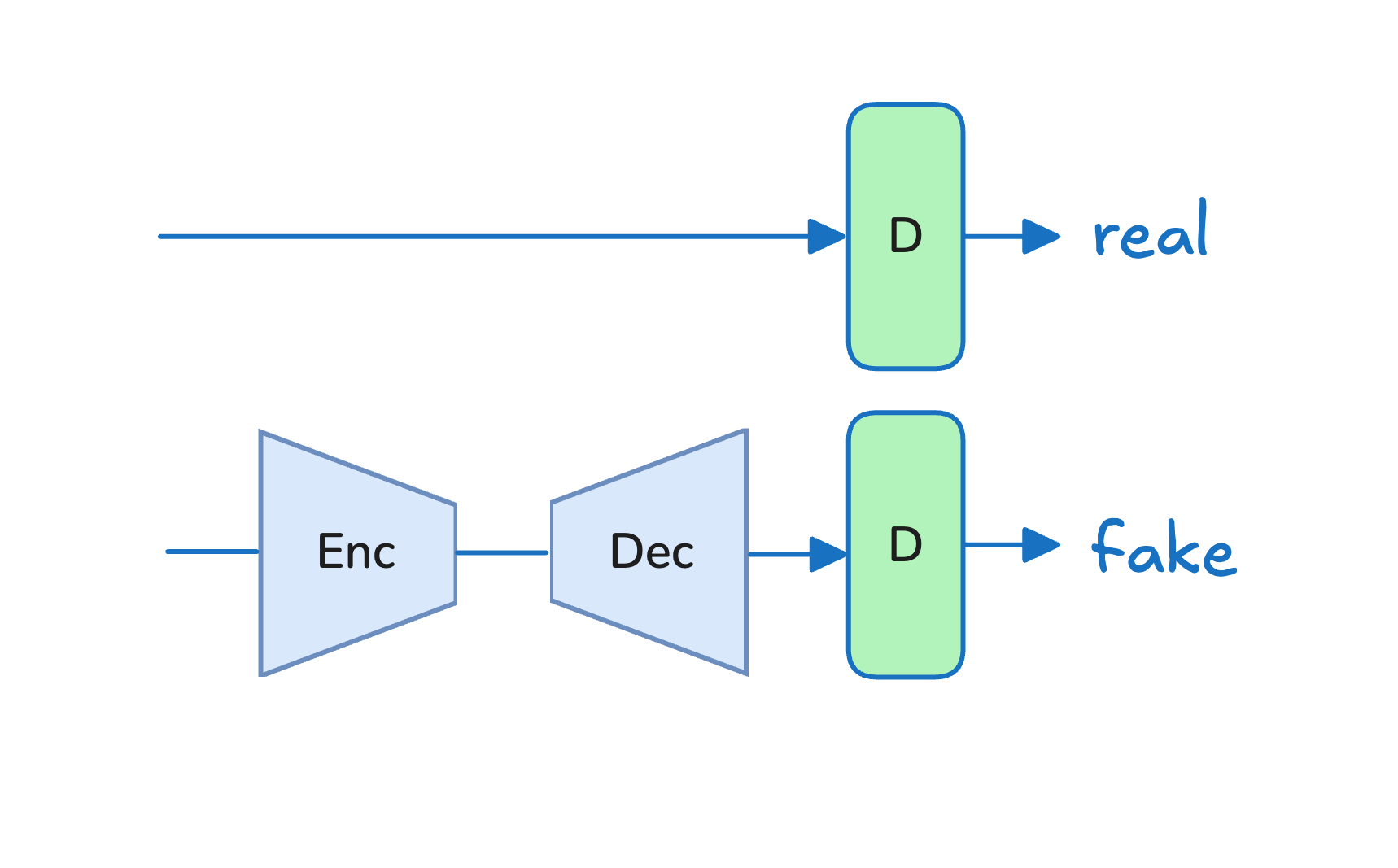}
        \caption{Traditional GAN}
        \label{fig:gan_default}
    \end{subfigure}
    \hspace{6mm}
    \begin{subfigure}[b]{0.425\textwidth}
        \centering
        \includegraphics[width=\textwidth]{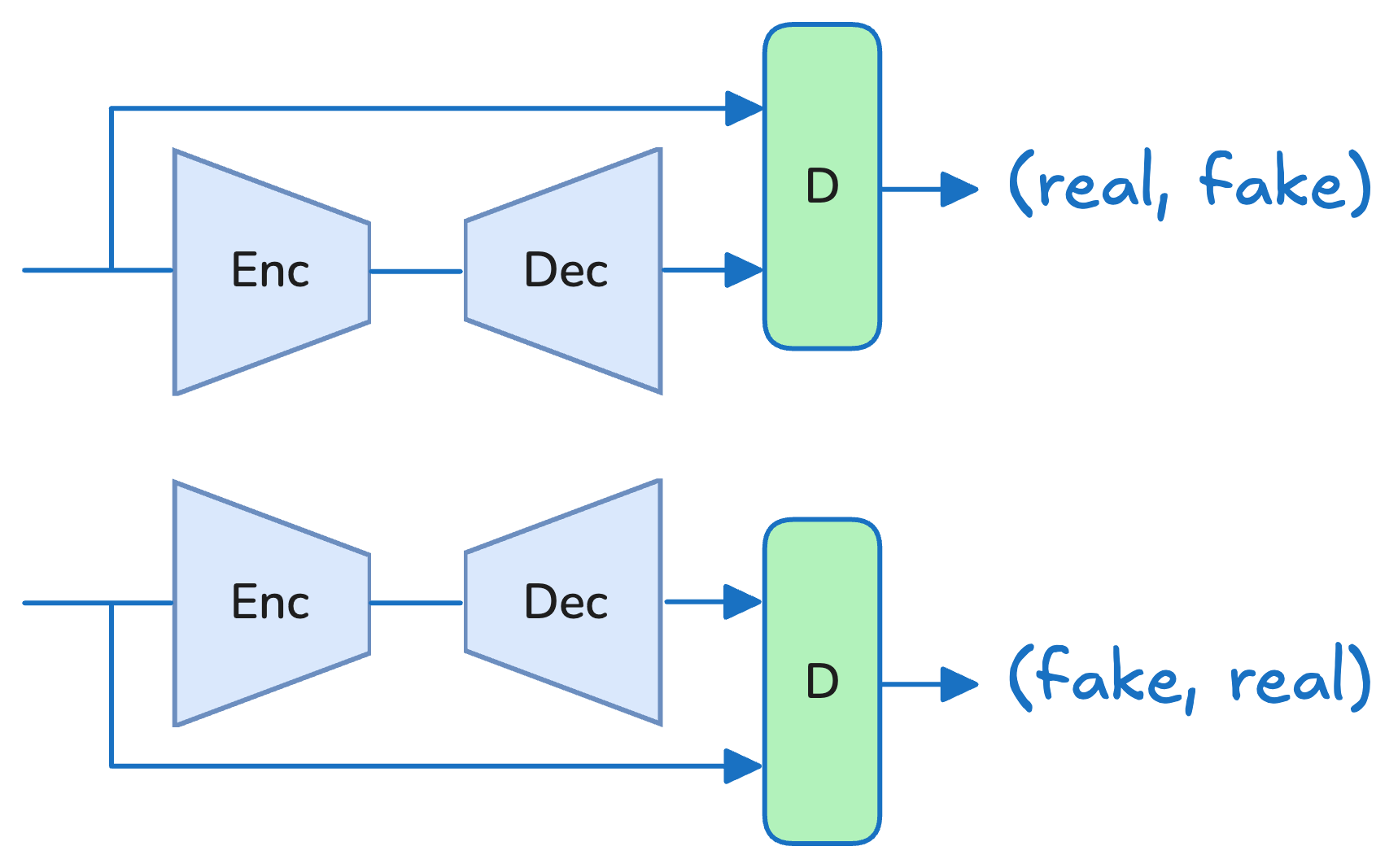}
        \caption{Reconstruction GAN}
        \label{fig:gan_rel}
    \end{subfigure}
    \caption{Our novel \textit{Reconstruction GAN} loss. (a) Traditional GAN -- the discriminator sees either a real or a reconstructed image. (b) Reconstruction GAN -- the discriminator sees both versions \textbf{of the same sample} (concatenated) and needs to decide which is the original and which is the reconstructed version.}
    \label{fig:rel_gan}
\end{figure}

To address this, we propose \textit{Reconstruction GAN} (see Fig~\ref{fig:rel_gan}), an adaptation of the traditional GAN training framework tailored to reconstruction tasks. In our approach, the discriminator is provided with both the input and the reconstructed samples for each iteration. Its goal is to determine which sample is the original (real) and which is reconstructed (fake). This relative comparison significantly simplifies the discriminator’s task and improves its ability to guide the generator.

Our experiments demonstrate that our proposed Reconstruction GAN greatly enhances GAN stability and performance. Moreover, it allows the discriminator to serve not only as a loss for matching the general distribution of real samples but also as a robust reconstruction loss, effectively balancing fidelity and perceptual quality.

\subsubsection{Multi-layer Noise Injection}
In current VAEs (\citeme{podell2023sdxl}, \citeme{chen2024deepcompressionautoencoderefficient}), stochasticity is introduced only by adding noise to the latents (according to the predicted log-variance values). Following \citeme{karras2019style}, we also inject noise at several layers of the VAE decoder allowing the generation of more diverse high-frequency details. The noise level is learned per-channel.

\subsubsection{Uniform log-variance}
We found that when using a wide latent space (large number of channels), standard KL loss tends to result in uneven latent space where some of the channels are not utilized for reconstruction but are ``sacrificed'' for satisfying the KL loss (the predicted means shrink and become close to zero and the predicted variances become close to one). To mitigate this issue we used a uniform variance for all latent channels -- a single predicted $logvar$ shared across channels. This uniformly distributes the effect of the KL loss across all channels.

\subsubsection{Video DWT Loss} To mitigate the known insufficiency of the L1 or L2 pixel loss in ensuring the reconstruction of high-frequency details, we introduced a spatio-temporal Discrete Wavelet Transform (DWT) loss. Specifically, we compute 8 3D DWT transforms for both the input and reconstructed videos and use their L1 distance as the loss.

\subsubsection{VAE Implementation Details}

The final set of losses used for training our VAE is: pixel reconstruction (MSE), Video-DWT (L1), Perceptual (LPIPS) and Reconstruction-GAN.

We tested both causal and non-causal VAEs. While non-causal VAEs are easier to train for better reconstruction, causal VAEs enable easier simultaneous training on images and videos and first-frame conditioned video generation.

We also tested architectures that employ separable convolutions, utilizing 2D spatial kernels followed by 1D temporal kernels. We found the 3D convolutions to work slightly better.

\subsection{Video Transformer}

\begin{figure}[htbp]
    \centering
    \includegraphics[width=0.46\textwidth]{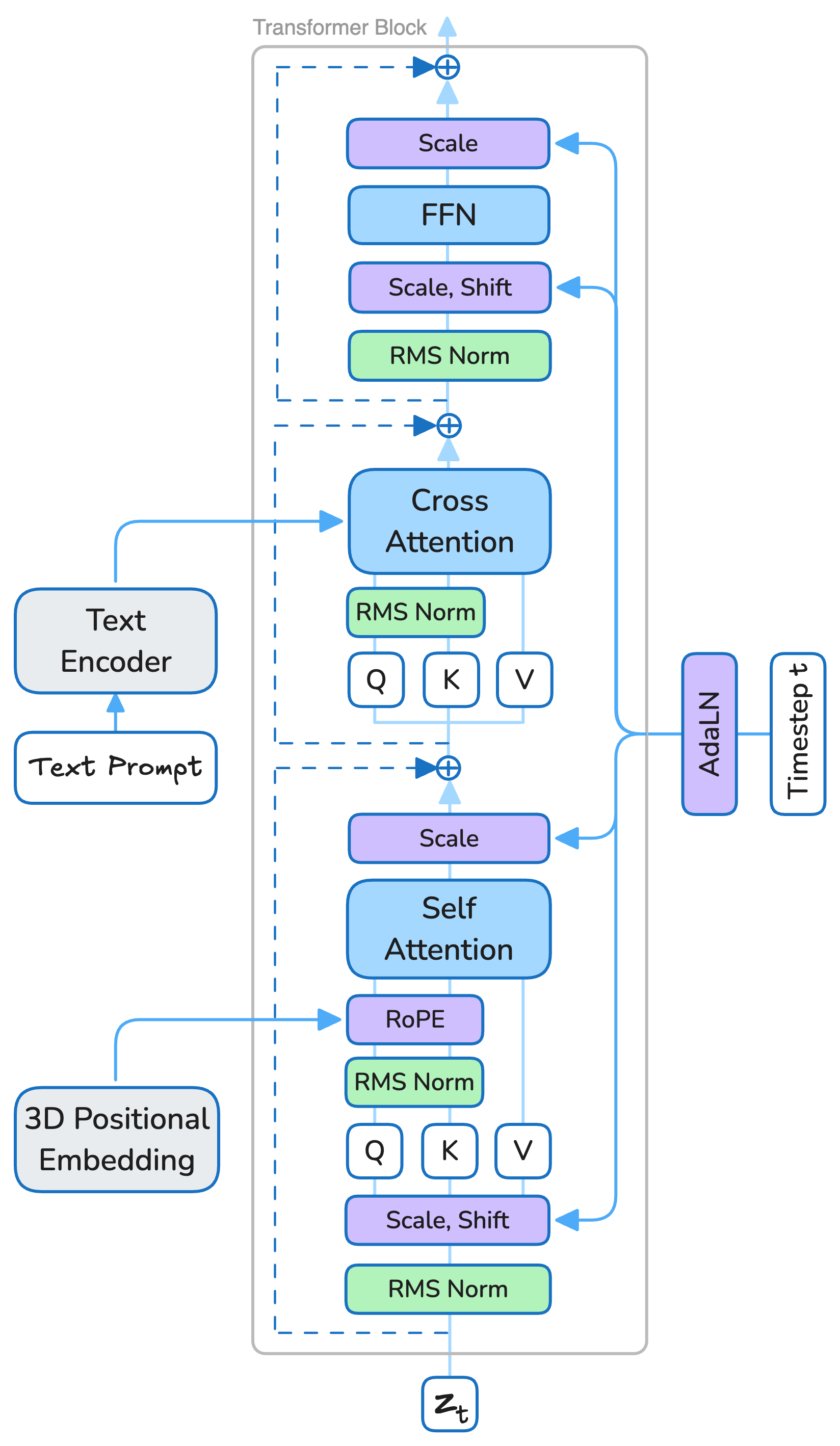}
    \caption{The \textit{LTX-Video} 3D transformer-block architecture. Our architecture builds upon \citeme{chen2023pixart}, replacing LayerNorm with RMSNorm and incorporating QK-normalization and RoPE positional embeddings.}
    \label{fig:arch}
\end{figure}

In an effort to optimize our transformer architecture for modeling diverse and complex data, we have incorporated several key modifications over the \citeme{chen2023pixart} baseline, which align with recent advancements in the field. See Fig~\ref{fig:arch} for the architecture of our 3D transformer block.

\subsubsection{Rotary Positional Embedding (RoPE)}
\label{sec:rope}
Following the practices established by \citeme{lu2024fit}, \citeme{largedit}, and \citeme{crowson2024scalable} for image generation, we replaced traditional absolute positional embeddings with Rotary Positional Embeddings (\citeme{su2024roformer}). RoPEs allow for a more dynamic and context-aware interpretation of positional information, which is crucial for managing the sequences of varying lengths and dimensions typical in video data.

\begin{figure}[htbp]
    \centering
    \begin{subfigure}[b]{0.25\textwidth}
        \centering
        \includegraphics[height=3.5cm]{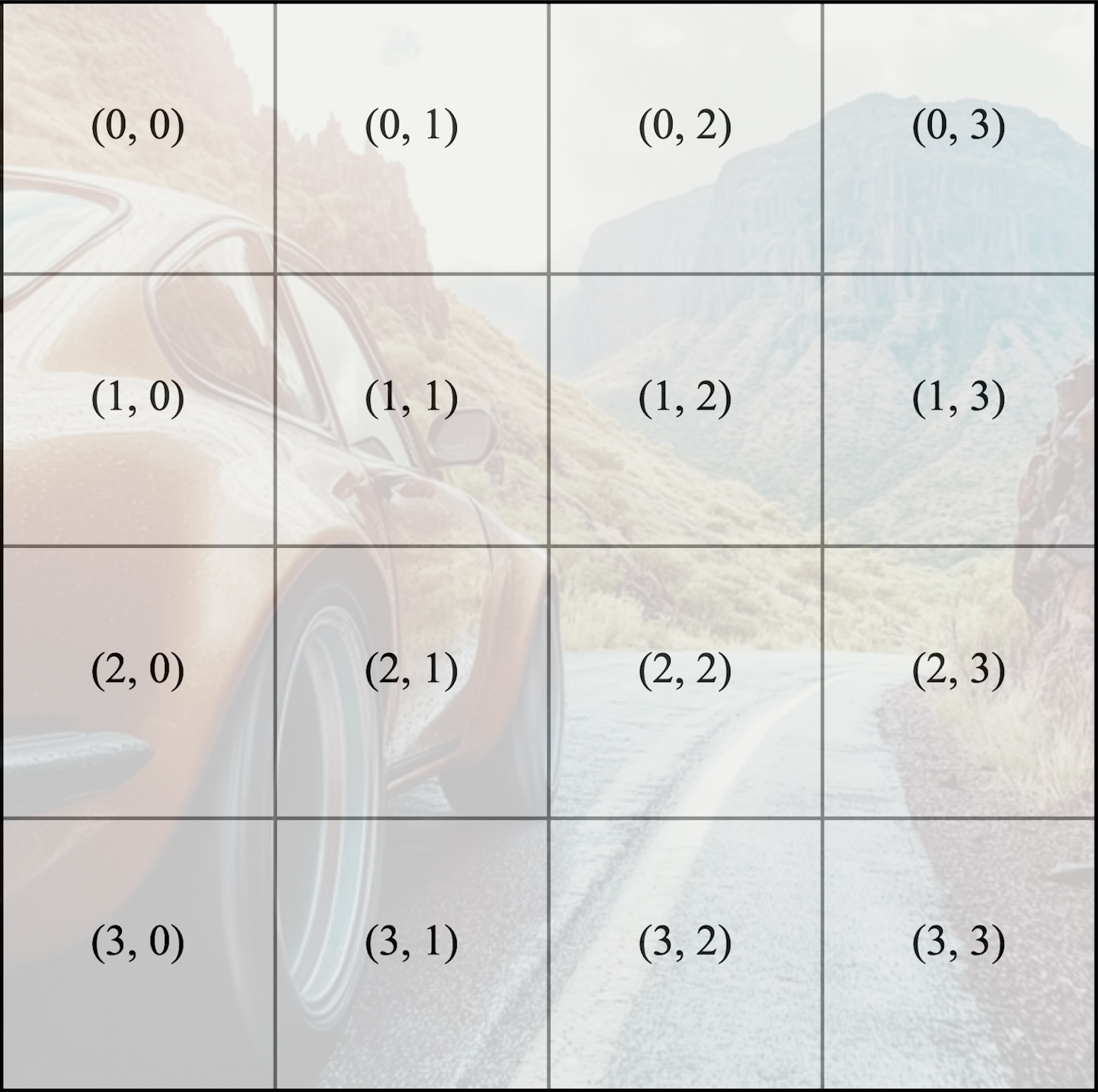}
        \caption{}
        \label{fig:pos_embs_abs}
    \end{subfigure}
    \hspace{6mm}
    \begin{subfigure}[b]{0.25\textwidth}
        \centering
        \includegraphics[height=3.5cm]{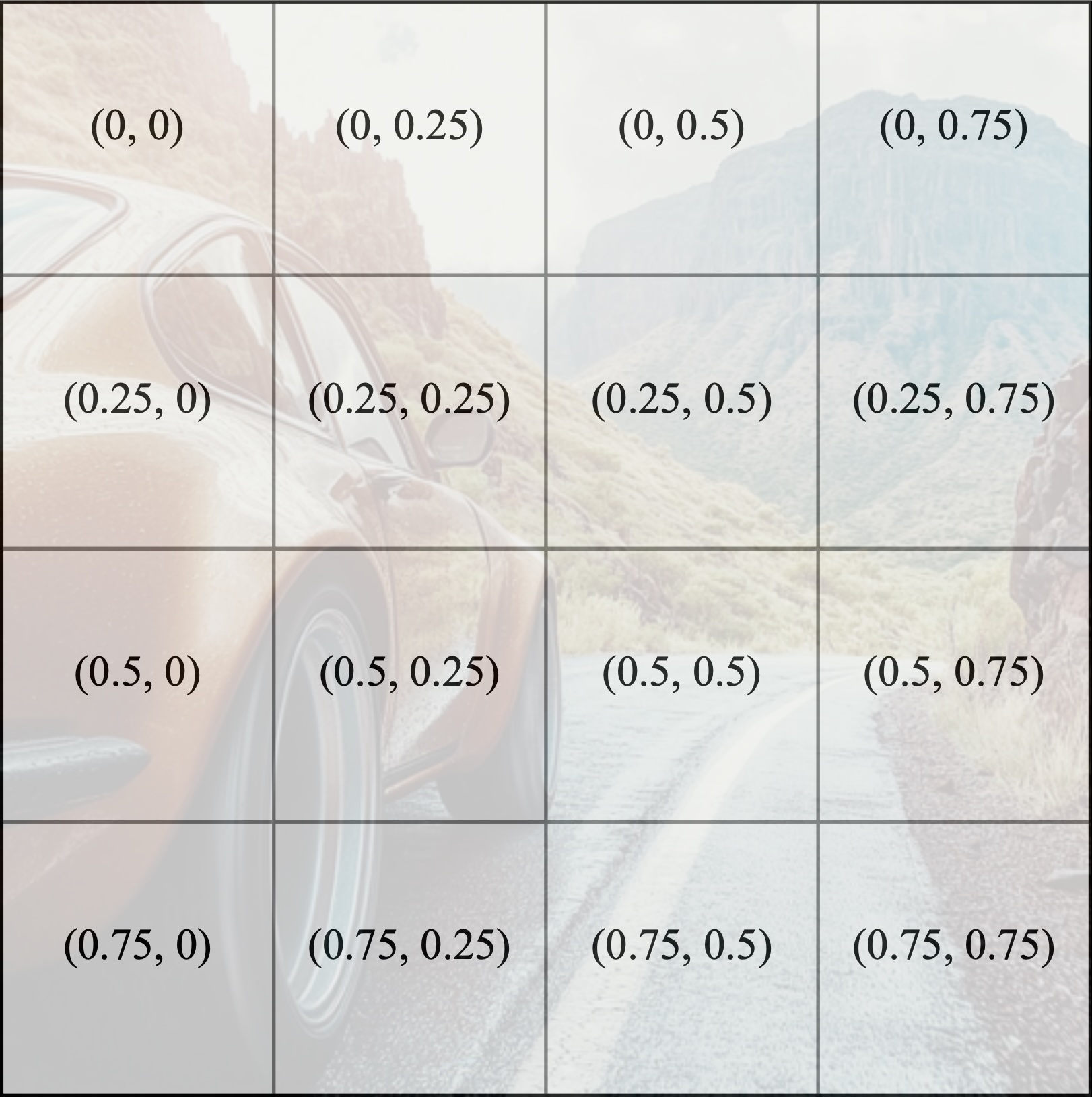}
        \caption{}
        \label{fig:pos_embs_frac}
    \end{subfigure}
    \hspace{6mm}
    \begin{subfigure}[b]{0.25\textwidth}
        \centering
        \includegraphics[height=3.5cm]{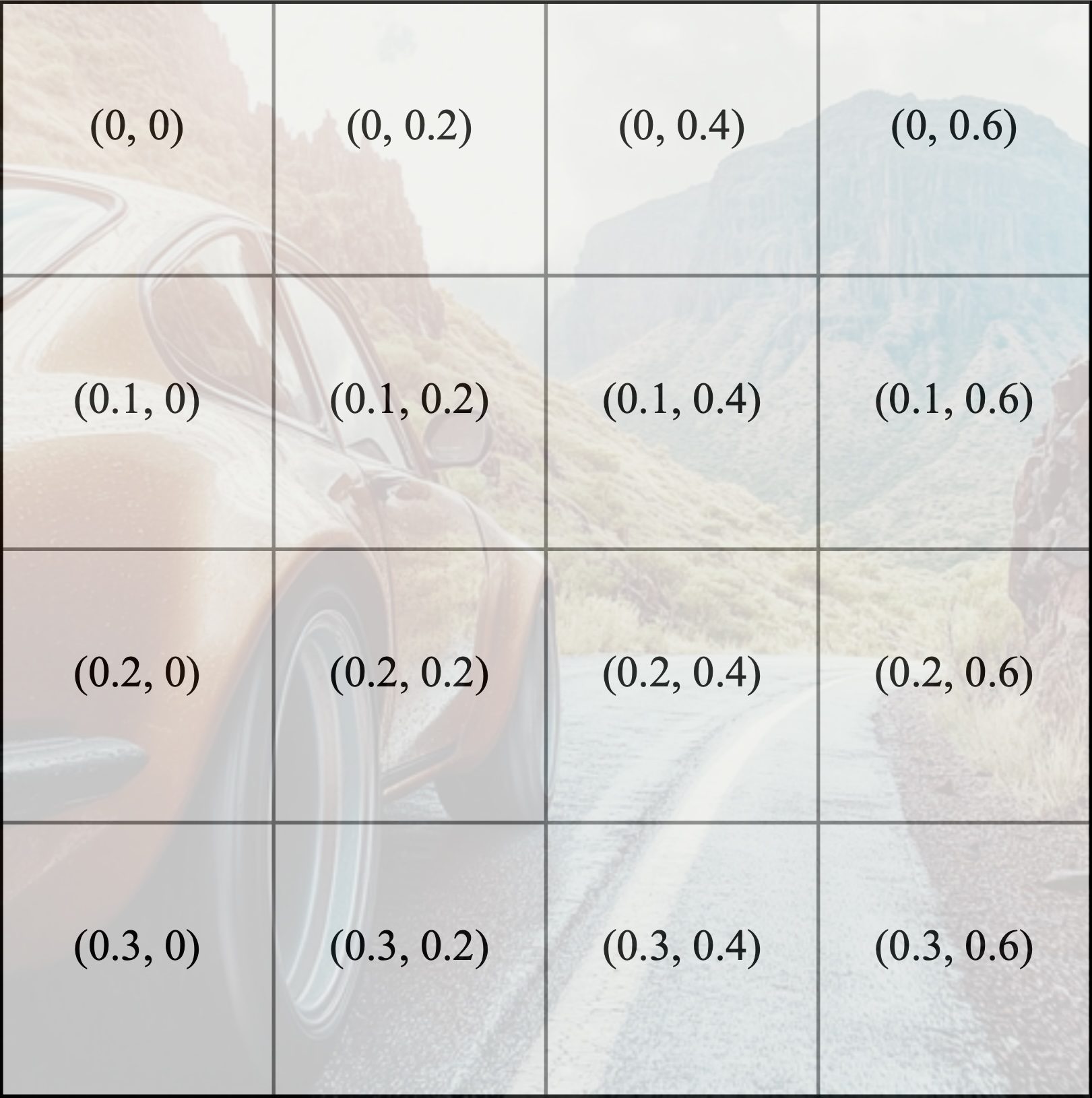}
        \caption{}
        \label{fig:pos_embs_rel}
    \end{subfigure}
    \caption{Positional encoding options: (a) Absolute positional encoding. (b) Fractional positional encoding. (c) Relative fractional positional encoding. Our experiments showed that relative-fractional positional embedding (option c) works best.}
    \label{fig:pos_embs}
\end{figure}

In our positional embedding experiments, we tested three primary variants: (1) absolute positional embeddings, (2) RoPE with fractional coordinates, and (3) RoPE with fractional coordinates normalized by predefined maximum coordinates. See Fig~\ref{fig:pos_embs}. Our empirical results demonstrated superior performance with the normalized fractional coordinate approach (Fig~\ref{fig:pos_embs_rel}). To ensure consistency over different resolutions, number of frames and frame rate, we compute the spatial and temporal RoPE coordinates in pixels and seconds respectively, relative to predefined max resolution and duration. Incorporating the original FPS into the temporal embedding allows the model to generate more natural-looking motion.

\begin{figure}[htbp]
    \centering
    \includegraphics[width=0.9\textwidth]{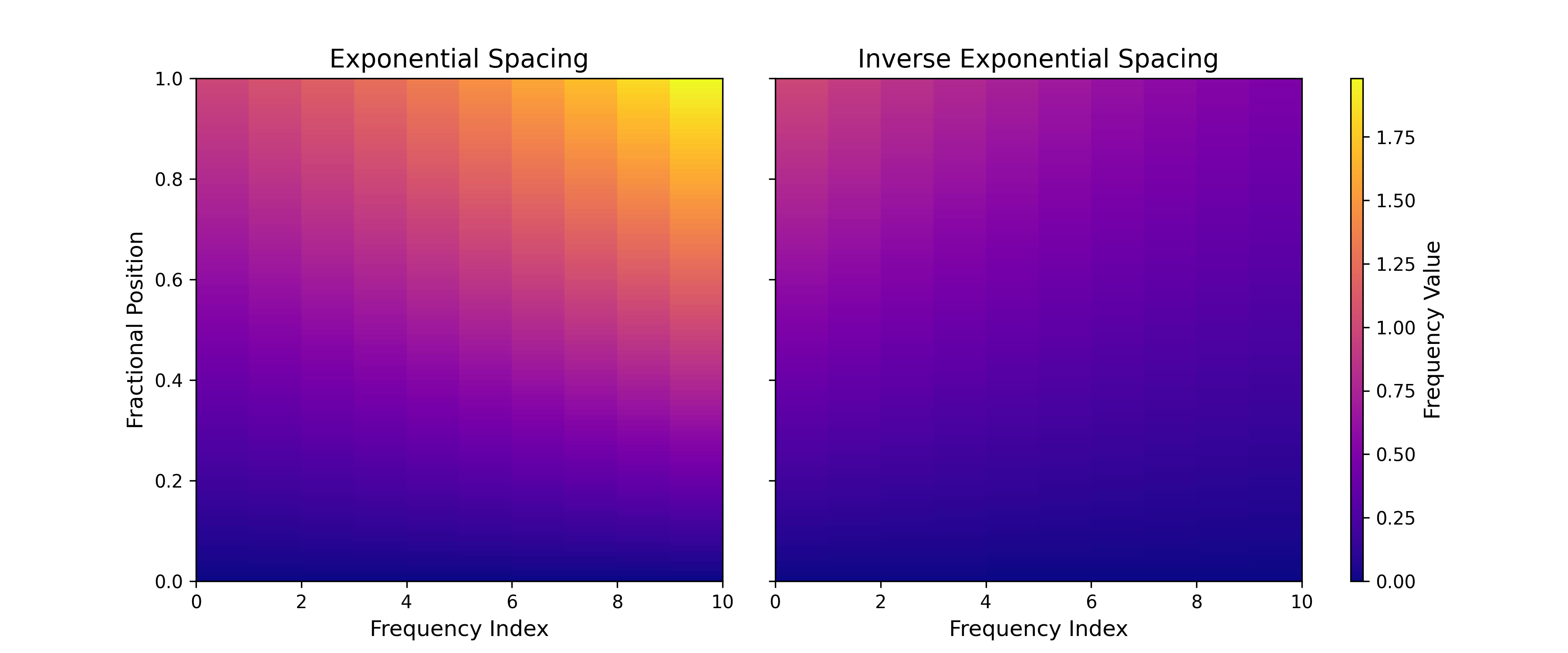}
    \caption{Different options for RoPE frequency spacing -- exponential (left) and inverse-exponential (right). \textit{LTX-Video} uses exponential spacing. See also section~\ref{sec:ablation-rope}.}
    \label{fig:rope}
\end{figure}

While many open-source implementations use inverse exponential frequency spacing, our experiments, including controlled toy problems, revealed better performance with exponentially increasing frequencies (Fig~\ref{fig:rope}). This finding aligns with recent theoretical work \citep{barbero2024round} suggesting that truncating lower frequencies can improve model performance.

\subsubsection{QK Normalization}

Following the findings presented in \citeme{chen2022scaling} and \citeme{largedit}, we apply a normalization layer to the queries and keys before the dot product attention computation to avoid extremely large values in attention logits, which lead to attention weights with near-zero entropy.

We compared RMSNorm and \citeme{lei2016layer} and found that RMSNorm performs better.

\vspace{2ex}
These enhancements are designed to significantly improve the model's performance and adaptability, particularly in addressing the unique challenges presented by video and image generation tasks. By integrating these advanced features, our model achieves a higher degree of accuracy and efficiency in synthesizing high-quality visual content.

\vspace{1ex}
See Table~\ref{tab:model_comparison} for a comparison of the \textit{LTX-Video} architecture to recent text-to-video models - \citeme{polyak2024movie}, \citeme{kong2024hunyuanvideo}, \citeme{jin2024pyramidal} and \citeme{yang2024cogvideox}.       

\begin{table}[htbp]
    \centering
    \scriptsize
    \renewcommand{\arraystretch}{1.4}
    \begin{tabular}{@{}lccccc@{}}
        \toprule
                                             & \textbf{MovieGen} & \textbf{HunyuanVideo} & \textbf{PyramidFlow}  & \textbf{CogVideoX} & \textbf{LTX-Video} \\
        \midrule
        Base model size                      & 30B               & 13B                   & 2B                  & 2B / 5B             & 1.9B          \\
        Transformer hidden dimension         & 6144              & 3072                   & 1536                & 1920 / 3072         & 2048          \\
        FFN dimension factor                 & $2/3 \times 4$    & 4                   & 4                   & 4                   & 4             \\
        VAE spatio-temporal compression      & $8 \times 8 \times 8$ & $8 \times 8 \times 4$ & $8 \times 8 \times 8$ & $8 \times 8 \times 4$ & $32 \times 32 \times 8$ \\
        VAE output channels                  & 16                & 16                   & 16                  & 16                  & 128           \\
        VAE total compression                & 1:96              & 1:48                   & 1:96                & 1:48                & 1:192         \\
        Transformer input patchifier         & $2 \times 2 \times 1$ & $2 \times 2 \times 1$                   & $2 \times 2 \times 1$ & $2 \times 2 \times 1$ & $1 \times 1 \times 1$ \\
        Tokens to pixels ratio               & 1:2048            & 1:1024                   & 1:2048              & 1:1024                & 1:8192       \\
        Number of Transformer blocks         & 48                & 60                   & 24                  & 30 / 42             & 28            \\
        Attention Blocks Architecture        & Self + Cross  & Self-only   & Self-only     & Self-only      & Self + Cross \\
        Upsampling model                     & 7B params         & N/A                   & N/A                 & N/A                 & N/A           \\
        \bottomrule
    \end{tabular}
    \caption{Comparison of Model Specifications.}
    \label{tab:model_comparison}
\end{table}

\subsection{Text Conditioning}
The choice of a text conditioning method plays a critical role in ensuring that the model accurately interprets and generates content based on textual input. To achieve robust text-to-image and text-to-video synthesis, we employ several strategies inspired by recent advancements in the field.

\textbf{Utilizing Pre-Trained Text Encoders}
\citeme{nichol2021glide} pioneered the learning of a transformer-based text encoder integrated with a denoising U-Net, setting a foundational approach for subsequent models.
\citeme{saharia2022photorealistic} and \citeme{ho2022imagen} recommended using a pre-trained text encoder to condition diffusion models, a practice that has been widely adopted in subsequent works, including the use of a learnable projection layer on the input text embeddings to refine feature mapping.

Consistent with the approaches in \citeme{saharia2022photorealistic}, \citeme{betker2023improving}, and \citeme{chen2023pixart}, we utilize the \citeme{raffel2020exploring} text encoder to generate initial text embeddings. This choice is motivated by the success of these models in leveraging pre-trained text encoders to enhance the semantic understanding of the input text.

\textbf{Cross-attention} The two common architectures for conditioning diffusion transformers on text embedding tokens are \textit{cross attention} (\citeme{chen2023pixart}, \citeme{polyak2024movie}) and MM-DiT (introduced by \citeme{esser2024scaling} and used by \citeme{flux}, \citeme{auraflow}, \citeme{yang2024cogvideox}). In MM-DiT, text embeddings are processed in parallel to image patch embeddings. Unified attention layers facilitate the exchange of information between these two types of embeddings, suggesting that part of the transformation applied to image embeddings could be decoupled from the noise and timestep parameters. We use cross attention as we found it to work better than MM-DiT.

\subsection{Image Conditioning}
In many content-creation workflows, video generation starts from a given first frame, either real or generated, which we wish to animate and extend based on the text prompt. Different approaches exist for injecting the image-conditioning signal into the model (\citeme{blattmann2023stable}, \citeme{ni2023conditional}, \citeme{zhang2023i2vgen}, \citeme{opensoraplan}), typically requiring special tokens and models trained specifically for the image-to-video task. We base on and extend the approach presented in \citeme{opensora}, which leverages the diffusion timestep as a conditioning indicator, allowing seamless conditioning on any part of the video.

In \citeme{peebles2023scalable} and \citeme{chen2023pixart}, the same timestep embedding is always injected to all tokens via AdaLN and all tokens are noised to the corresponding noise level. We relax this restriction by allowing a different timestep and corresponding noising level per token.

To train the model for first-frame conditioning, we occasionally set the timestep of the tokens belonging to the first video frame to a small random value and noise these tokens to the corresponding level. The model quickly learns to utilize this new information (when provided) as a conditioning signal. 

\begin{figure}[htbp]
    \centering
    \includegraphics[width=0.7\textwidth]{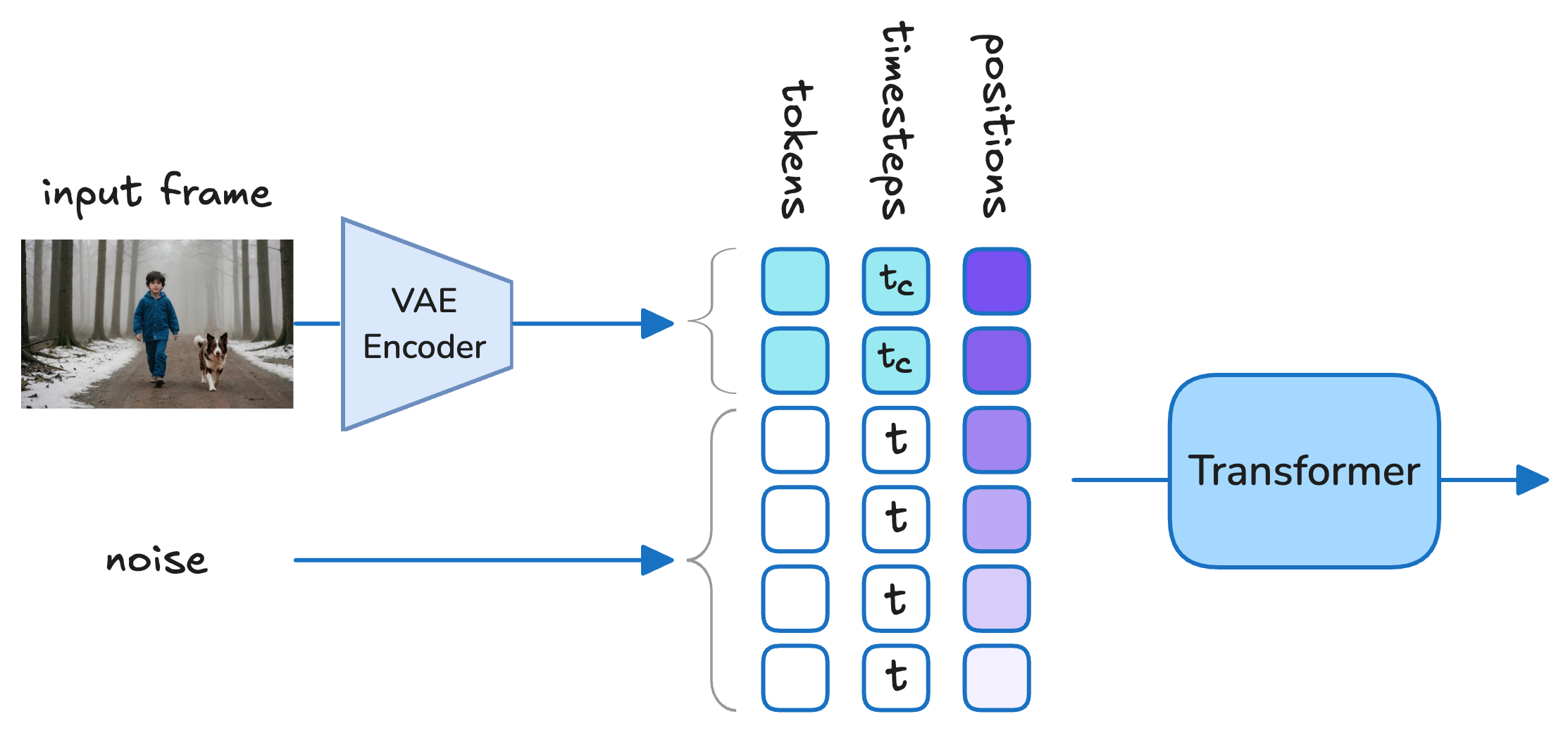}
    \caption{\textit{LTX-Video} image-to-video inference pipeline -- first-frame conditioning. The diffusion timestep and corresponding noise level is defined per-token. For example, conditioning tokens can have diffusion timesteps of $t_c=0$ and contain un-noised encoded tokens.}
    \label{fig:img2vid-inference}
\end{figure}

During inference, the conditioning image is encoded into a latent tensor with a temporal dimension of 1 using our causal VAE encoder. This tensor is concatenated with random noise latents and flattened to form the initial set of tokens. The per-token denoising timesteps are set to a small value $t_c$ for conditioning tokens and to $t=1$ for all other tokens. Figure~\ref{fig:img2vid-inference} illustrates this process.

\subsection{Rectified-Flow Training}

In this section, we describe specific design choices for the training procedure and the loss. These design choices significantly affect the training time and the resulting model quality.

\subsubsection{Background}

In \textit{Rectified Flow} (\cite{lipman2022flow}, \citeme{esser2024scaling}), the clean input $z_0$ is noised linearly in the forward process by the timestep $t \in [0, 1]$, according to $z_t = (1-t) z_0 + t \epsilon$, where the noise $\epsilon$ is sampled from the standard-normal distribution $\mathcal{N}(0, I)$.

During training, $t$ is randomly sampled from some distribution (uniform in the original diffusion paper -- $t\sim \mathcal{U}(0, 1)$). 
In the original diffusion setting, the model was tasked with predicting the noise $\epsilon$. Since this prediction task is not evenly-difficult, \citeme{esser2024scaling} propose to task the network with predicting the \textit{velocity} $v = \epsilon - z_0$ instead.

During inference, initial pure noise $z_1$ is gradually denoised towards a clean image $z_0$. 
At each step, the denoising process is $z_{t-\Delta t} = z_t - \Delta t \, v_t^\theta $, where $v_t^\theta$ is the velocity predicted by the model at timestep $t$.

\subsubsection{Timestep scheduling}

\citeme{esser2024scaling} proposed sampling the diffusion timestep $t$ during training from the \textit{log-normal} distribution, replacing the uniform distribution $\mathcal{U}(t)$ used in earlier models. The motivation is to assign more training steps to timesteps in which the velocity-prediction task is more difficult. \citeme{hoogeboom2023simple} shows that at higher image resolutions, a higher level of noising is required to maintain the SNR. We adopt this recommendation and shift the timestep scheduler towards the higher-noise regions, depending on the number of tokens. To prevent starvation at the tail of the resolution we clamp the \textit{pdf} at percentiles $0.5$ and $99.9$.

\begin{figure}[htbp]
    \centering
    \includegraphics[width=0.49\textwidth]{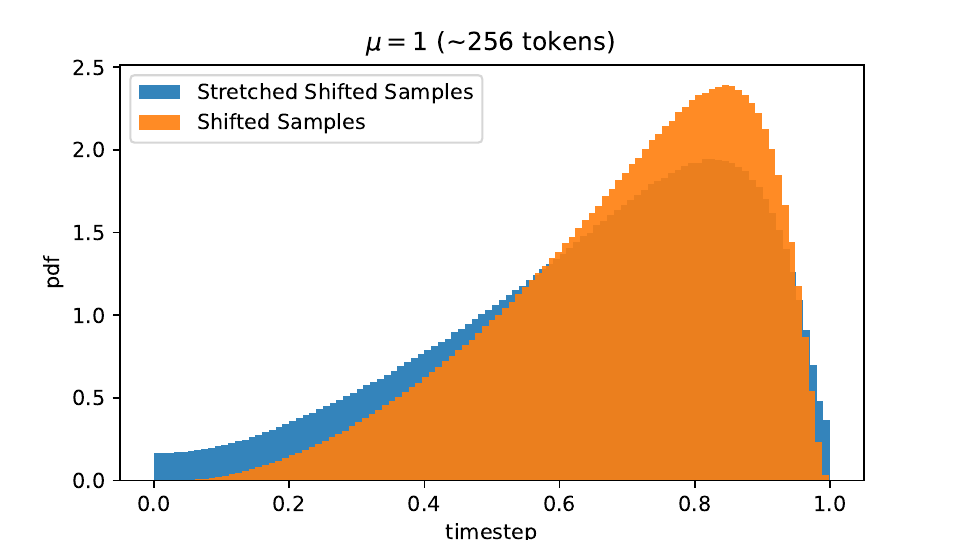}
    \includegraphics[width=0.49\textwidth]{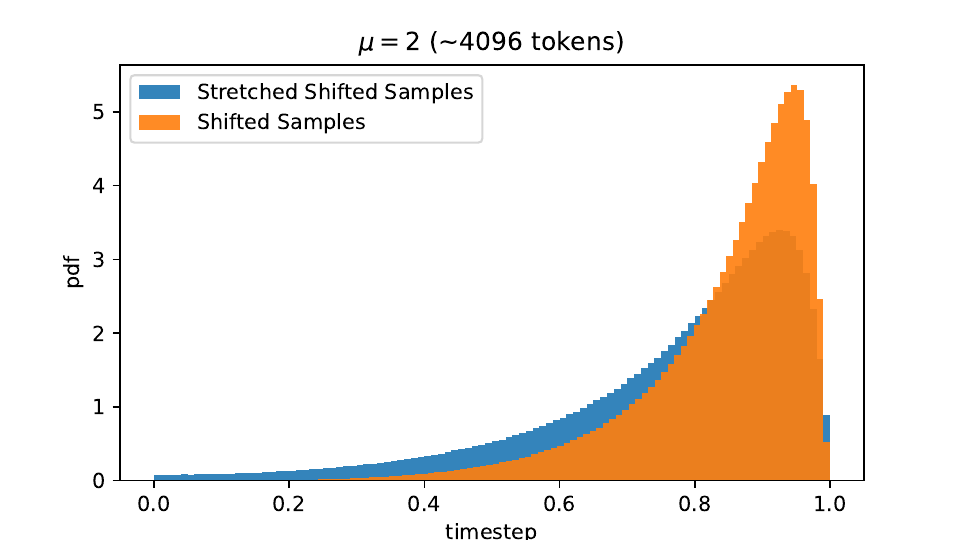}
    \caption{Timestep $t$ sampling distribution during training, shown with two shifting parameters $\mu$. We use the distributions shown in blue, which prevent near-zero probability at the tails.}
    \label{fig:scheduler}
\end{figure}

See Fig~\ref{fig:scheduler} for two timestep distribution shifting for two different resolutions.

\subsubsection{Multi-resolution Training}

To enable the model to generate videos at various resolutions and durations, we trained simultaneously on multiple combinations of resolutions and durations. We observed that after being exposed to a diverse range of width, height, and duration combinations, the model generalized well to unseen configurations.

During training, we ensured that all input samples contained approximately the same number of tokens by resizing the original videos to comparable token counts. To fix token counts across all sequences, we applied stochastic token dropping at rates ranging from 0\% to 20\%. This simple and efficient approach eliminated the need for complex token-packing or padding strategies while preserving diversity in the training data.

\subsubsection{Training on Images}

We include image training alongside regular video training, treating it as one of the resolution-duration combinations. Image datasets can enrich the set of concepts encountered during training, incorporating those not typically present in video datasets.

\section{Data Preparation}

\begin{figure}[htbp]
    \centering
    \includegraphics[trim=0.2in 0in 0.2in 0.5in, clip, scale=0.65]{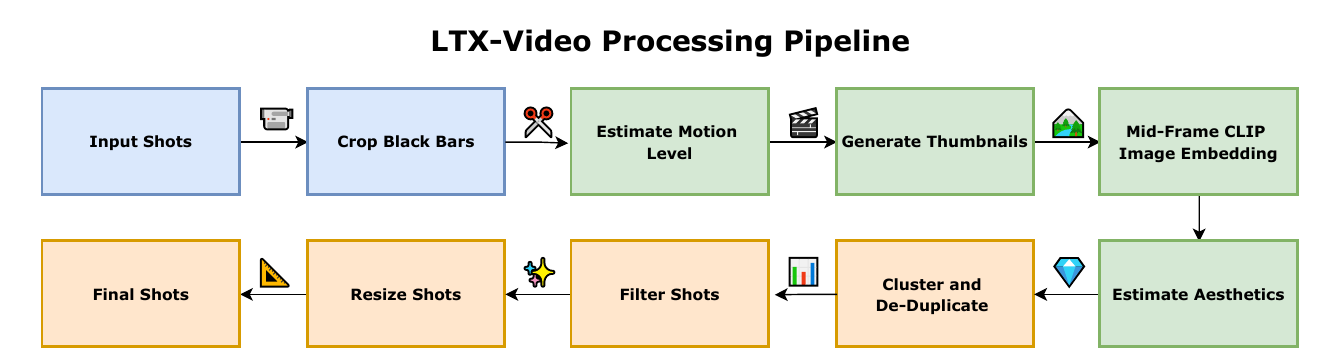}
    \caption{Our data processing pipeline.}
    \label{fig:data_pipeline}
\end{figure}

Our training dataset comprises a robust collection of publicly available data, supplemented with licensed material, ensuring a diverse and comprehensive training environment and enabling our model to generate a wide variety of visual content. This section describes our data processing pipeline -- see Fig~\ref{fig:data_pipeline}.

\textbf{Quality Control and Filtering}
Quality control is a critical aspect of our data preparation process. We train and employ an aesthetic model to evaluate both videos and images. The model was trained on tens of thousands of image pairs, manually tagged to identify the superior image in terms of aesthetic quality.

To sample pairs for manual tagging, we labeled millions of samples using a multi-labeling network and sampled only pairs that share at least one of the top three labels. This approach helped minimize distribution shifts when filtering data based on aesthetics.

The training data was used to train a \citeme{siamese} designed to predict an aesthetic score that preserves the order relations established by the tagged pairs. Once trained, the model computes an aesthetic score for each sample, and we filter out those with scores below a certain threshold.

This process ensures that we select only the most visually appealing content, which is crucial for training our models to generate high-quality outputs.

\textbf{Motion and Aspect Ratio Filtering}
In addition to aesthetic filtering, we actively remove videos that exhibit insignificant motion to ensure that our dataset focuses on dynamic content, which is more relevant to our model's target capabilities. Videos are also processed to crop out any black bars, thus standardizing the aspect ratios and enhancing the usable visual data.

\textbf{Fine-Tuning with Aesthetic Content}
For fine-tuning, we selectively use the most aesthetic content identified by our filtering processes. This approach helps in achieving more visually appealing results in the generated outputs, aligning with high standards of image and video quality.

\begin{figure}[htbp]
    \centering
    \begin{subfigure}[b]{0.48\textwidth}
        \centering
        \includegraphics[height=5.2cm]{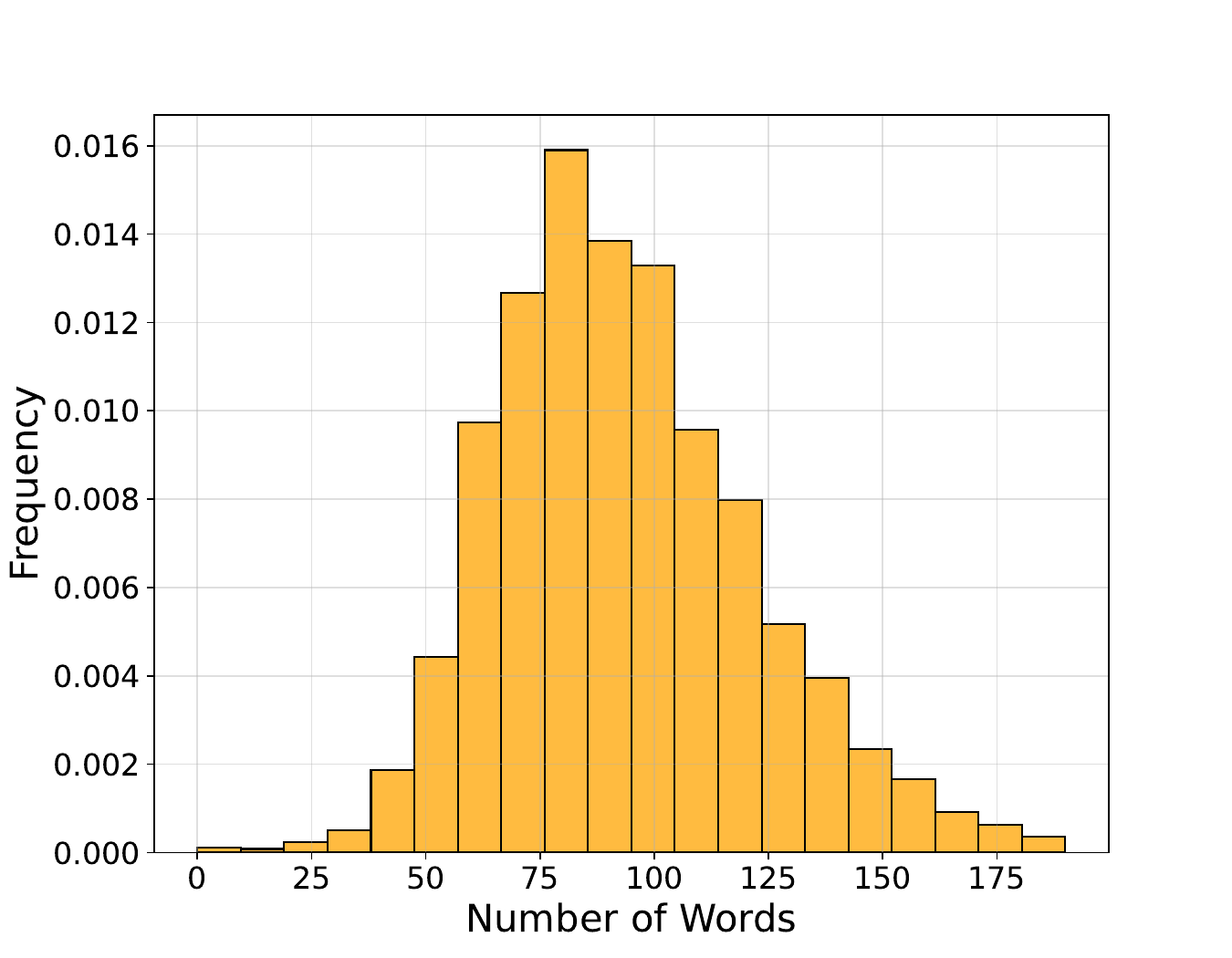}
        \caption{Number of words per caption}
        \label{fig:caption_hist}
    \end{subfigure}
    \hspace{3mm}
    \begin{subfigure}[b]{0.48\textwidth}
        \centering
        \includegraphics[height=5.2cm]{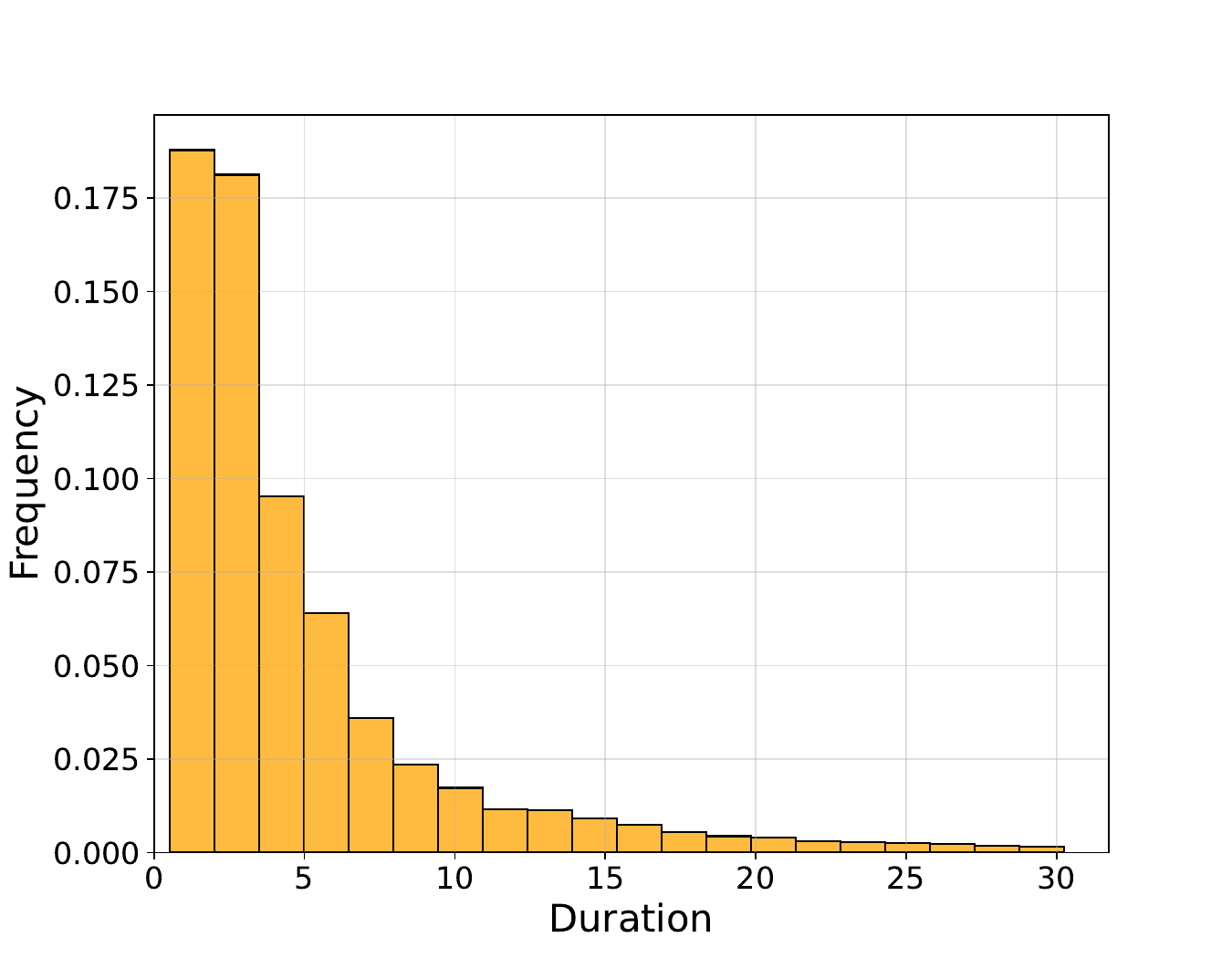}
        \caption{Clip duration (seconds)}
        \label{fig:duration_hist}
    \end{subfigure}
    \caption{Training data statistics.}
\end{figure}

See Fig~\ref{fig:duration_hist} for the distribution of the clip durations in our filtered data.

\begin{figure}[htbp]
\centering
\scriptsize
\setlength\tabcolsep{1pt} 
\renewcommand\arraystretch{1.0} 

\begin{tabular}{@{}cccc@{}}
    \includegraphics[width=0.24\textwidth]{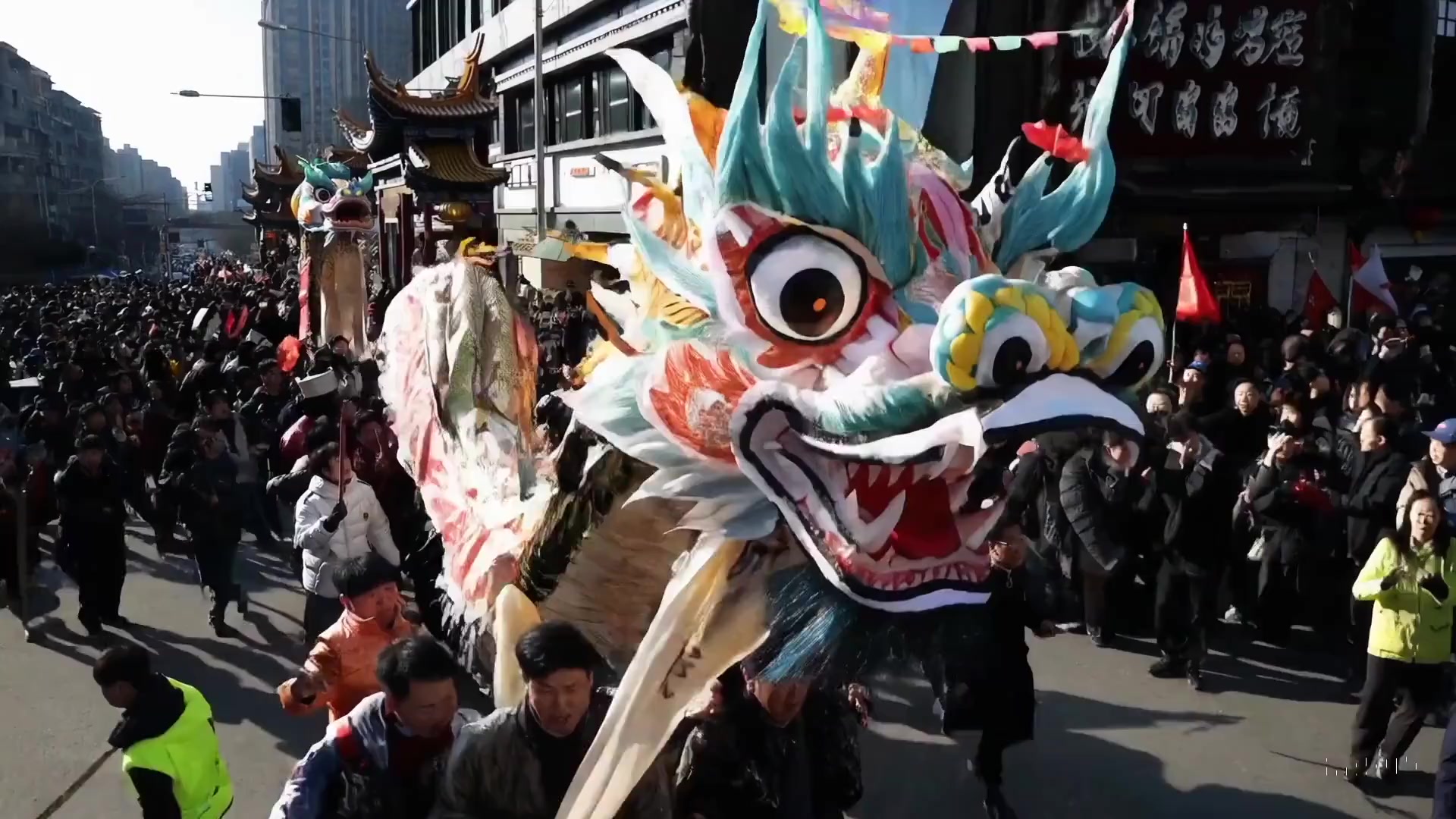} & \includegraphics[width=0.24\textwidth]{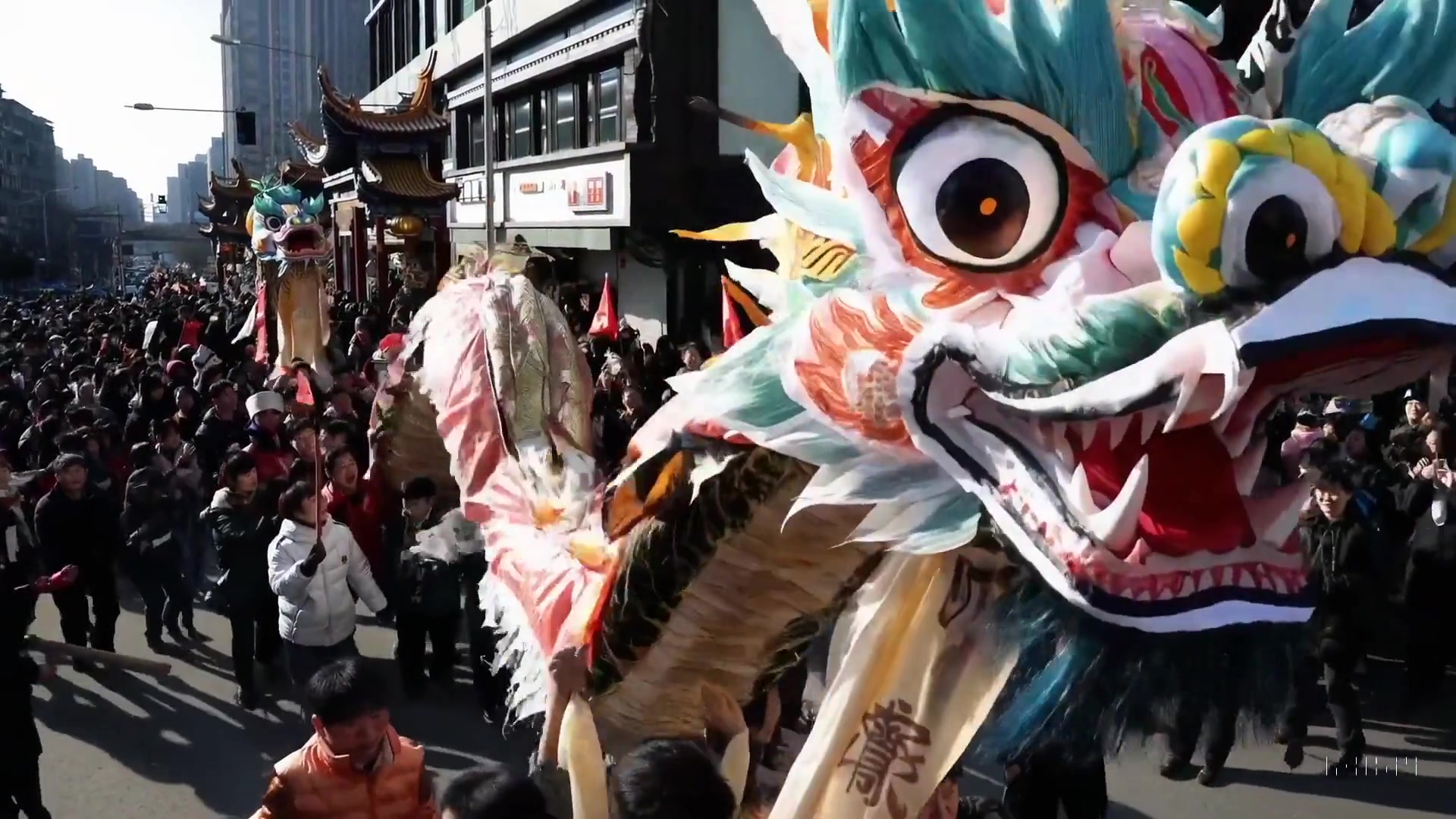} & \includegraphics[width=0.24\textwidth]{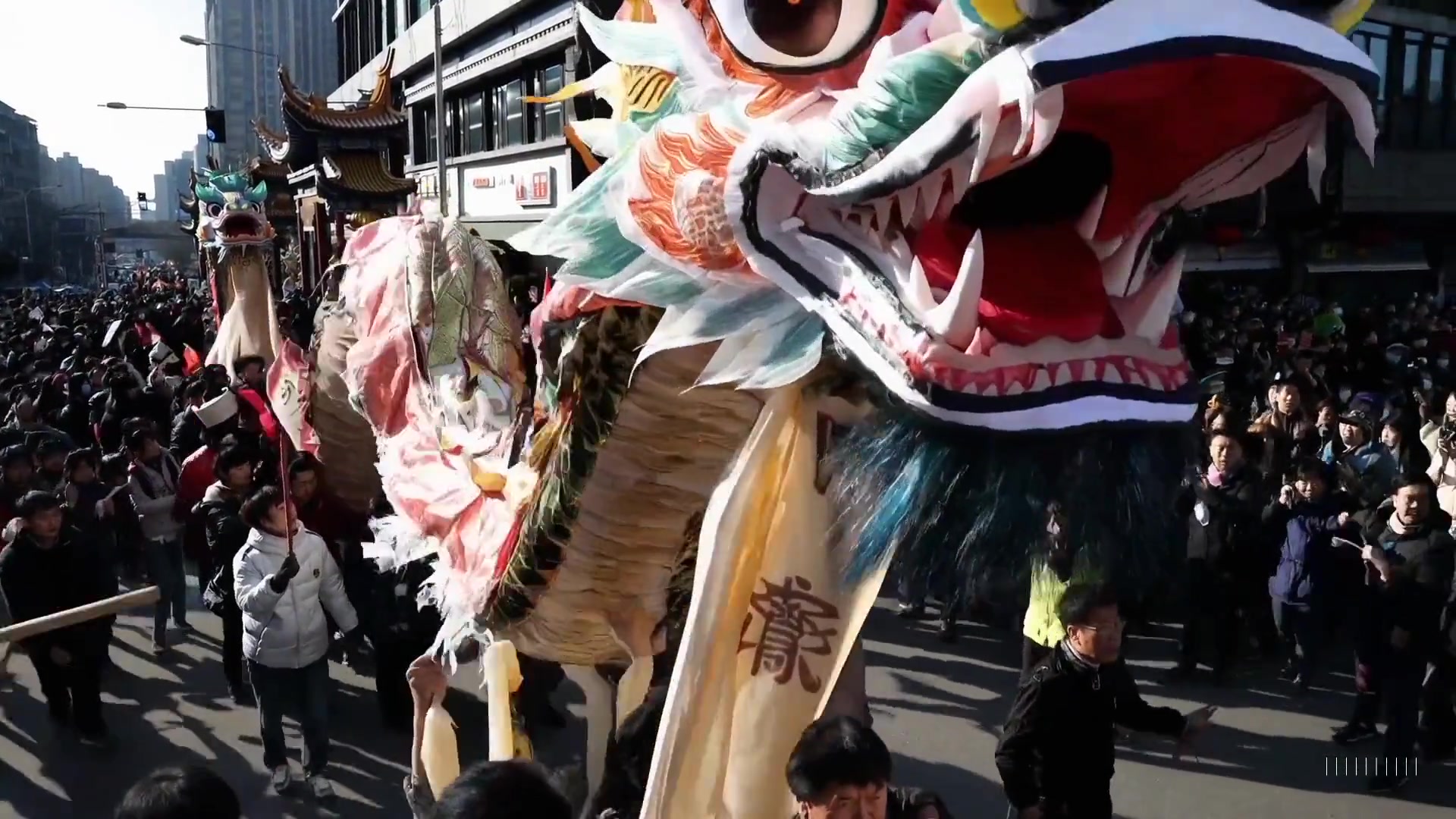} & \includegraphics[width=0.24\textwidth]{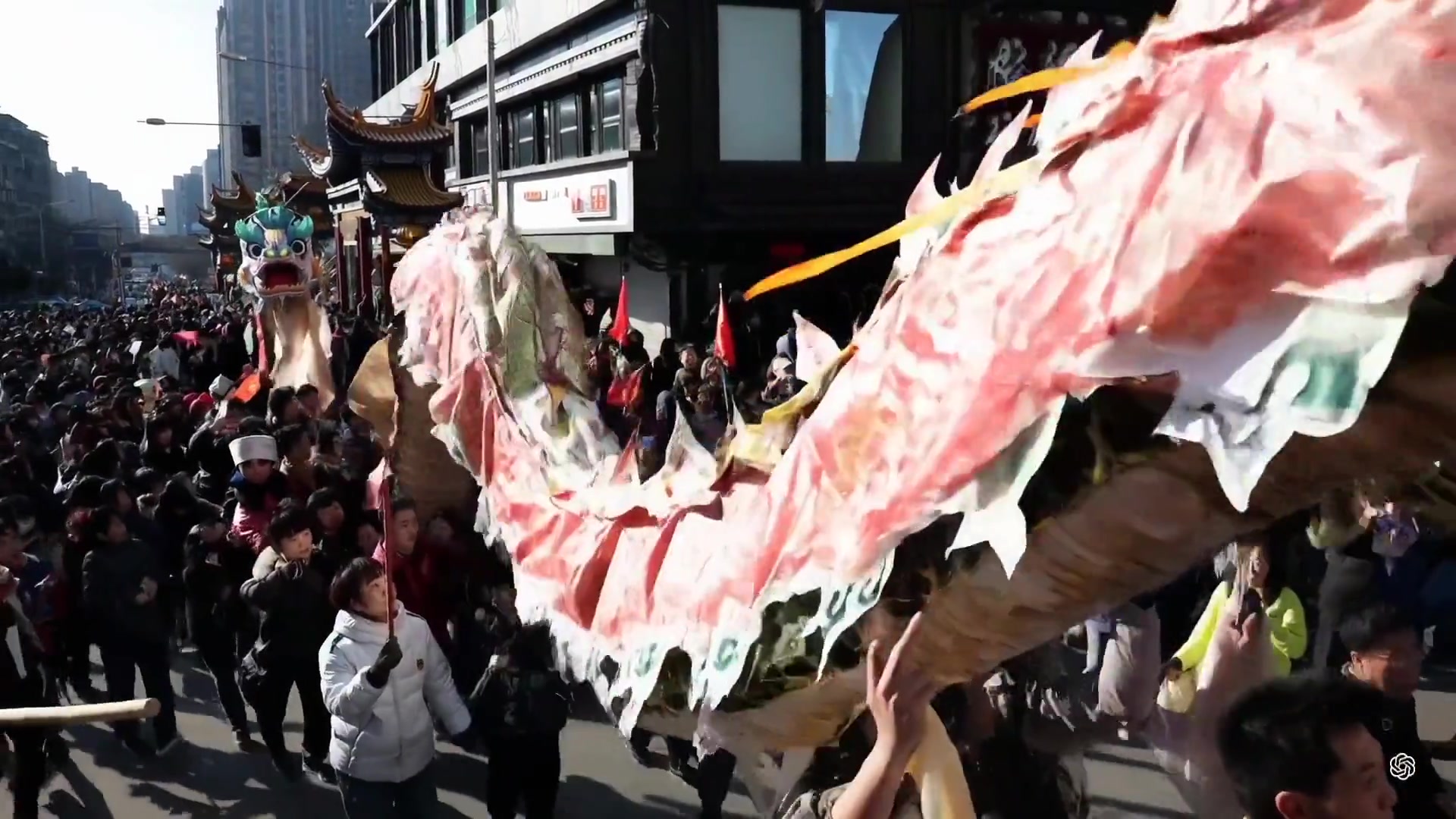} \\
\end{tabular}

A large, colorful dragon puppet is carried down a street by a crowd of people. The dragon puppet is predominantly white with blue, green, and red accents, and it has large, expressive eyes and a long, serpentine body; it is suspended from a pole held by a person in the crowd, and its head moves slightly as it is moved down the street. The crowd is composed of people of all ages, many of whom are wearing winter clothes; they are cheering and waving their arms in the air, and some are taking pictures or videos with their phones. The street is lined with buildings, some of which have signs in Chinese characters; there are also a few trees and streetlights visible. The camera follows the dragon puppet from behind as it moves down the street, and it pans slightly to capture the surrounding crowd and the buildings; the lighting is natural and slightly overcast, casting soft shadows on the street. The scene is captured in real-life footage.

\vspace{0.2cm} 

\begin{tabular}{@{}cccc@{}}
    \includegraphics[width=0.24\textwidth]{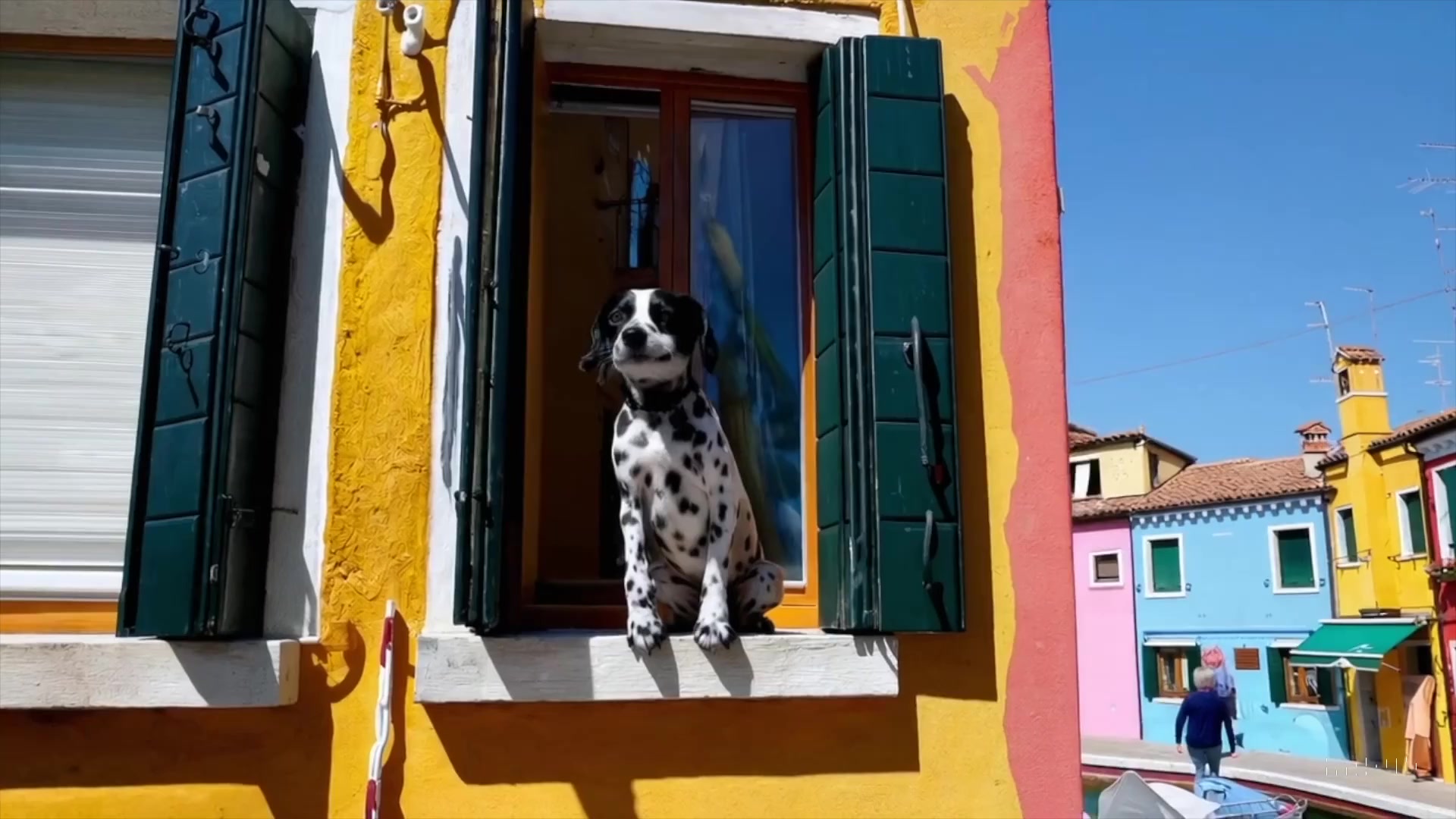} &
    \includegraphics[width=0.24\textwidth]{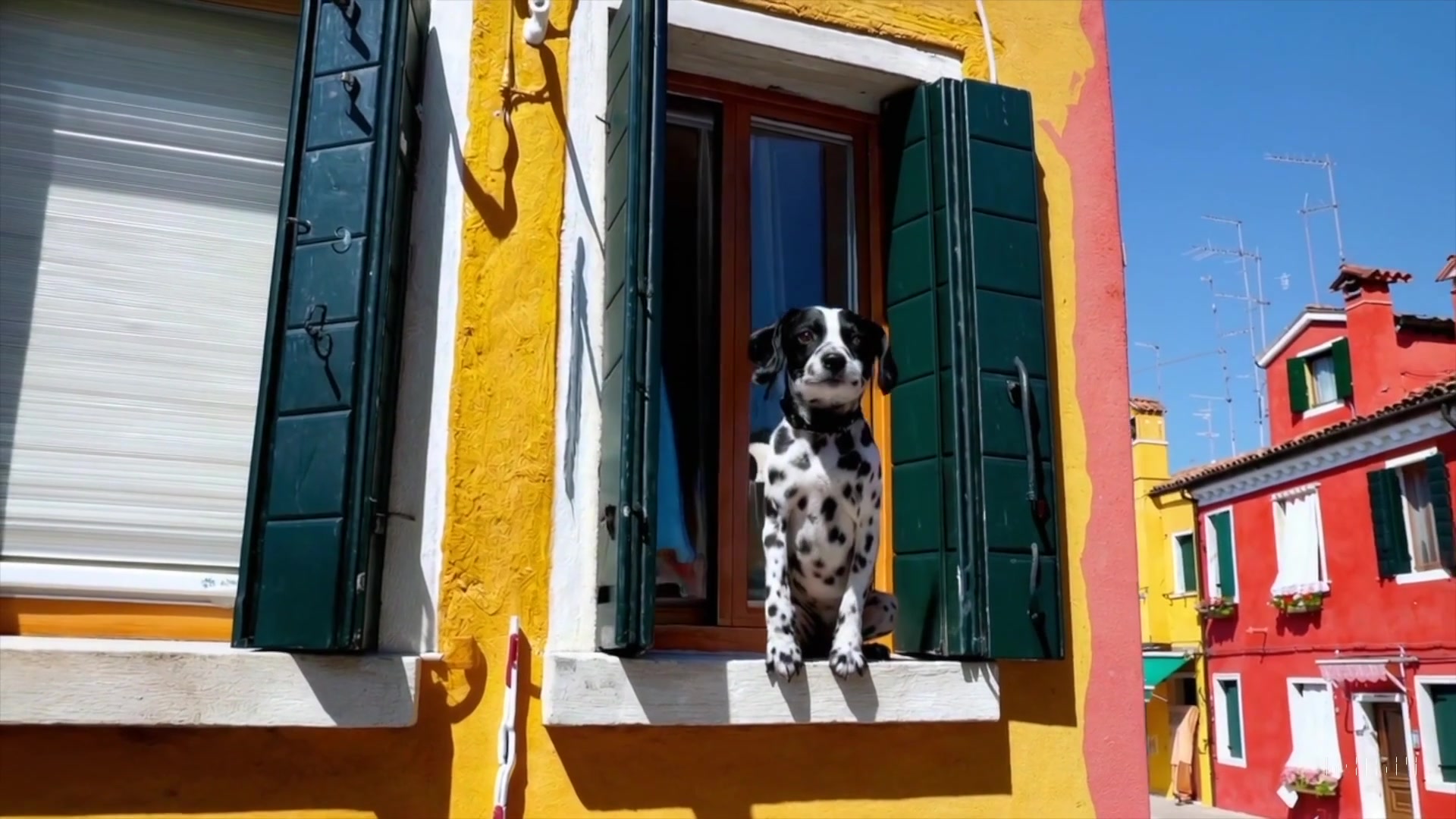} &
    \includegraphics[width=0.24\textwidth]{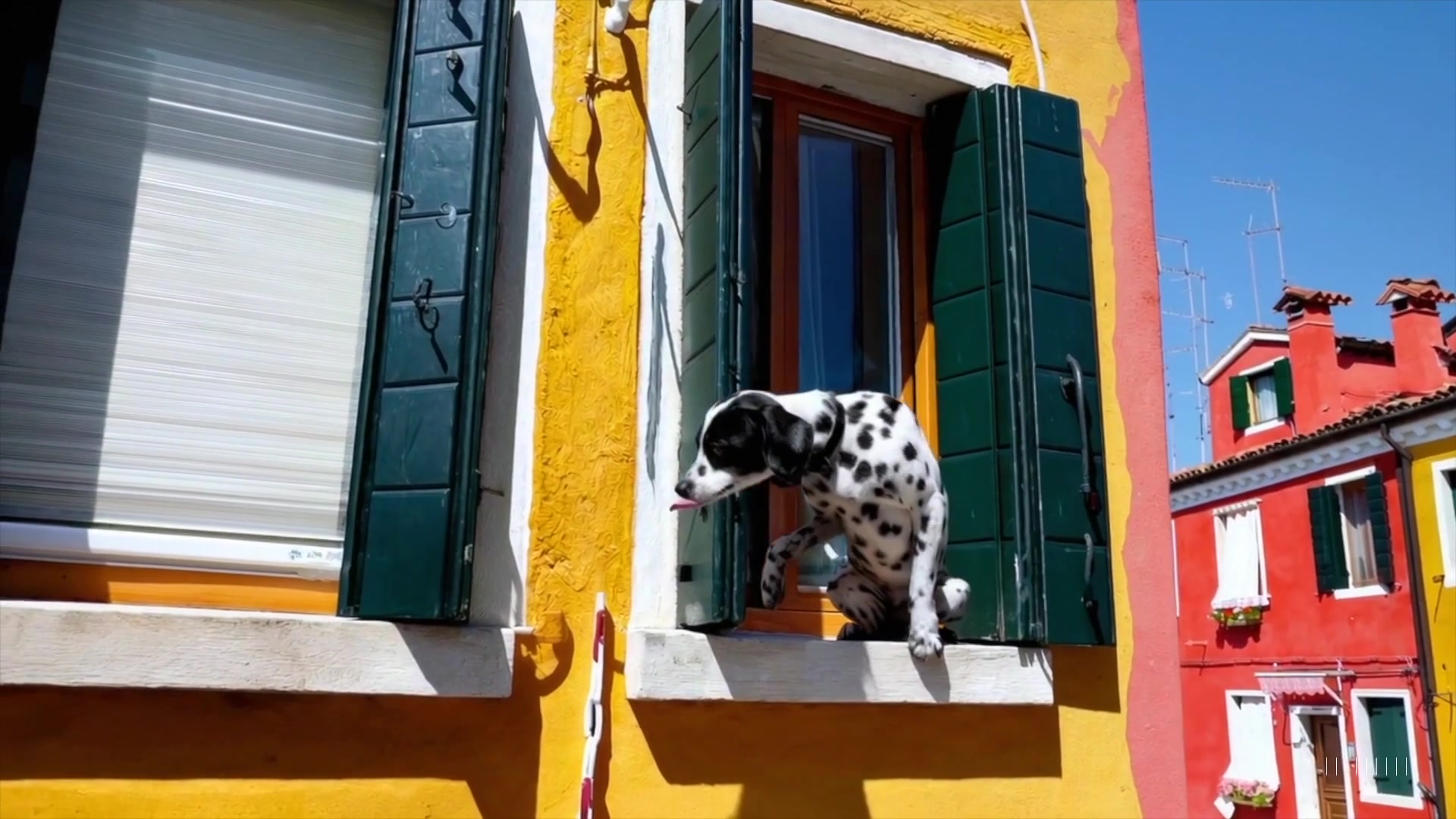} &
    \includegraphics[width=0.24\textwidth]{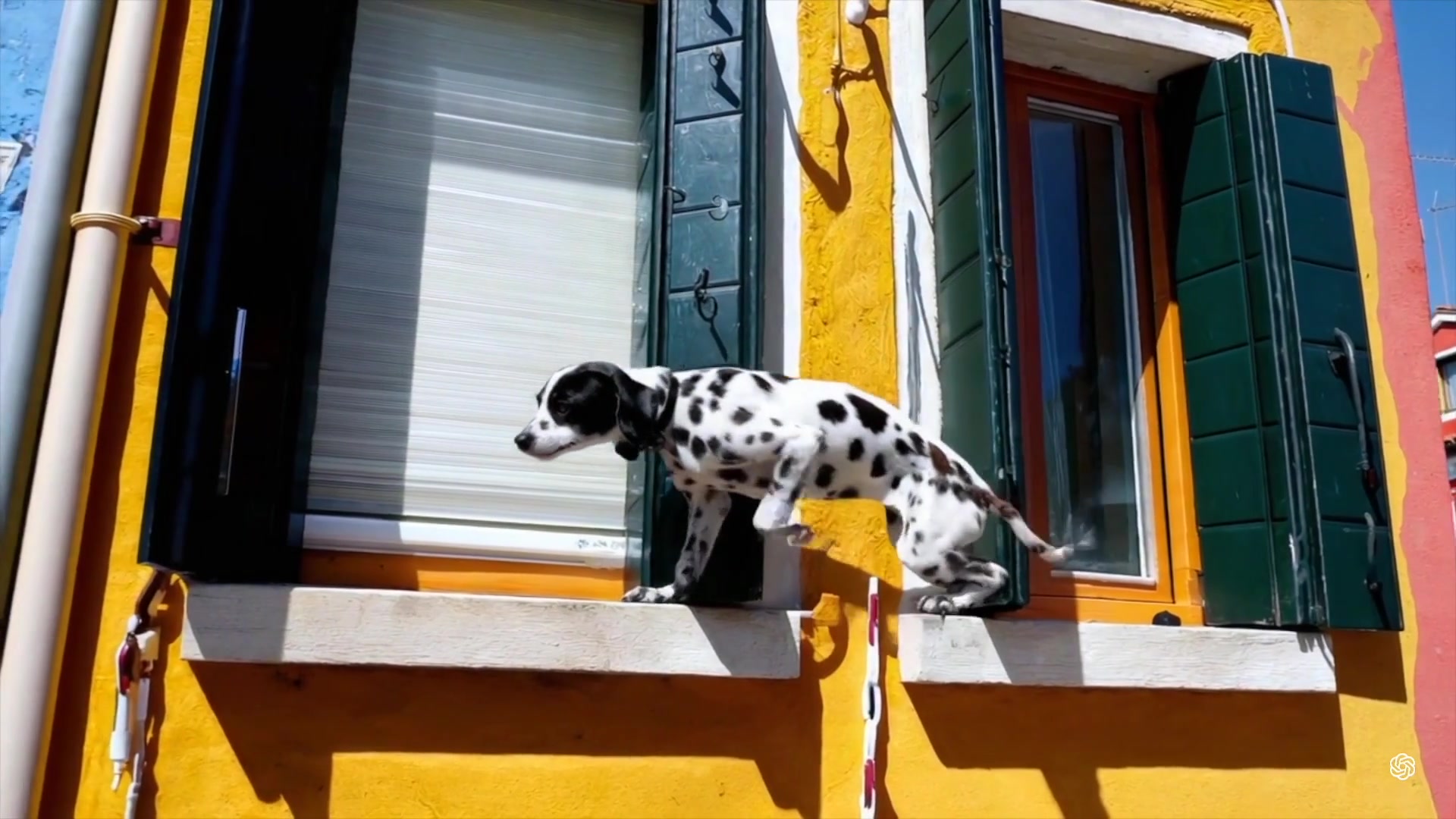} \\
\end{tabular}

A Dalmatian dog stands on a window ledge, looks around, and then jumps. The dog has black spots on its white fur and is standing on a window ledge of a yellow building with green shutters. The dog looks to the left and then jumps from the window ledge. The camera is positioned across the street from the dog and remains stationary throughout the video. The lighting is natural sunlight, and the colors in the video are bright and vibrant. The scene is captured in real-life footage.

\caption{Our video captioner -- input video (generated by \citeme{brooks2024video}) and our generated captions.}
\label{fig:captioner_results}
\end{figure}

\begin{figure}[htbp]
    \centering
    \includegraphics[width=\textwidth]{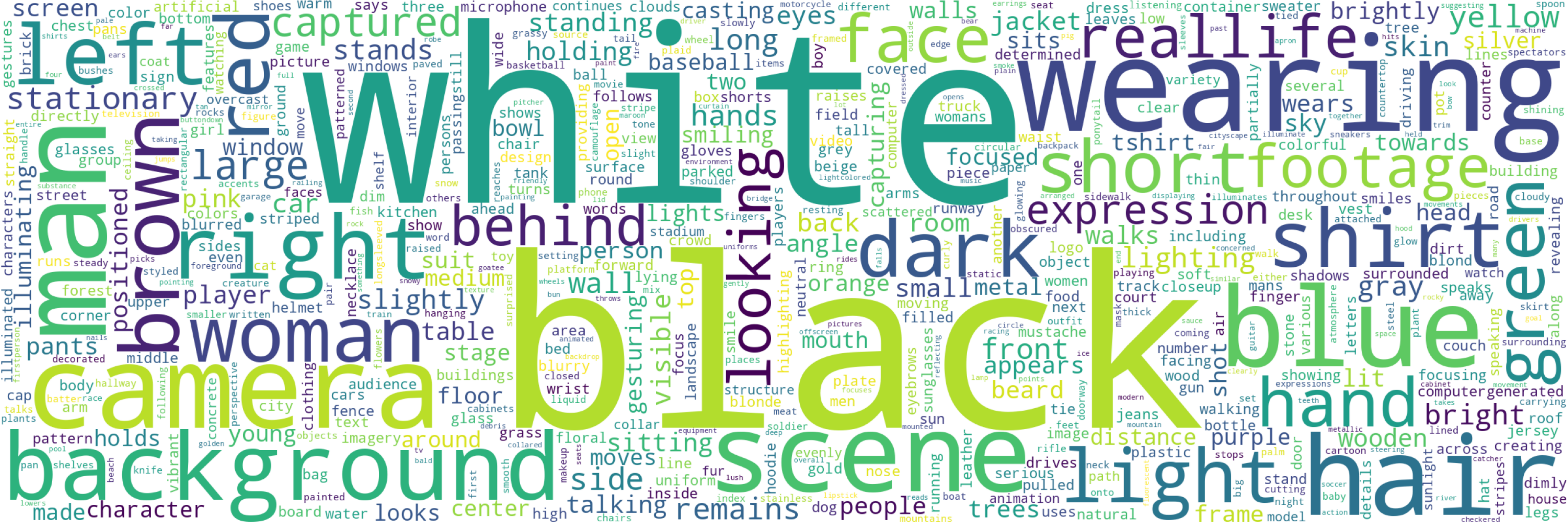}
    \caption{A word cloud of our captioned training data.}
    \label{fig:wordcloud}
\end{figure}

\textbf{Captioning and Metadata Enhancement}
To enhance the metadata of our training data, we utilize an internal automatic image and video captioner to re-caption the entire training set. This re-captioning process ensures that the text descriptions are accurate and relevant, providing better context for the training models and improving the alignment between visual content and textual annotations. Examples of input videos and generated captions are shown in Fig~\ref{fig:captioner_results}. Fig~\ref{fig:wordcloud} shows a word-cloud visualizing the distribution of caption words in our entire data, and the distribution of the number of words per captions is shown in Fig~\ref{fig:caption_hist}.

\section{Experiments}

\subsection{Training}
We trained our model using the \textit{ADAM-W} optimizerAfter pre-training, we fine-tuned our model on a subset of the data containing high-aesthetic videos.

\subsection{Evaluation}

Following \citeme{polyak2024movie}, we conducted a human survey to evaluate the quality of \textit{LTX-Video} compared to current state-of-the-art models of similar size: \citeme{opensoraplan}, \citeme{yang2024cogvideox} (2B) and \citeme{jin2024pyramidal}.
We used 1,000 prompts for the text-to-video task and 1,000 pairs of images and prompts for the image-to-video task (the images were generated using \citeme{flux}). We then generated 5-second videos at resolution of $768 \times 512$ using all evaluated models with their default configurations. All videos were generated using 40 diffusion steps.

The survey involved 20 participants, each of whom was shown each time a randomly ordered pair of videos from a randomly selected pair of evaluated models (not necessarily including \textit{LTX-Video}). The model names were not revealed to the participants. For each pair, the videos were generated using the same prompt (and the same initial frame for the image-to-video task). Participants ranked the videos based on their overall preference, considering visual quality, motion fidelity, and prompt adherence. Participants could skip voting on a pair if they were unsure, using a provided 'skip' option. To ensure that our survey included a sufficient number of evaluators, we divided the participants into two groups of 10 and found that the win rates between groups differed by less than 2\%.

\begin{figure}[htbp]
    \centering
    \begin{subfigure}[b]{0.4\textwidth}
        \centering
        \includegraphics[width=\textwidth]{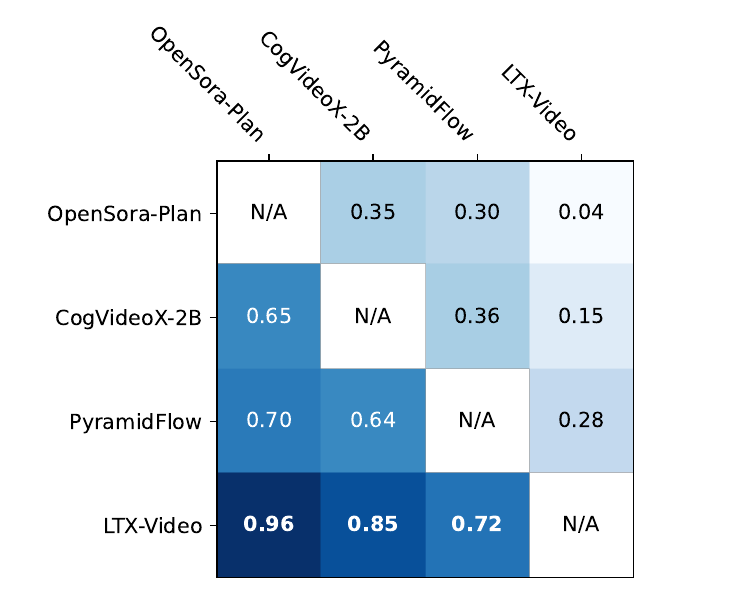}
        \caption{Text-to-Video}
    \end{subfigure}
    \hspace{3mm}
    \begin{subfigure}[b]{0.4\textwidth}
        \centering
        \includegraphics[width=\textwidth]{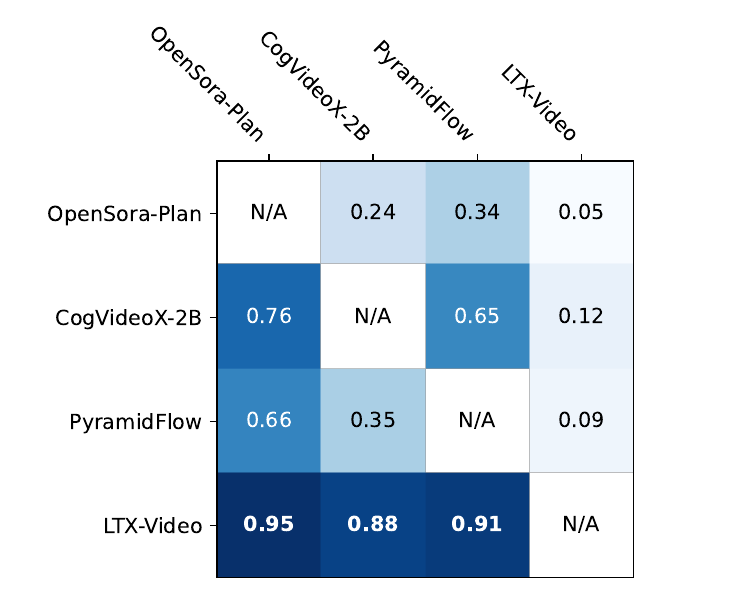}
        \caption{Image-to-Video}
    \end{subfigure}
    \caption{Pairwise performance matrix: each row indicates the win ratios of a model against all other models.}
    \label{fig:pairwise_wins}
\end{figure}

The results for each category were summarized as the percentage of tests in which each model won (calculated as $wins / (wins + losses))$. The survey results are presented in Table~\ref{tab:survey}. As shown, \textit{LTX-Video} significantly outperforms other similar-sized models, even with its considerable speed advantage. Figure~\ref{fig:pairwise_wins} illustrates the pairwise win ratios. Since we evaluated all model pairs (not just \textit{LTX-Video} against others), we present the full pairwise results.

\begin{table}[htbp]
\centering
\small
\caption{Survey results showing the percentage of tests in which each model won for text-to-video and image-to-video tasks.}
\label{tab:survey}
\begin{tabular}{lcccc}
\toprule
 & \textbf{Open-Sora Plan} & \textbf{CogVideoX 2B} & \textbf{PyramidFlow} & \textbf{\textit{LTX-Video}} (Ours) \\
\midrule
Text-to-video   & 20\% & 38\% & 51\% & \textbf{85\%} \\
Image-to-video  & 20\% & 47\% & 35\% & \textbf{91\%} \\
\bottomrule
\end{tabular}
\end{table}

\subsection{Ablations}
\label{sec:ablations}

\subsubsection{Reconstruction GAN vs. traditional GAN}

\begin{figure}[htbp]
    \centering
    \includegraphics[width=\textwidth]{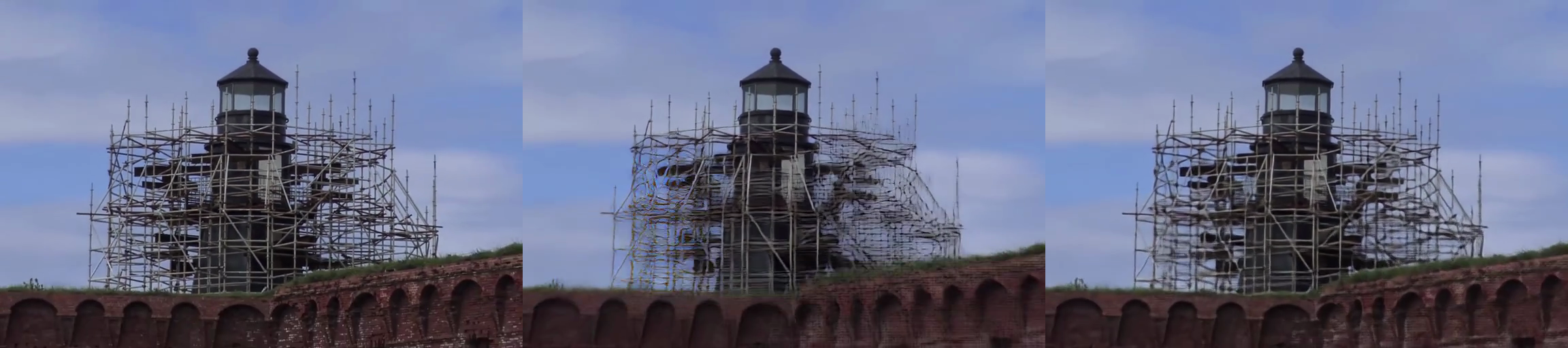}
    \caption{VAE reconstruction example -- input frame (left), reconstruction with standard GAN loss (center), improved reconstruction with our proposed \textit{Reconstruction GAN} loss (right)}
    \label{fig:vae_gan_compare}
\end{figure}

The high compression rate employed by our VAE challenges the reconstruction of high-frequency details, particularly in video frames that combine fast motion with intricate detail. Figure~\ref{fig:vae_gan_compare} illustrates such an example. At a high compression rate of 1:192, the commonly used combination of reconstruction and GAN losses fails to achieve consistent reconstruction of the original video (in this example, artifacts are visible in a single frame, but more commonly, inconsistencies become evident only when the video is played). Our proposed \textit{Reconstruction GAN} loss significantly reduces these visible artifacts.

\subsubsection{RoPE frequency spacing}
\label{sec:ablation-rope}
\begin{figure}[htbp]
    \centering
    \includegraphics[width=0.8\textwidth]{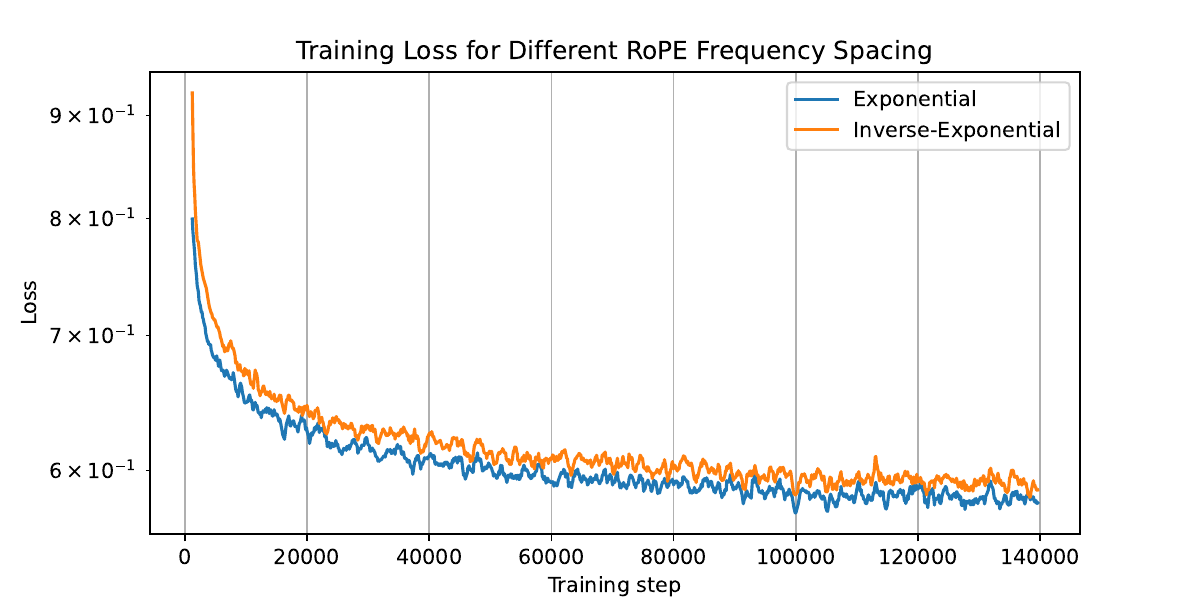}
    \caption{Video diffusion transformer training with two RoPE frequency spacing options - exponential and inverse-exponential. The loss is consistently better with exponential spacing.}
    \label{fig:rope_compare}
\end{figure}

As discussed in Sec~\ref{sec:rope}, we compared two variants for the RoPE frequency spacing -- exponential and inverse-exponential. Fig~{\ref{fig:rope_compare}} shows the training loss of two diffusion training experiments with the two frequency spacing options. The experiments were conducted using the same architecture, hyper-parameters and data that was used in our \textit{LTX-Video} training. As can be seen, the loss with inverse-exponential spacing remains higher.

\subsection{Denoising VAE Decoder}

Our holistic approach tasks the VAE decoder with performing the last denoising step in conjunction with converting the latents into pixels. To validate this design choice, we performed an internal user study comparing videos generated according to our approach to videos generated with the common approach, where denoising is performed solely by the diffusion-transformer, in latent space. 

For the first set of results, our VAE decoder was conditioned on timestep $t=0.05$. For the second set, the VAE decoder was conditioned on timestep $t=0.0$ and did not perform any denoising.

The survey results indicated that videos generated by our method were strongly preferred over the standard results. 
The improvement was particularly evident in high-motion videos, where artifacts caused by the strong compression were mitigated by the VAE decoder's last-step denoising.

\section{Limitations}

\textbf{Model Sensitivity to Prompt Formulation}
While \textit{LTX-Video} demonstrates strong prompt adherence, its performance can vary significantly based on the quality and clarity of textual prompts. Poorly formulated or very ambiguous prompts might result in less coherent outputs.

\textbf{Limited Support for Long Videos}
Currently, the model focuses on generating short videos of up to ten seconds. Extending the architecture to support longer durations while maintaining temporal consistency and prompt fidelity remains an open area for future research.

\textbf{Domain-Specific Generalization}
The model’s ability to adapt to domain-specific tasks (e.g., multi-view synthesis or fine-grained editing) has not been extensively tested, and further experimentation is required to assess its performance in specialized applications.

\section{Social Impact}

\textbf{Accessibility and Democratization}
Our model is designed with accessibility in mind. Unlike many state-of-the-art text-to-video models that require high-end hardware and significant computational resources, our model is optimized for efficiency and can run on consumer-grade GPUs. This design choice makes advanced text-to-video generation more accessible to researchers, developers, and enthusiasts who may not have access to expensive hardware setups.

\textbf{Open-Source Contribution}
By releasing our model as open-source, we aim to foster innovation and collaboration in the AI community. Open access to the model encourages diverse applications, including educational tools, creative content generation, and rapid prototyping for small and medium-sized enterprises, which may lack resources for training large models.

\textbf{Environmental Considerations}
The relatively small size of our model not only reduces the hardware requirements but also lowers the energy consumption associated with training and deployment. This contributes to a more sustainable approach to deploying AI technologies.

\textbf{Potential Risks and Mitigation}
While our model lowers barriers to entry, we recognize that making such technologies widely available could pose risks, such as misuse for generating misleading content. To mitigate this, we have included clear guidelines and disclaimers in our documentation to encourage responsible use.

\section{Conclusion}
In this paper, we introduced \textit{LTX-Video}, a state-of-the-art transformer-based latent diffusion model designed for text-to-video and image-to-video generation. By addressing key limitations of existing methods, such as constrained temporal modeling and inefficient spatial compression, \textit{LTX-Video} achieves faster-than-real-time generation while maintaining high motion fidelity, temporal consistency, and strong alignment with input prompts or conditioning frames.

At the core of \textit{LTX-Video} is a holistic approach to latent diffusion, which seamlessly integrates the Video-VAE and the denoising transformer. This integration is achieved by relocating the patchifying operation from the transformer’s input to the VAE encoder, enabling efficient processing within a compressed latent space. Furthermore, the model introduces a novel shared diffusion objective between the VAE decoder and the transformer, effectively fusing the final diffusion step with the latent-to-pixel decoding stage. This innovation ensures fine-detail generation without the need for additional upsampling modules.

\textit{LTX-Video} sets a new benchmark for text-to-video generation, outperforming SOTA open-source models of similar scale in speed and quality. Its ability to efficiently generate high-resolution videos while preserving coherence and adherence to prompts underscores the potential of latent diffusion models for video generation tasks.

The accessibility of \textit{LTX-Video} further amplifies its impact, as its efficient design allows it to run on consumer-grade GPUs, democratizing advanced video generation capabilities. By lowering hardware requirements, \textit{LTX-Video} opens the door to researchers, developers, and creative professionals who may not have access to high-end compute resources.

Future work may explore extending \textit{LTX-Video}’s architecture to support longer videos, incorporate advanced temporal coherence techniques, and investigate its adaptability to domain-specific tasks such as multi-view video synthesis or fine-grained motion editing. By enabling faster, high-quality outputs with lower compute demands, \textit{LTX-Video} represents a significant step forward in creative content creation, accessible AI, and scalable video modeling.

\begin{figure}[htbp]
\centering
\tiny
\setlength\tabcolsep{1pt} 
\renewcommand\arraystretch{1.0} 

\successprompt{A young woman in a traditional Mongolian dress is peeking through a sheer white curtain, her face showing a mix of curiosity and apprehension. The woman has long black hair styled in two braids, adorned with white beads}, and her eyes are wide with a hint of surprise. \successprompt{Her dress is a vibrant blue with intricate gold embroidery, and she wears a matching headband with a similar design}...
\begin{tabular}{@{}cccc@{}}
    \includegraphics[width=0.24\textwidth]{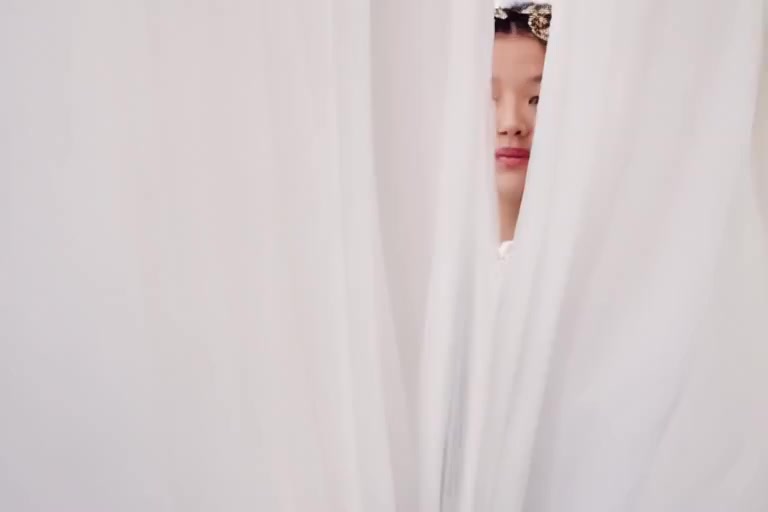} &
    \includegraphics[width=0.24\textwidth]{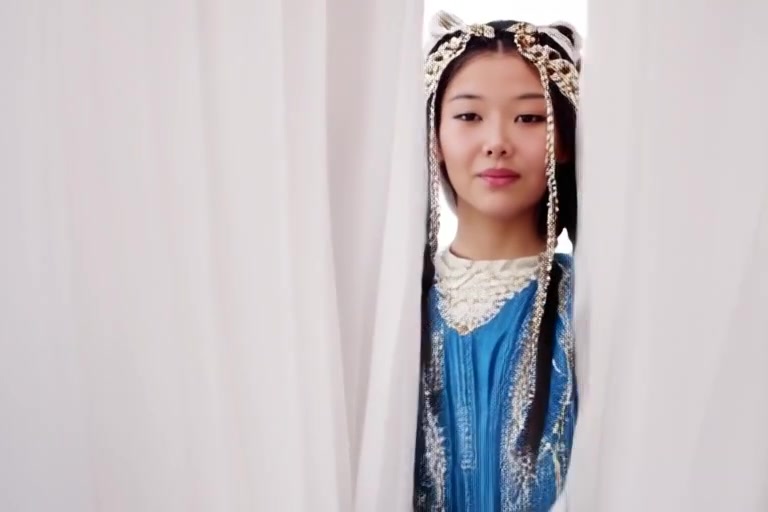} &
    \includegraphics[width=0.24\textwidth]{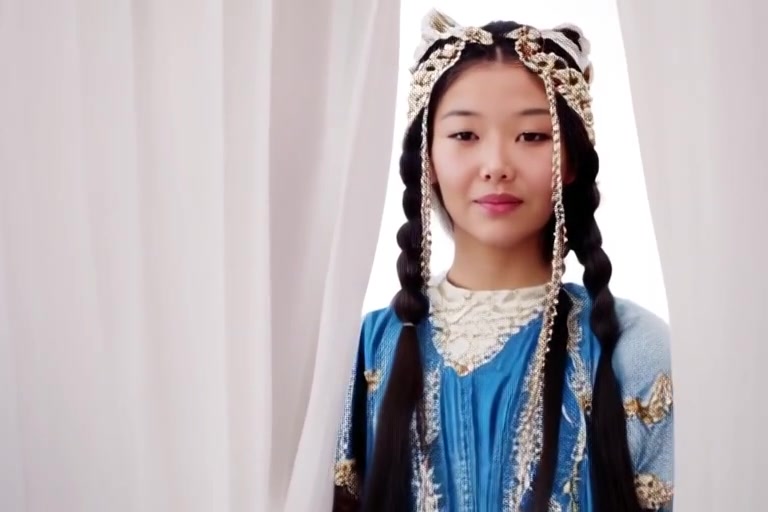} &
    \includegraphics[width=0.24\textwidth]{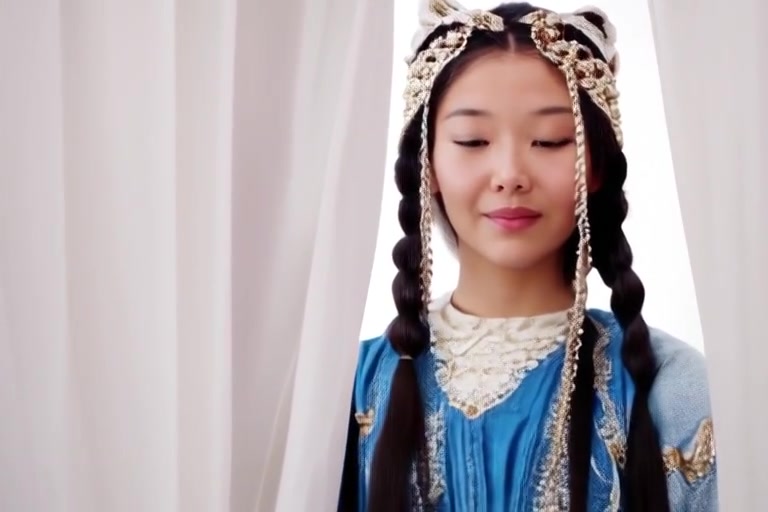} \\
\end{tabular}

\successprompt{A woman with long brown hair and bangs sits in the driver’s seat of a car, smiling slightly as she looks to her} right. \successprompt{She is wearing a black jacket over a white shirt. The camera is positioned outside the car}, looking in through the driver’s side window. \successprompt{The car is dark-colored and the background is blurred}, suggesting a residential street...
\begin{tabular}{@{}cccc@{}}
    \includegraphics[width=0.24\textwidth]{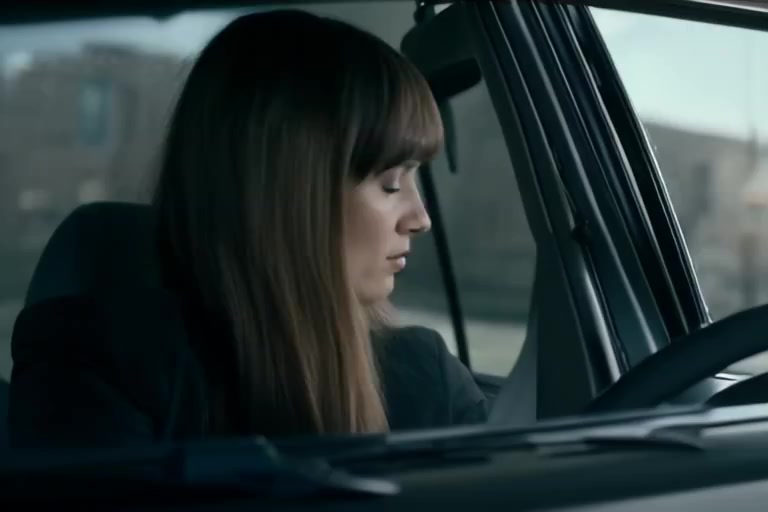} &
    \includegraphics[width=0.24\textwidth]{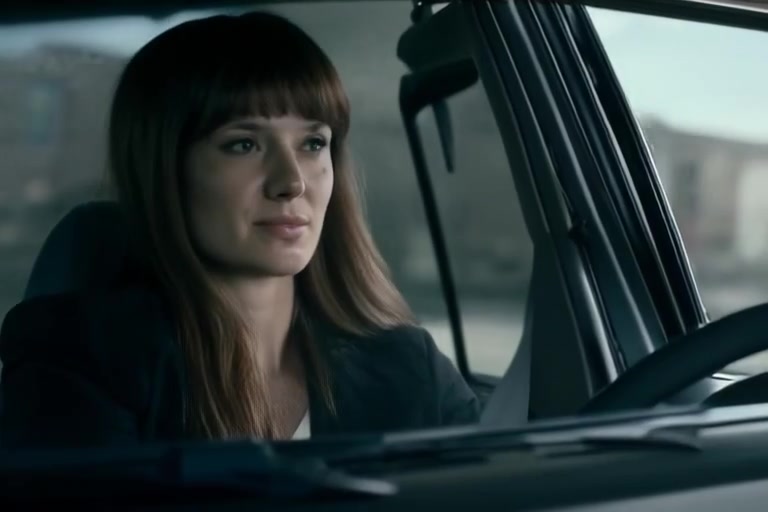} &
    \includegraphics[width=0.24\textwidth]{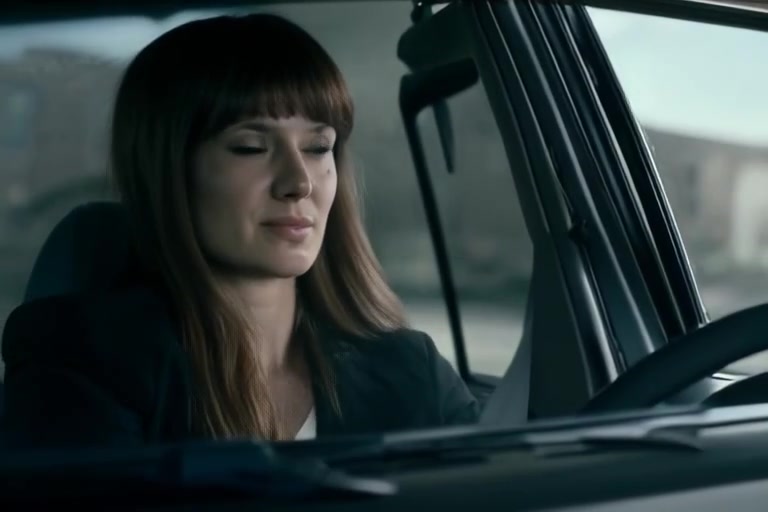} &
    \includegraphics[width=0.24\textwidth]{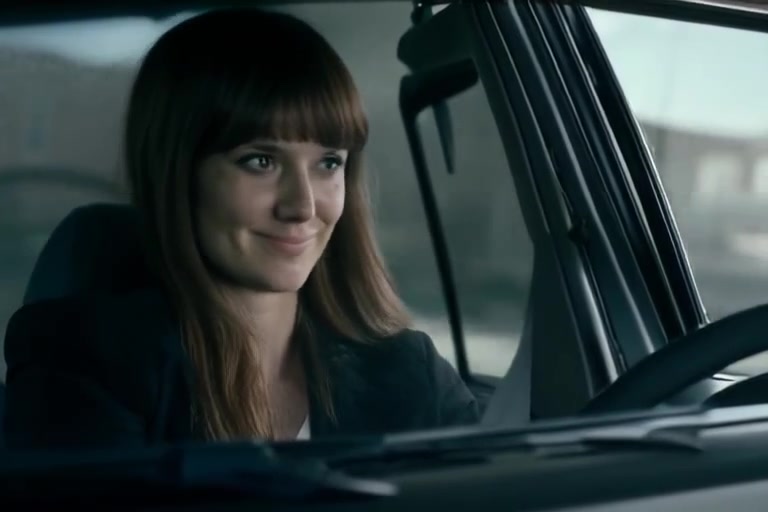} \\
\end{tabular}

\successprompt{A woman with blood on her face and a white tank top looks down and to her right, then back up as she speaks. She has dark hair pulled back, light skin, and her face and chest are covered in blood. The camera angle is a close-up, focused on the woman's face and upper torso. The lighting is dim and blue-toned, creating a somber and intense atmosphere. The scene appears to be from a movie or TV show.}
\begin{tabular}{@{}cccc@{}}
    \includegraphics[width=0.24\textwidth]{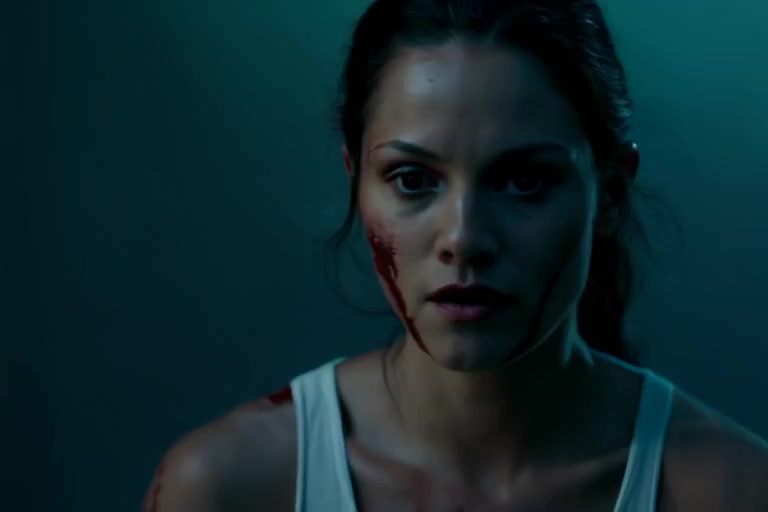} &
    \includegraphics[width=0.24\textwidth]{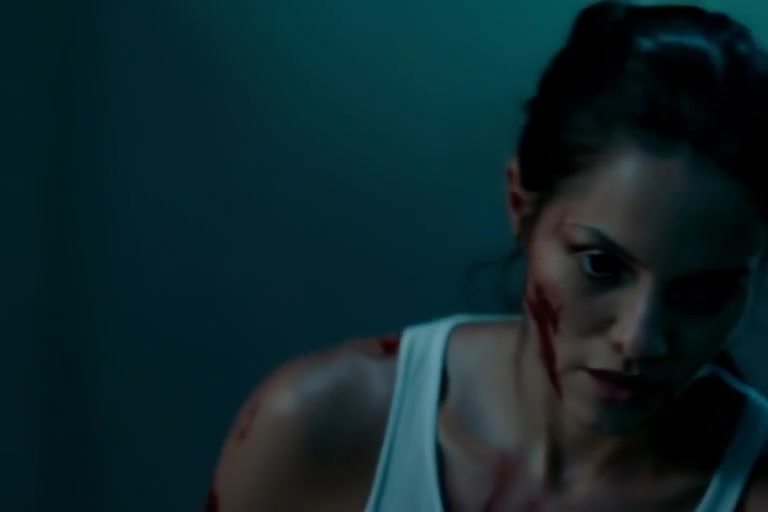} &
    \includegraphics[width=0.24\textwidth]{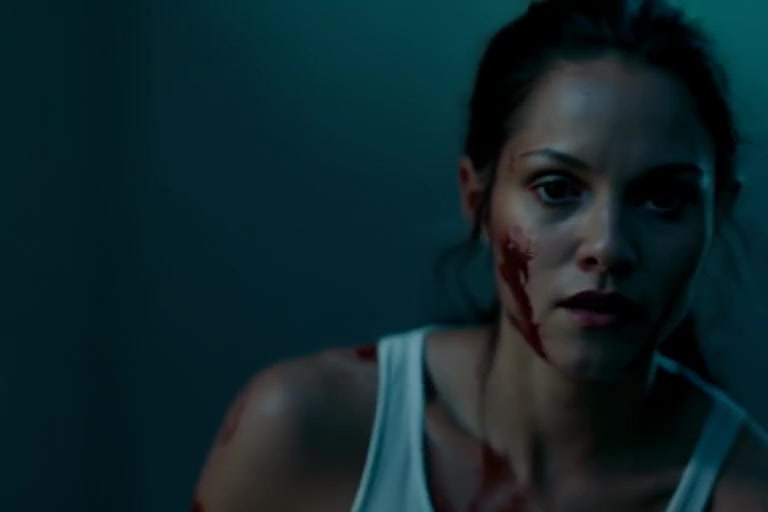} &
    \includegraphics[width=0.24\textwidth]{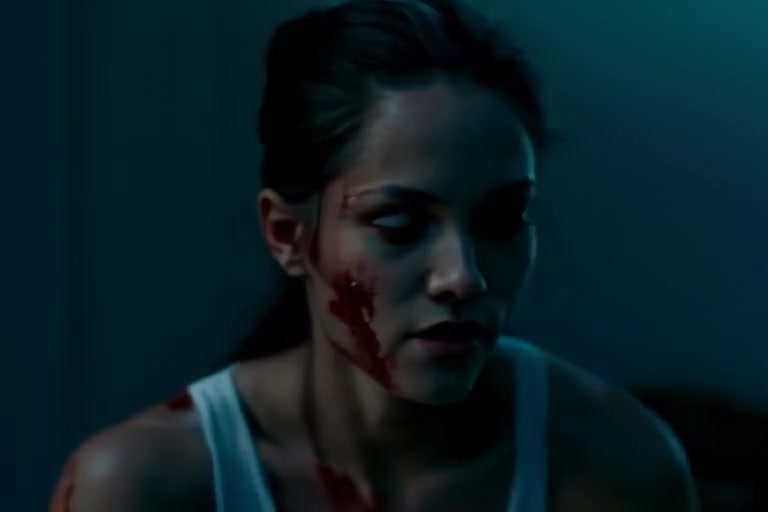} \\
\end{tabular}

\successprompt{A man in a dimly lit room talks on a vintage telephone, hangs up, and looks down with a sad expression. He holds the black rotary phone to his right ear with his} right hand, his left hand holding a rocks glass with amber liquid. \successprompt{He wears a brown suit jacket over a white shirt}, and a gold ring on his left ring finger. \successprompt{His short hair is neatly combed, and he has light skin with visible wrinkles around his eyes}. The camera...
\begin{tabular}{@{}cccc@{}}
    \includegraphics[width=0.24\textwidth]{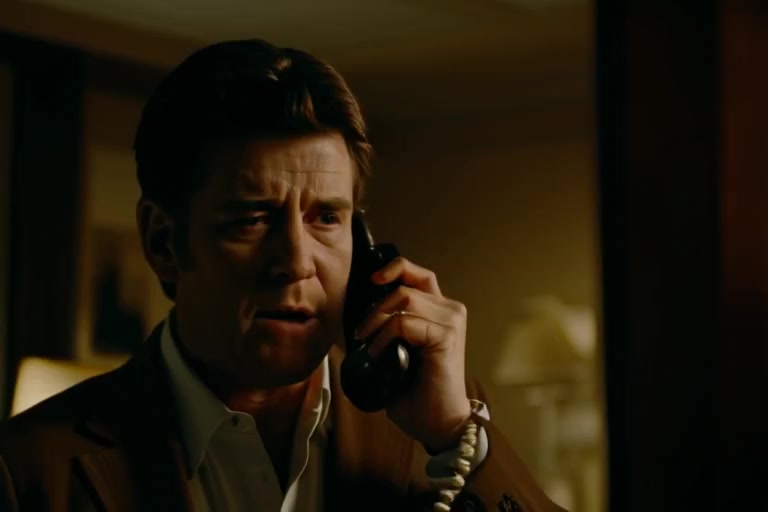} &
    \includegraphics[width=0.24\textwidth]{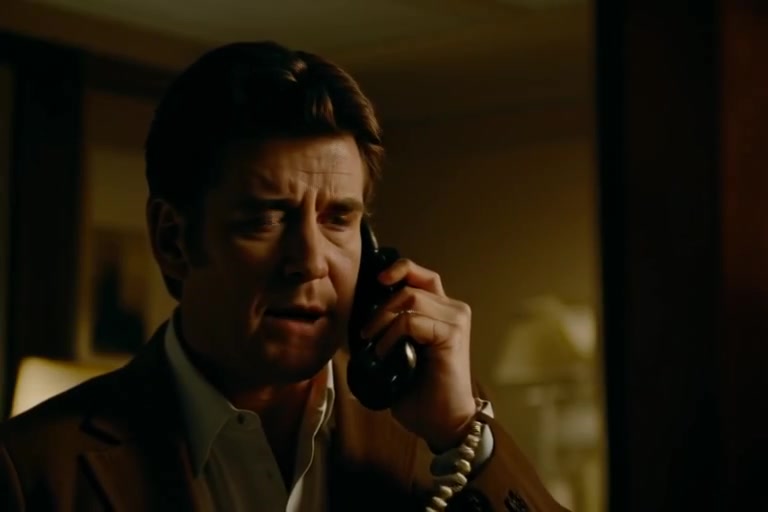} &
    \includegraphics[width=0.24\textwidth]{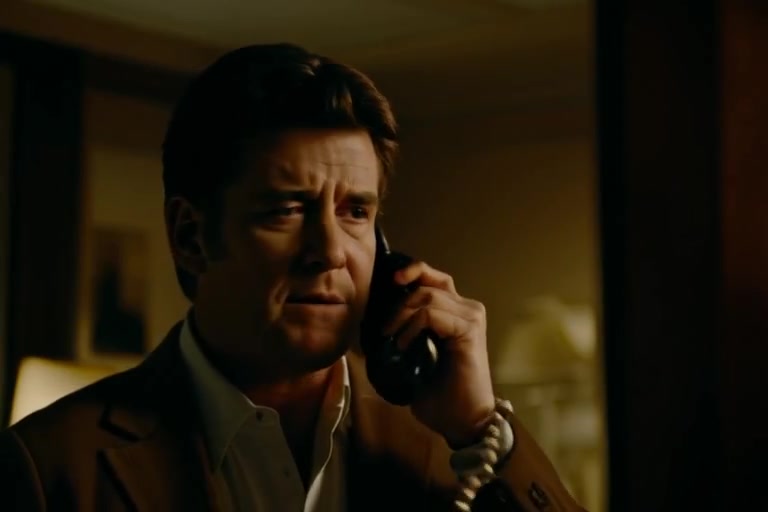} &
    \includegraphics[width=0.24\textwidth]{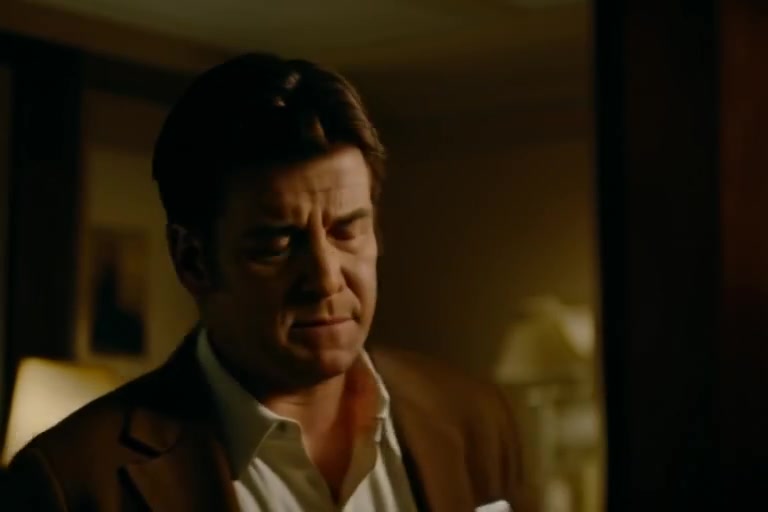} \\
\end{tabular}

\successprompt{A man in a suit enters a room and speaks to two women sitting on a couch. The man, wearing a dark suit with a gold tie, enters the room from the left and walks towards the center of the frame. He has short gray hair, light skin, and a serious expression.} He places his right hand on the back of a chair as he approaches the couch. Two women are seated on a light-colored couch in the background. \successprompt{The woman on the left wears a light blue} sweater \successprompt{and has short blonde hair.}  \successprompt{The scene appears to be from a film or television show}.
\begin{tabular}{@{}cccc@{}}
    \includegraphics[width=0.24\textwidth]{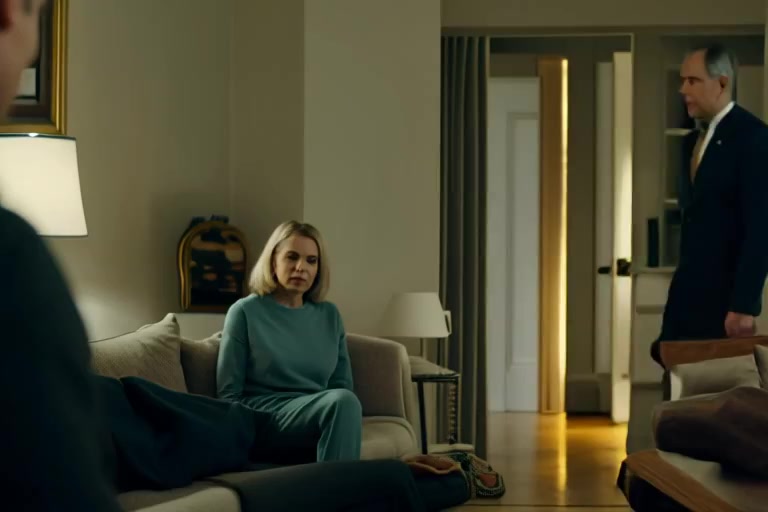} &
    \includegraphics[width=0.24\textwidth]{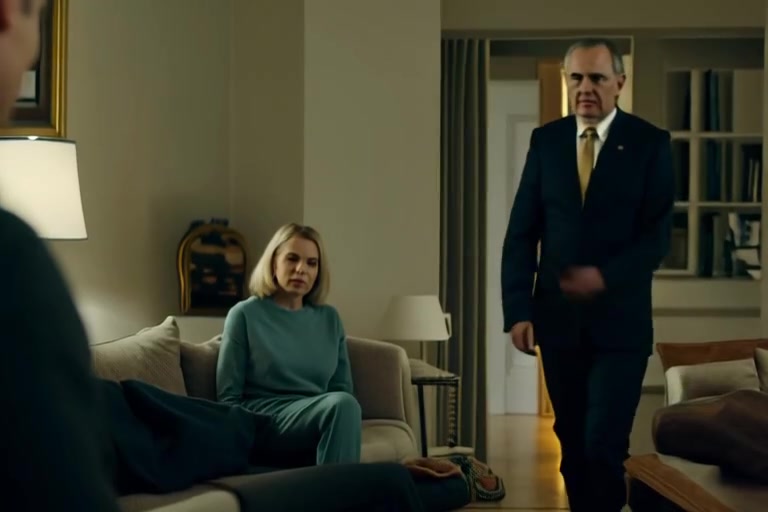} &
    \includegraphics[width=0.24\textwidth]{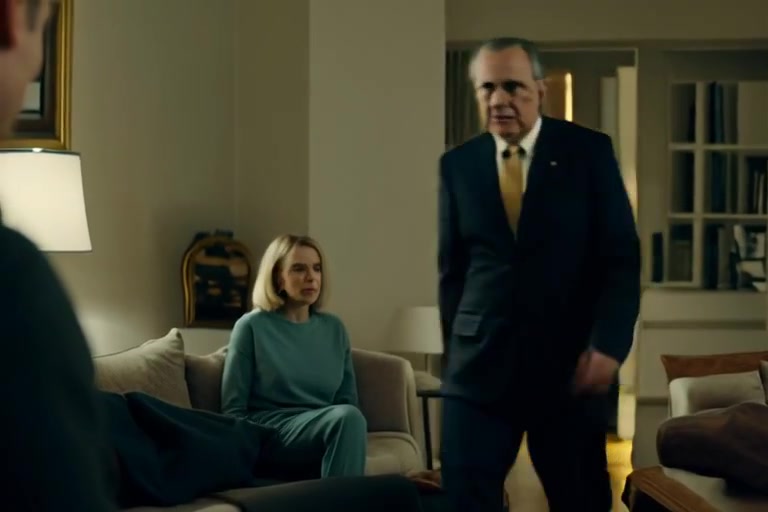} &
    \includegraphics[width=0.24\textwidth]{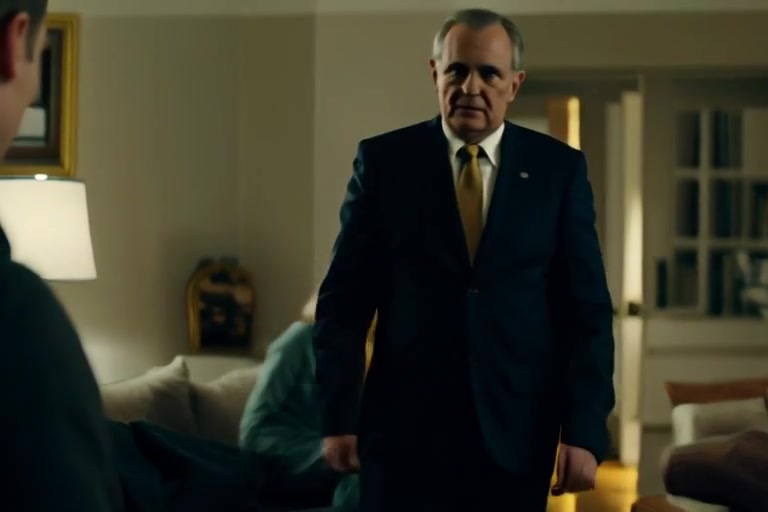} \\
\end{tabular}

\successprompt{A young woman} in a classroom \successprompt{receives a phone call, listens intently}, and hangs up. \successprompt{The woman, with dark skin and long black hair styled in braids, wears a white turtleneck sweater, an orange cardigan, and a silver chain necklace; she holds a white smartphone to her right ear, her expression shifting from a smile to concern as she listens}; she raises her eyebrows...
\begin{tabular}{@{}cccc@{}}
    \includegraphics[width=0.24\textwidth]{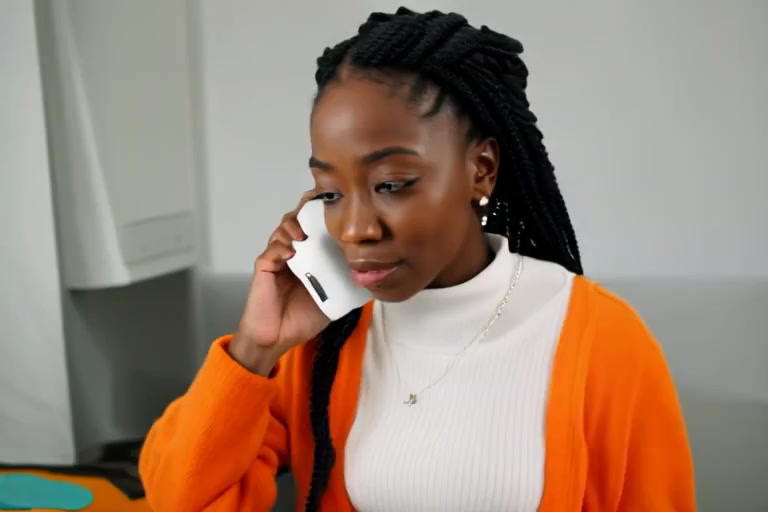} &
    \includegraphics[width=0.24\textwidth]{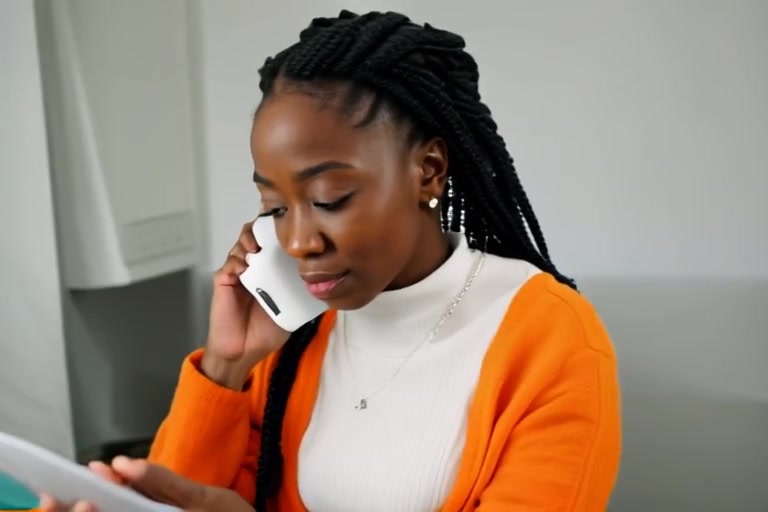} &
    \includegraphics[width=0.24\textwidth]{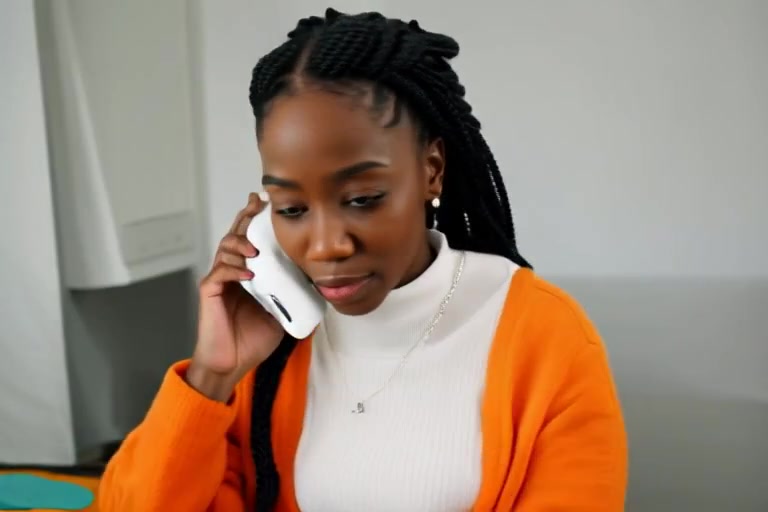} &
    \includegraphics[width=0.24\textwidth]{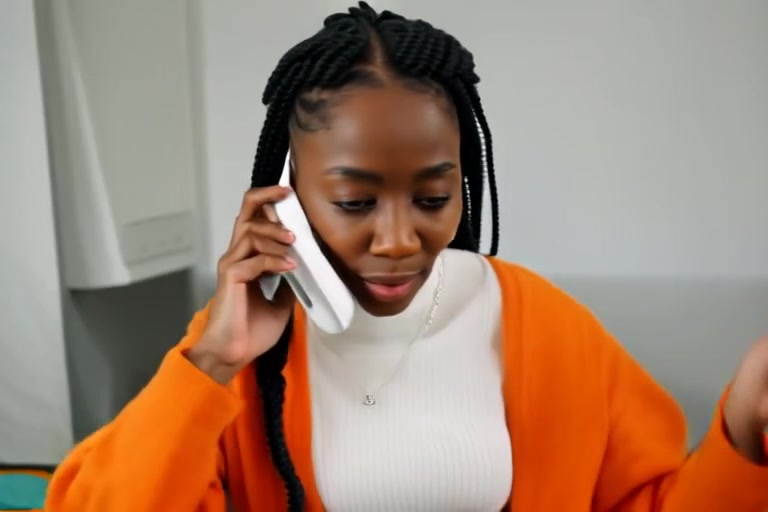} \\
\end{tabular}

\successprompt{A man walks towards a window, looks out, and then turns around. He has short, dark hair, dark skin, and is wearing a brown coat over a red and gray scarf. He walks from left to right towards a window, his gaze fixed on something outside. The camera}...
\begin{tabular}{@{}cccc@{}}
    \includegraphics[width=0.24\textwidth]{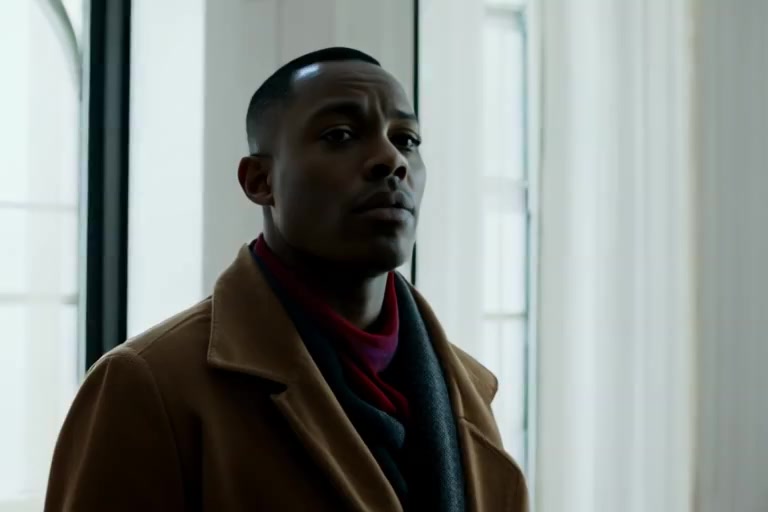} &
    \includegraphics[width=0.24\textwidth]{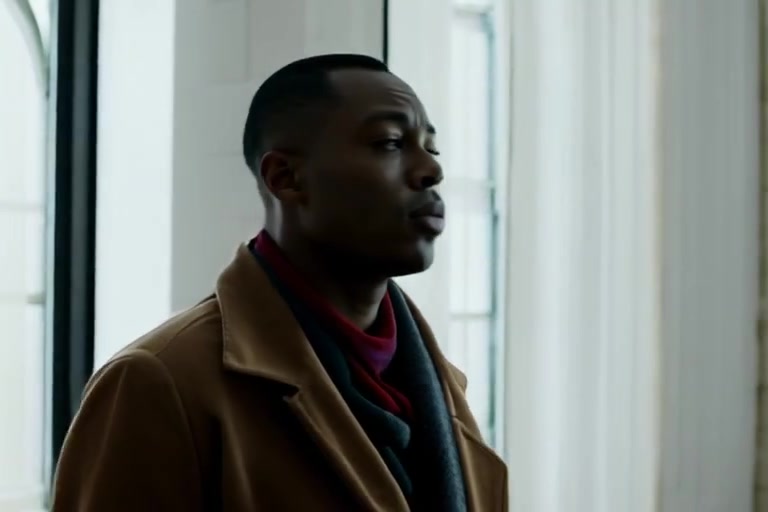} &
    \includegraphics[width=0.24\textwidth]{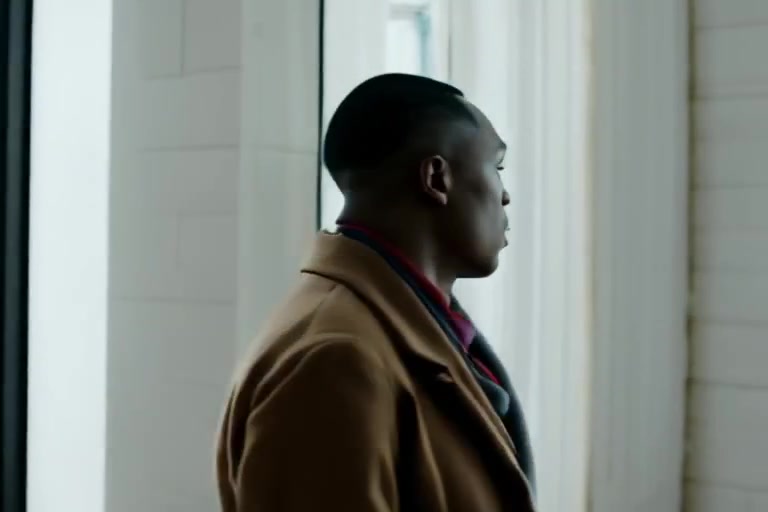} &
    \includegraphics[width=0.24\textwidth]{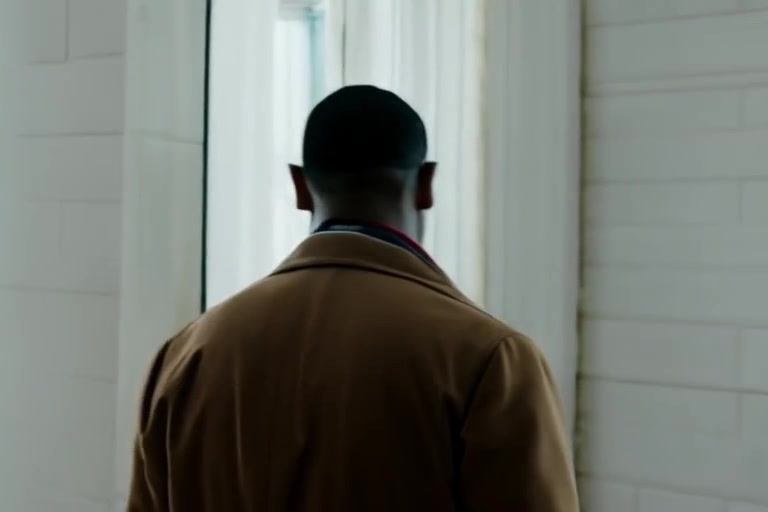} \\
\end{tabular}

\caption{Additional text-to-video samples generated by \textit{LTX-Video}, highlighting our model's high level of prompt adherence, visual quality and motion fidelity. Each row shows evenly-spaced frames from a generated 5-second video. Prompt adherence is indicated in \successprompt{green}.}
\label{fig:more_results}
\end{figure}

\begin{figure}[htbp]
\centering
\setlength\tabcolsep{1pt} 
\renewcommand\arraystretch{0.8} 
\begin{tabular}{@{}c|ccc@{}}

\includegraphics[width=0.245\textwidth]{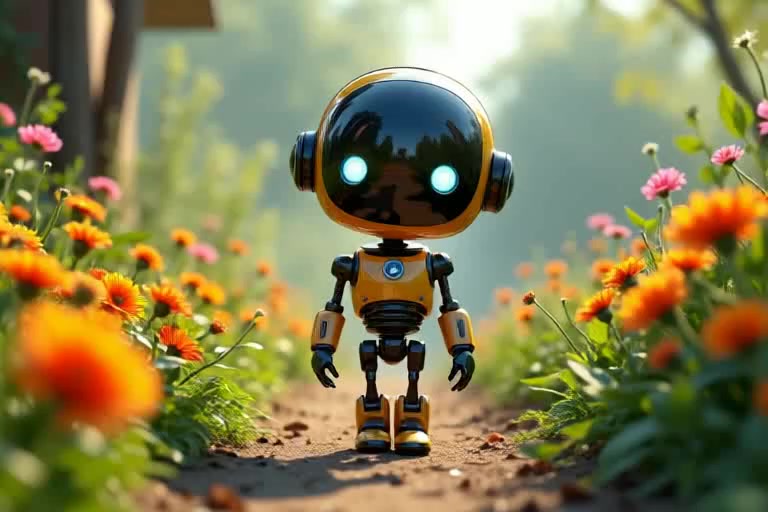} &
\includegraphics[width=0.245\textwidth]{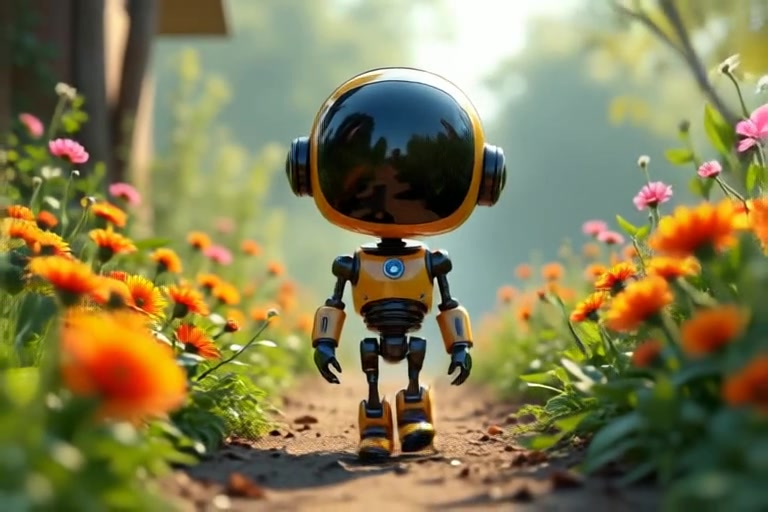} &
\includegraphics[width=0.245\textwidth]{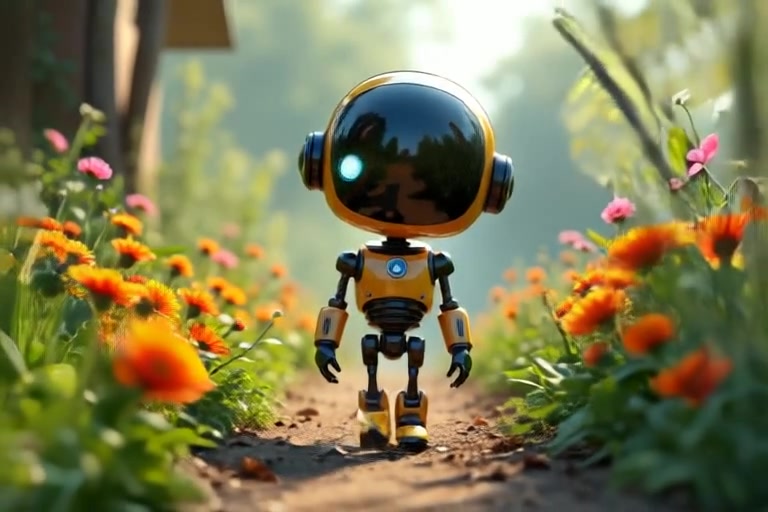} &
\includegraphics[width=0.245\textwidth]{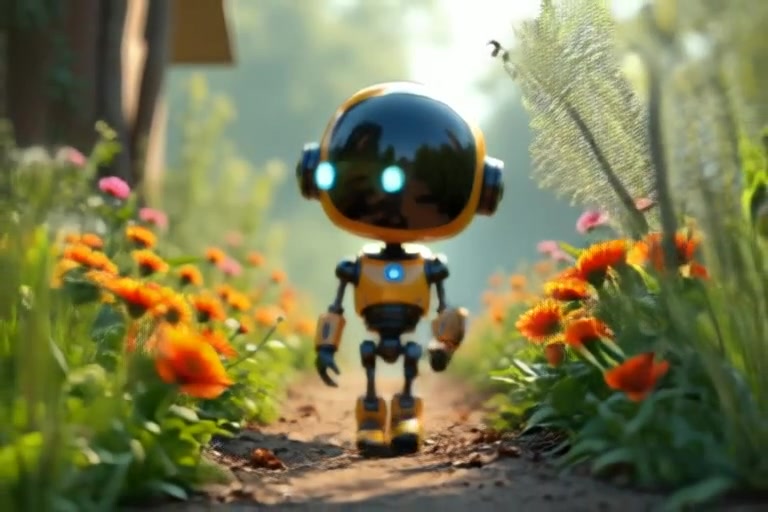} \\

\includegraphics[width=0.245\textwidth]{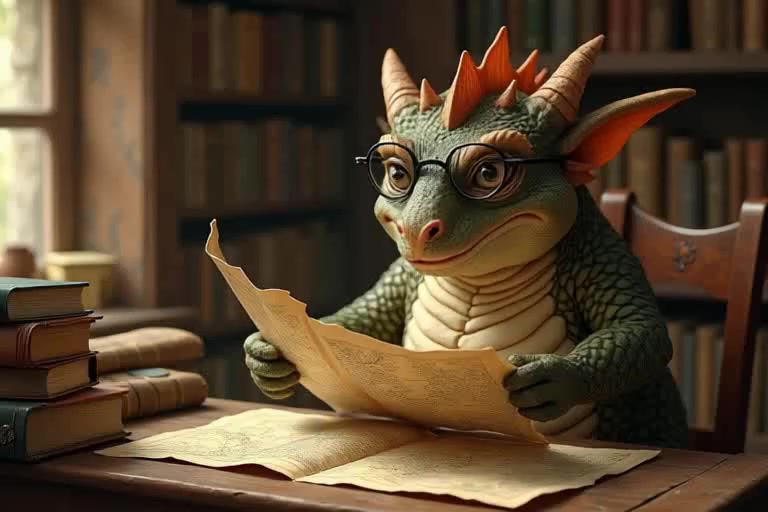} &
\includegraphics[width=0.245\textwidth]{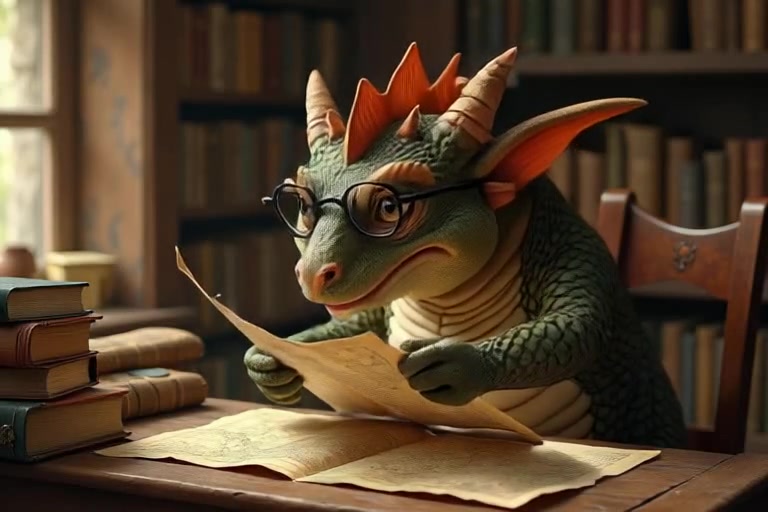} &
\includegraphics[width=0.245\textwidth]{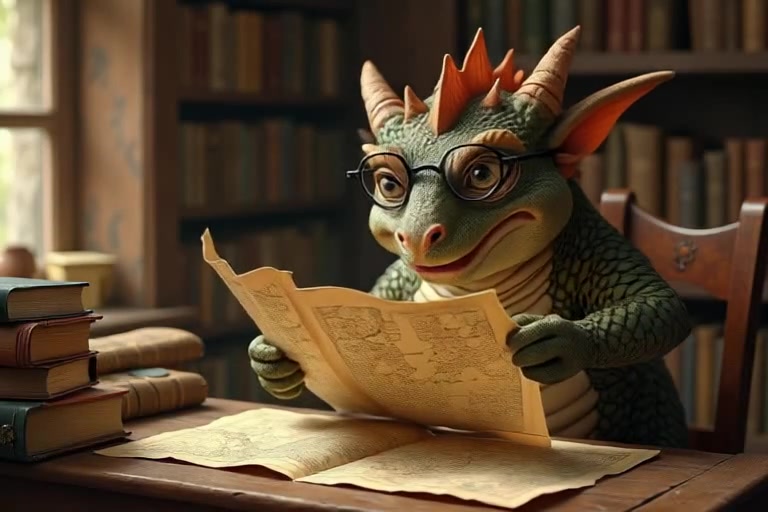} &
\includegraphics[width=0.245\textwidth]{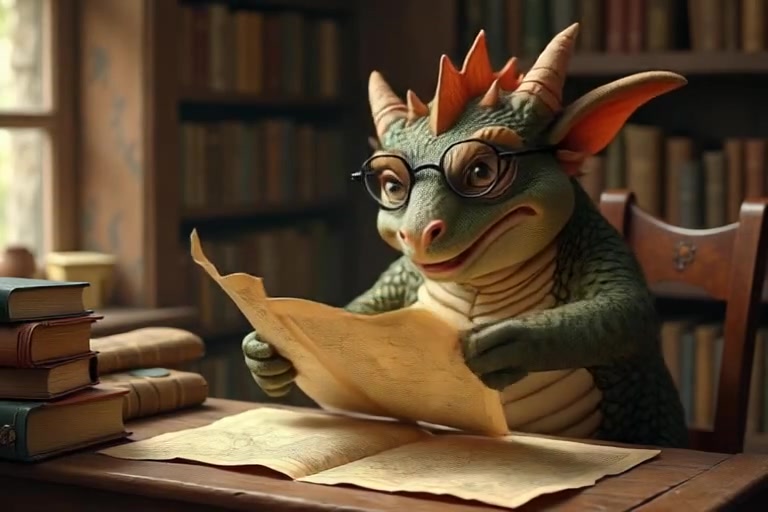} \\

\includegraphics[width=0.245\textwidth]{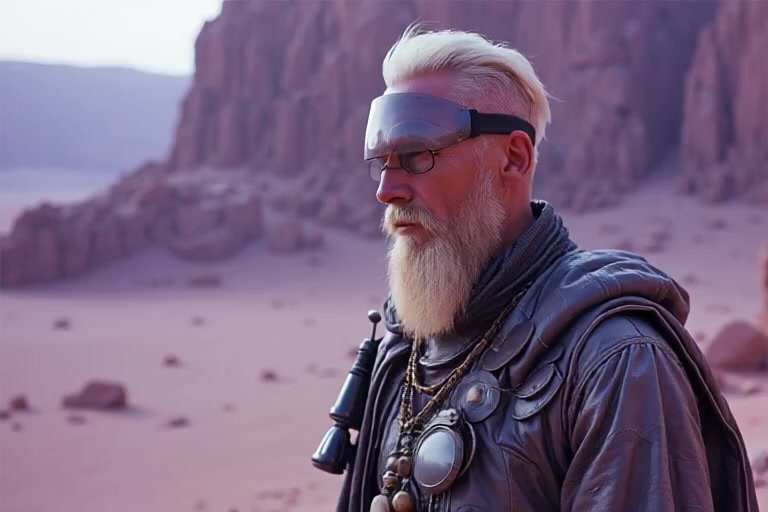} &
\includegraphics[width=0.245\textwidth]{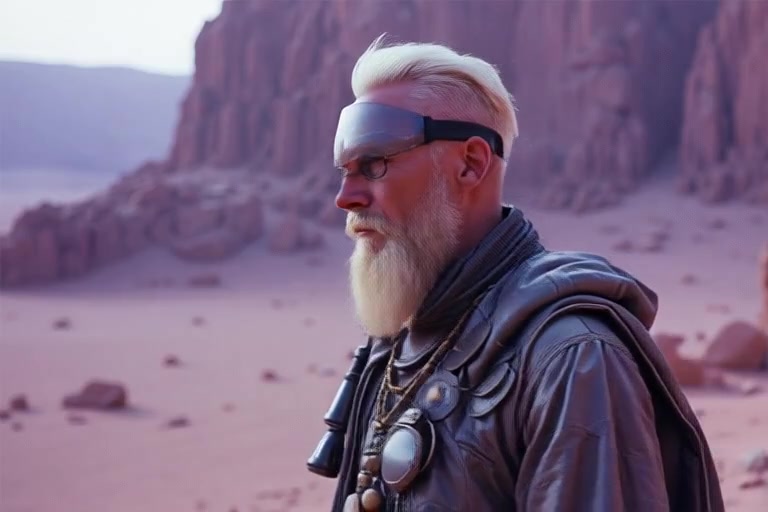} &
\includegraphics[width=0.245\textwidth]{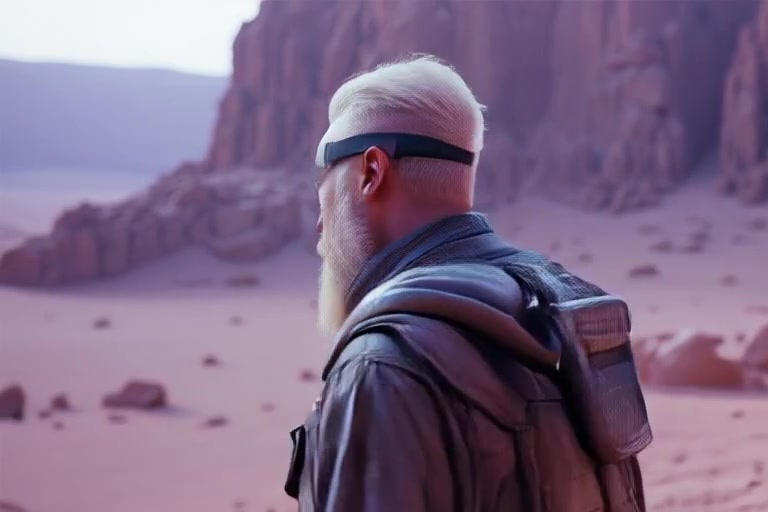} &
\includegraphics[width=0.245\textwidth]{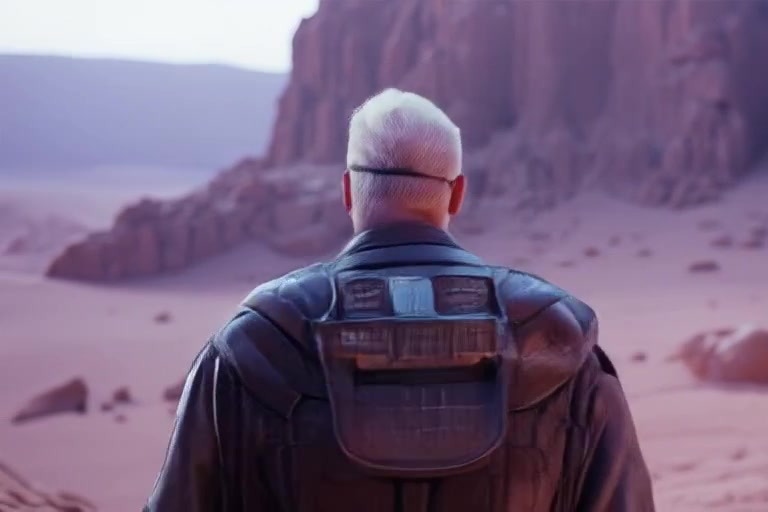} \\

\includegraphics[width=0.245\textwidth]{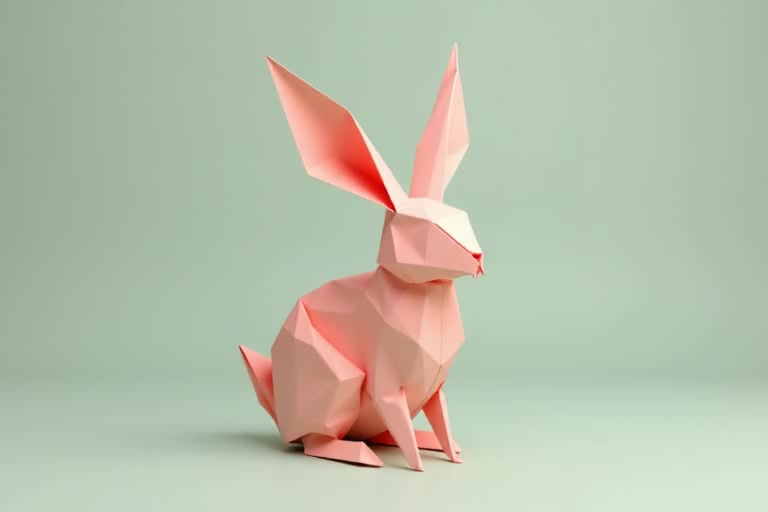} & \includegraphics[width=0.245\textwidth]{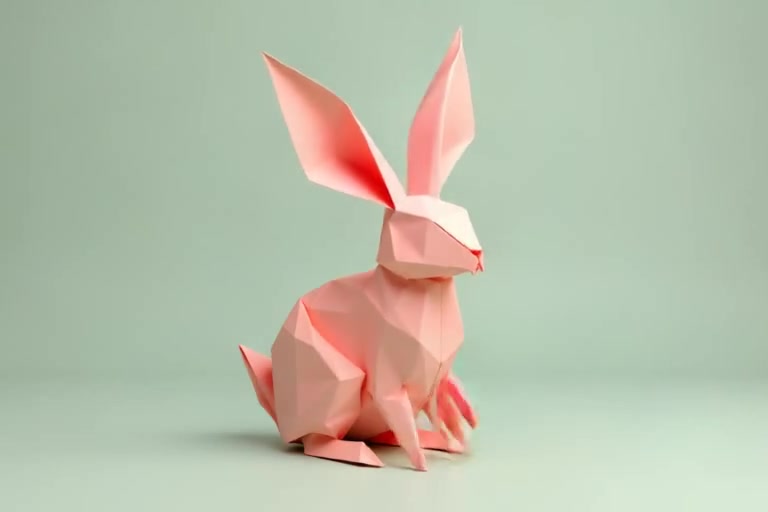} & \includegraphics[width=0.245\textwidth]{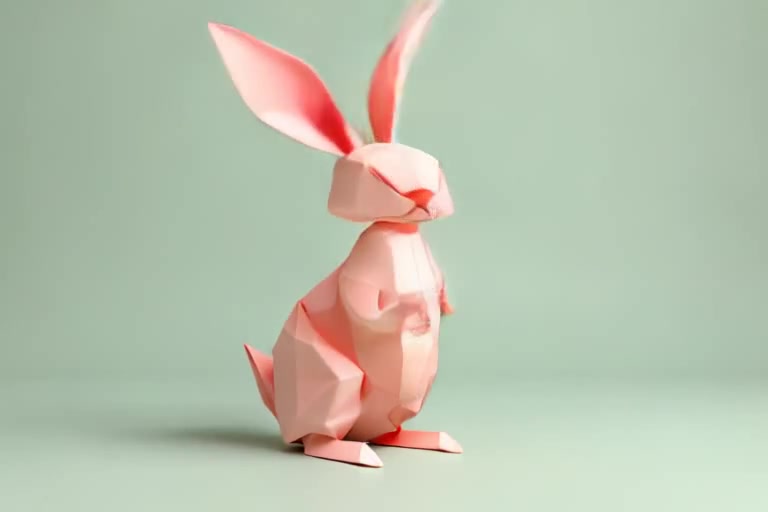} & \includegraphics[width=0.245\textwidth]{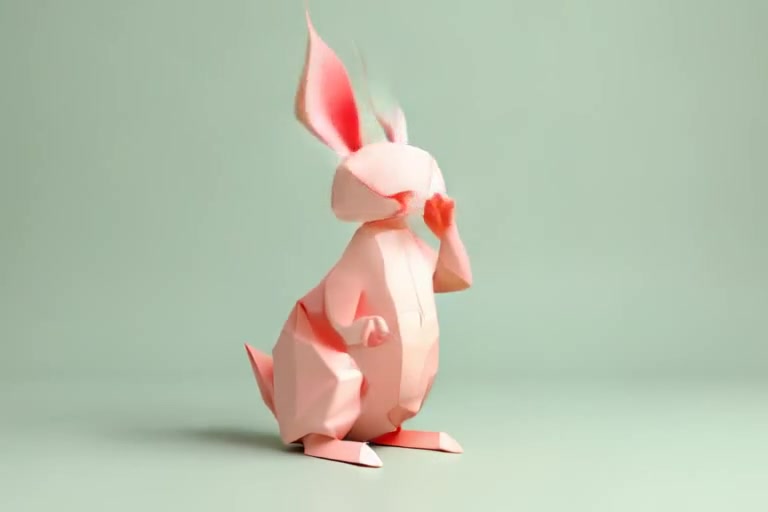} \\

\includegraphics[width=0.245\textwidth]{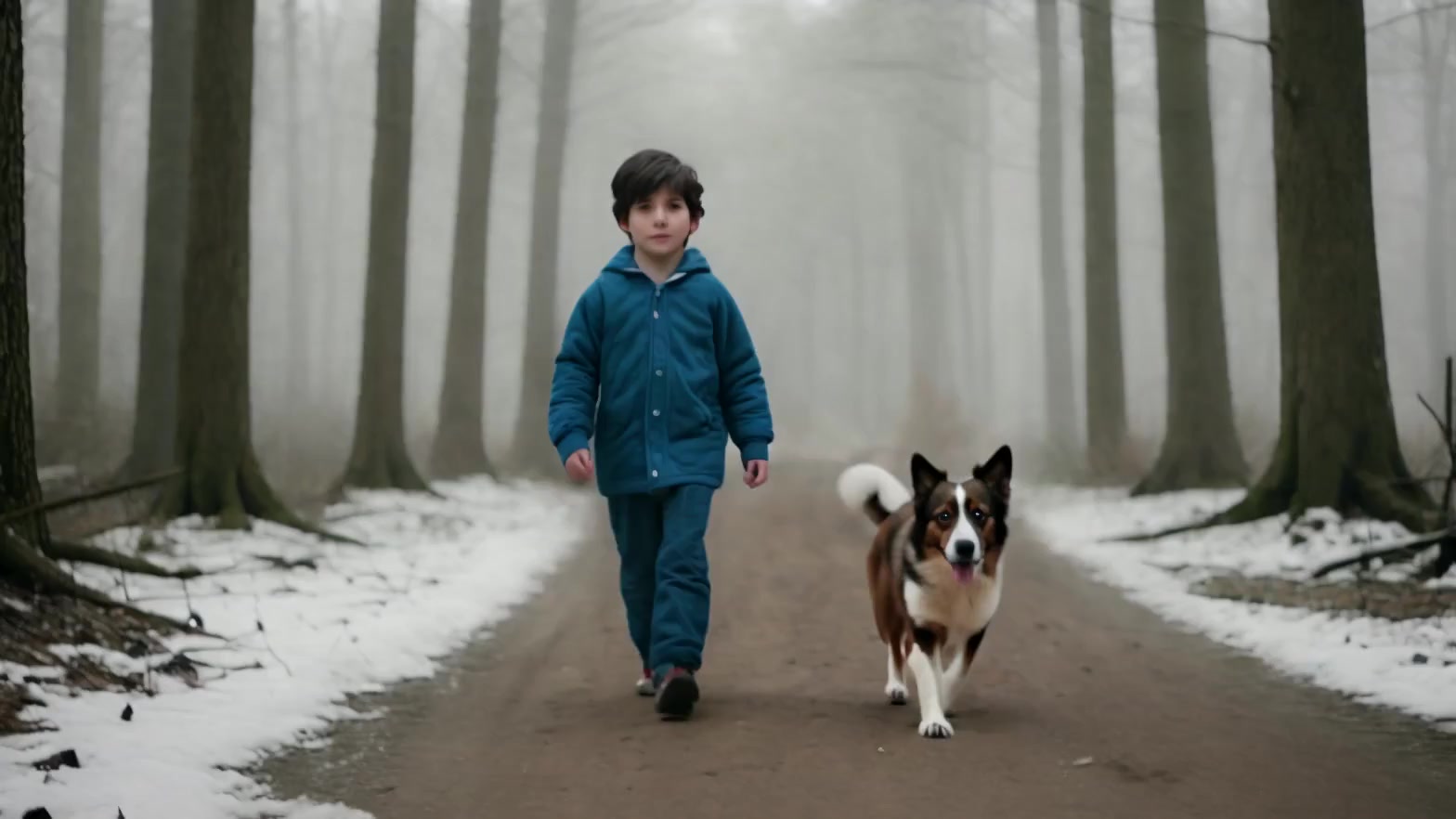} & \includegraphics[width=0.245\textwidth]{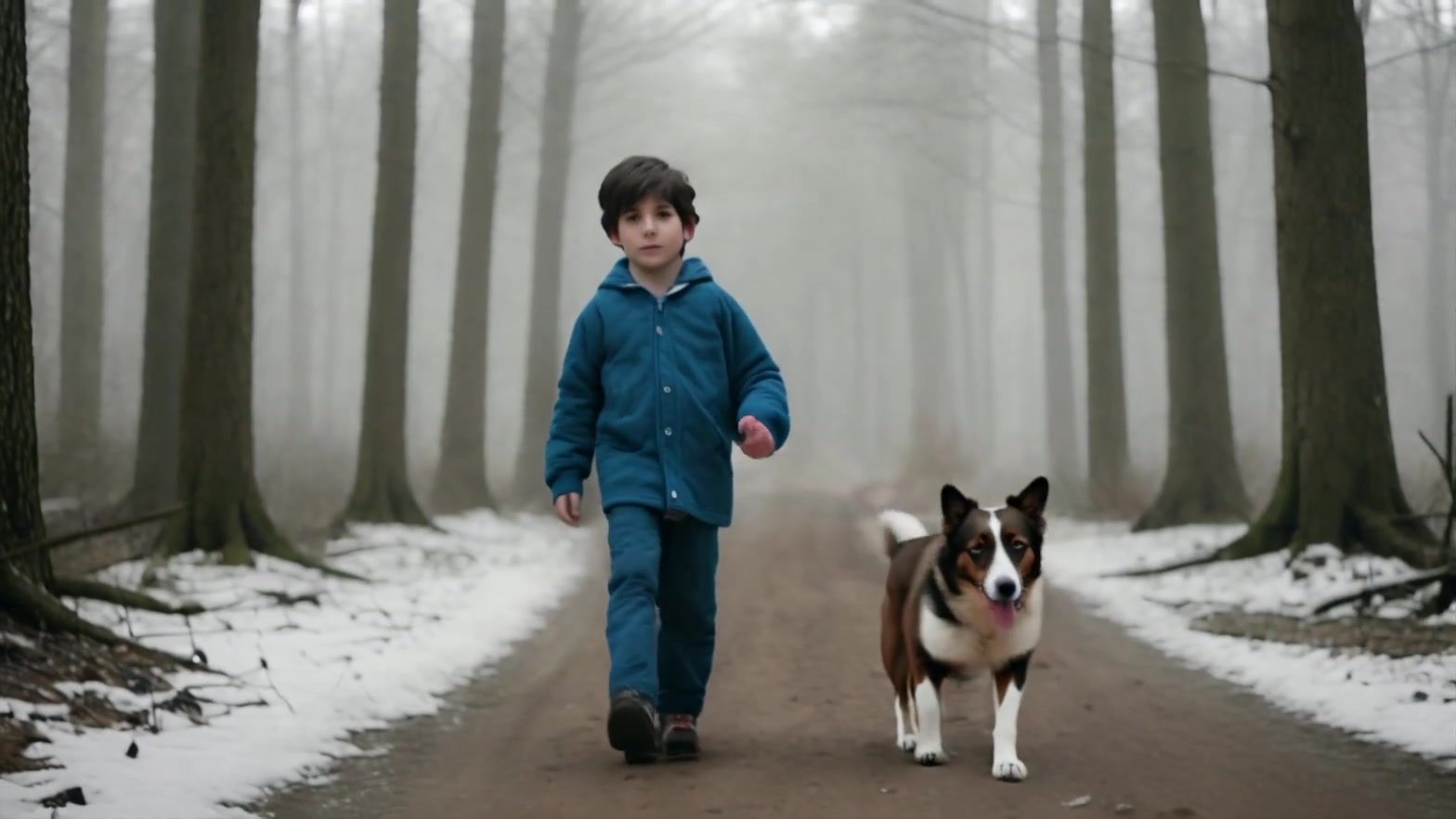} & \includegraphics[width=0.245\textwidth]{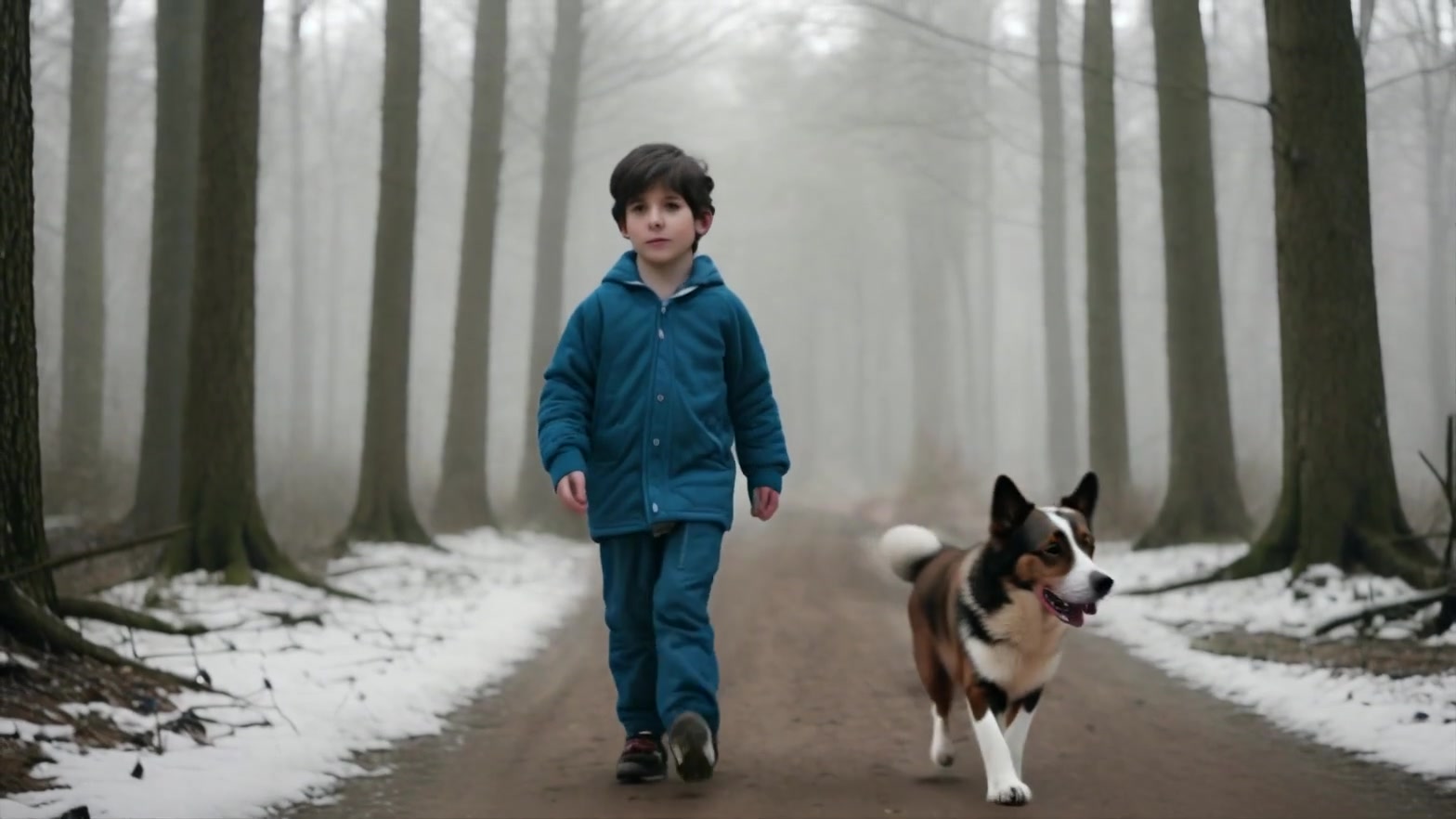} & \includegraphics[width=0.245\textwidth]{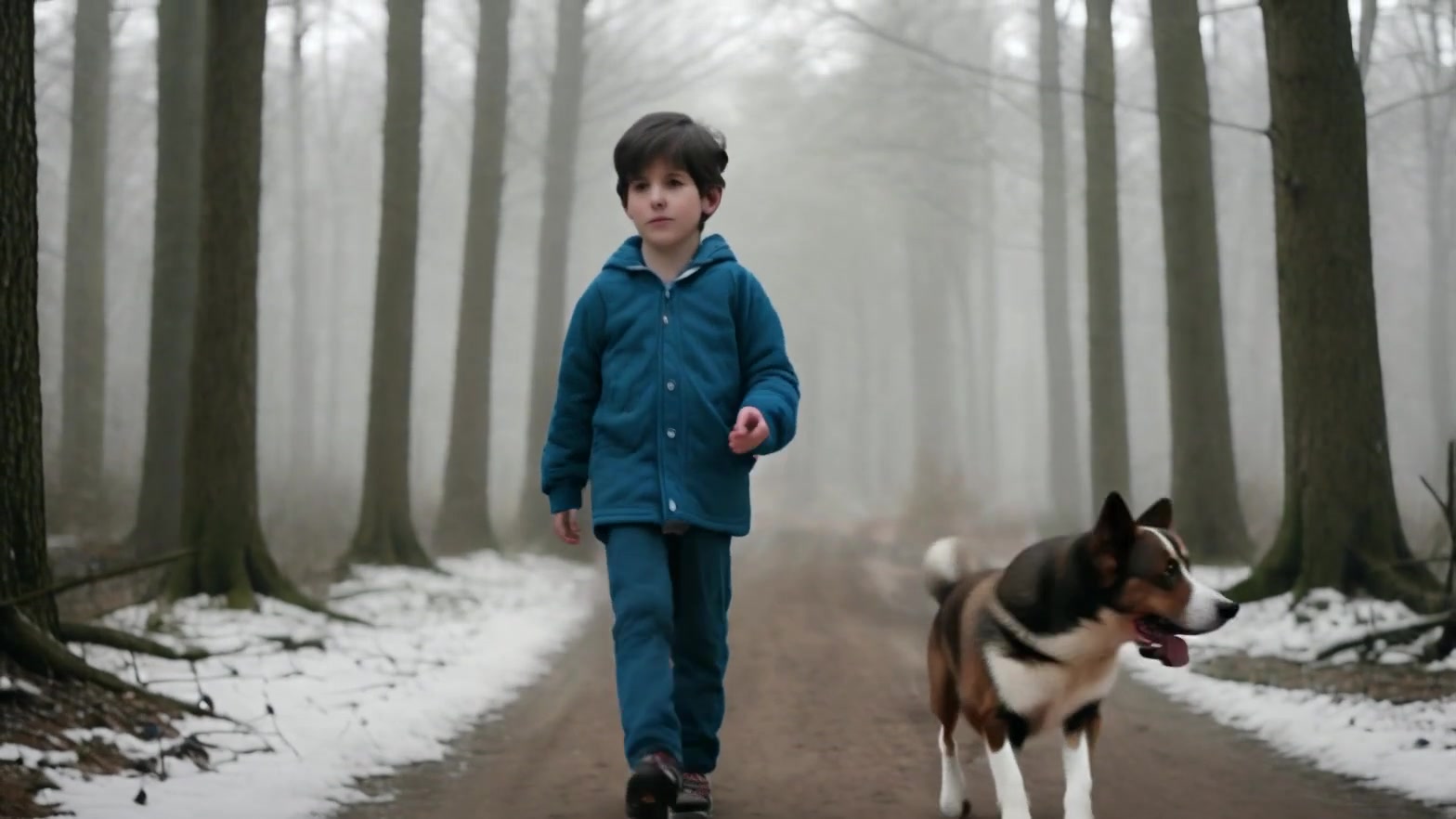} \\

\includegraphics[width=0.245\textwidth]{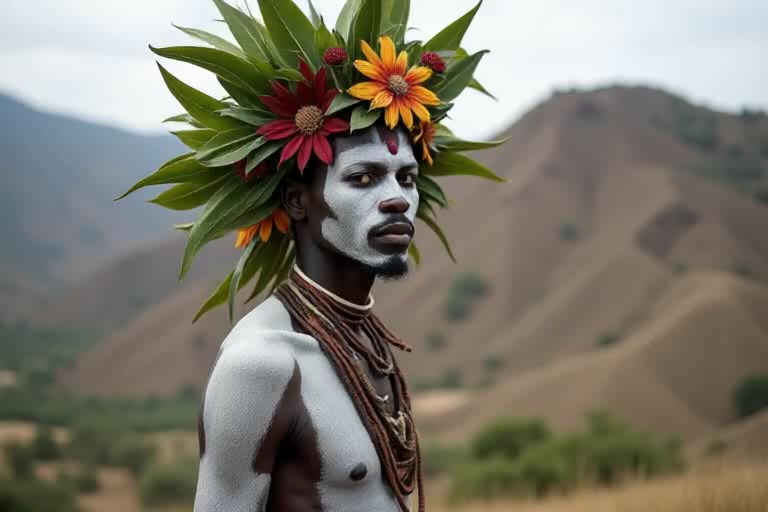} &
\includegraphics[width=0.245\textwidth]{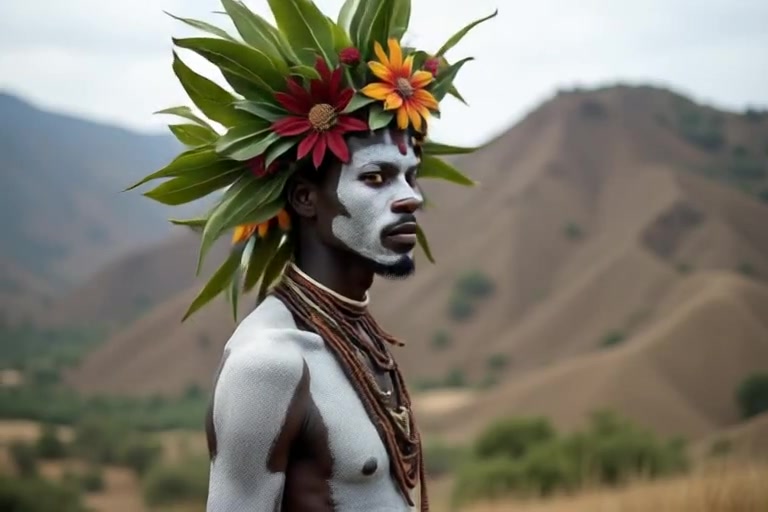} &
\includegraphics[width=0.245\textwidth]{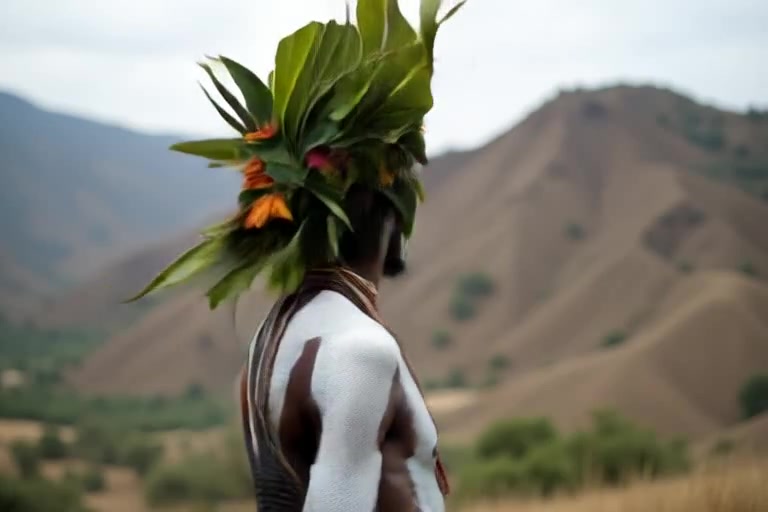} &
\includegraphics[width=0.245\textwidth]{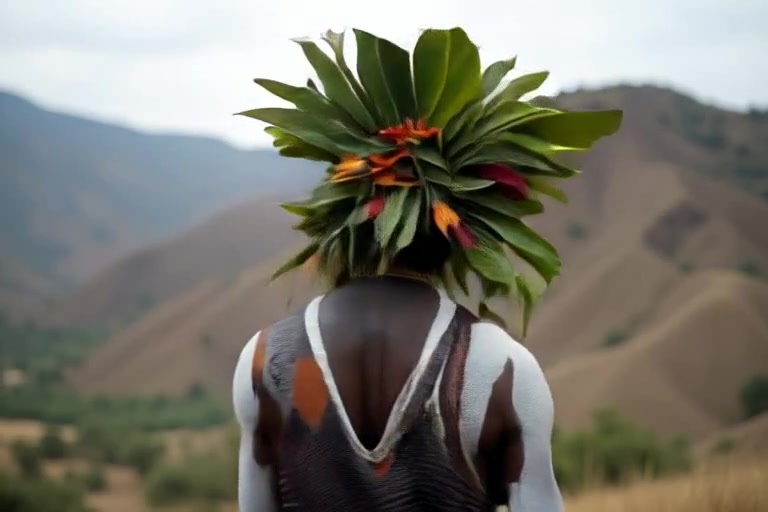} \\

\includegraphics[width=0.245\textwidth]{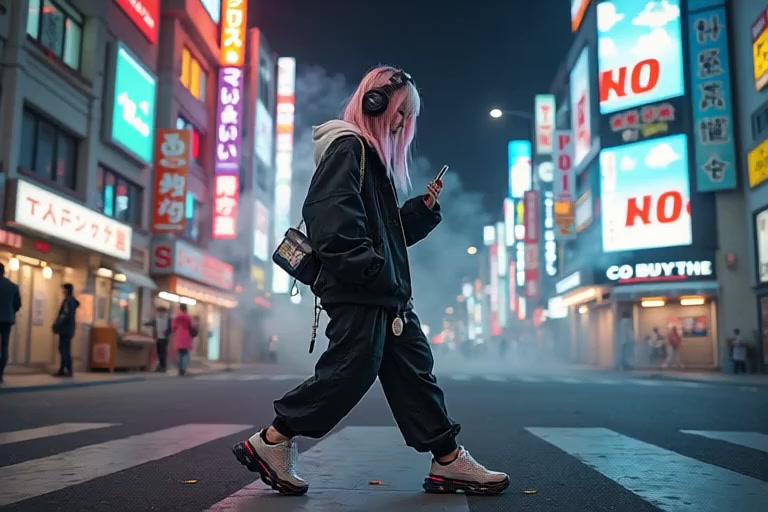} & \includegraphics[width=0.245\textwidth]{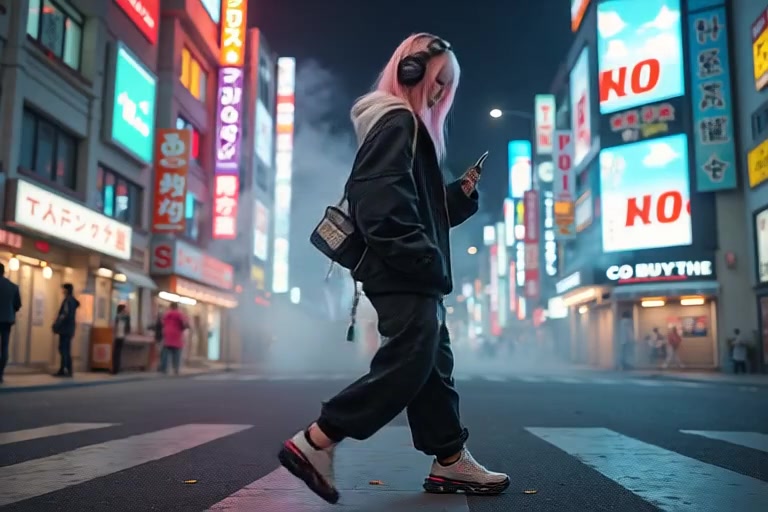} & \includegraphics[width=0.245\textwidth]{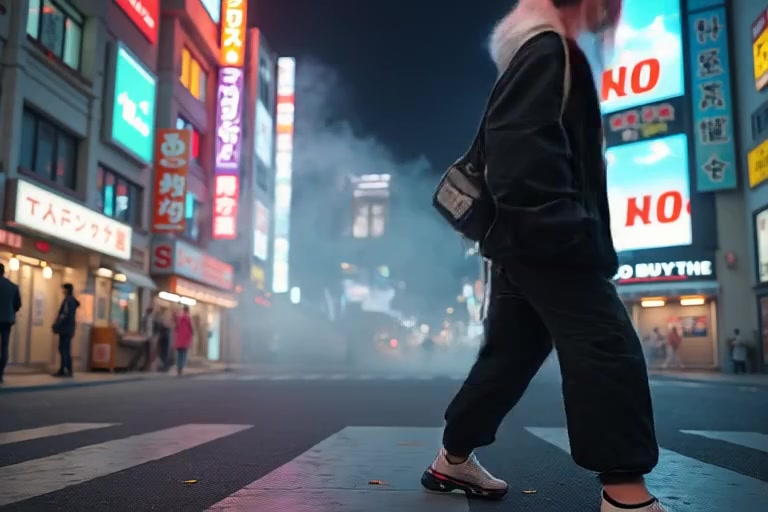} & \includegraphics[width=0.245\textwidth]{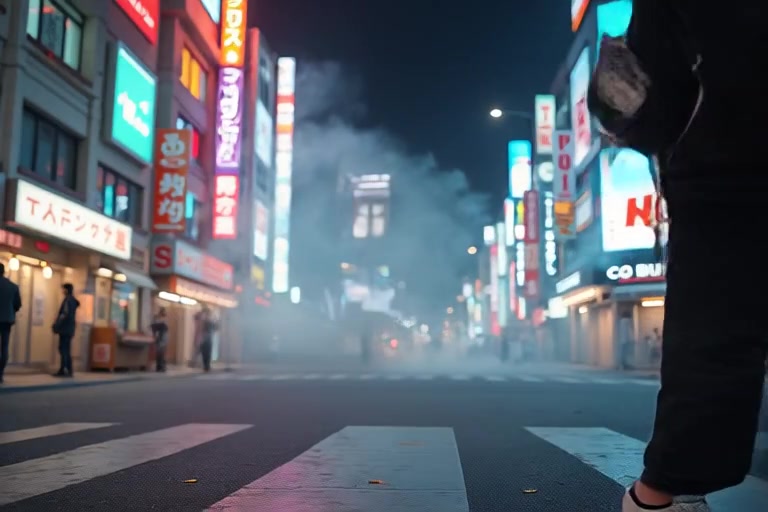} \\

\includegraphics[width=0.245\textwidth]{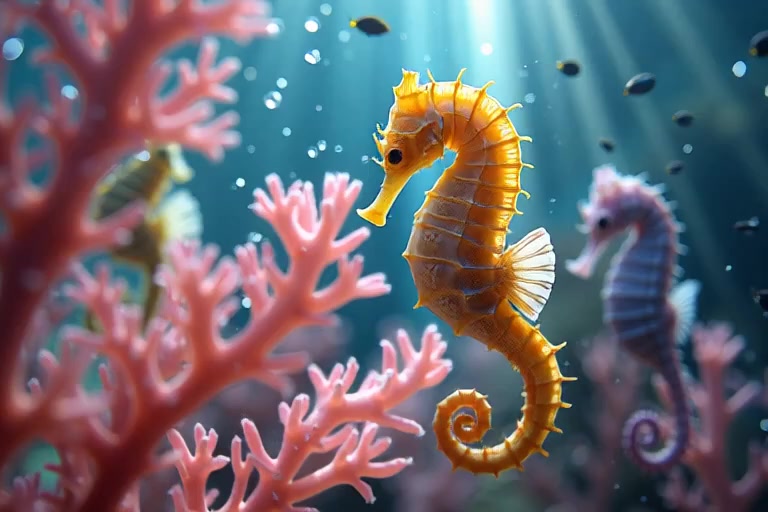} & \includegraphics[width=0.245\textwidth]{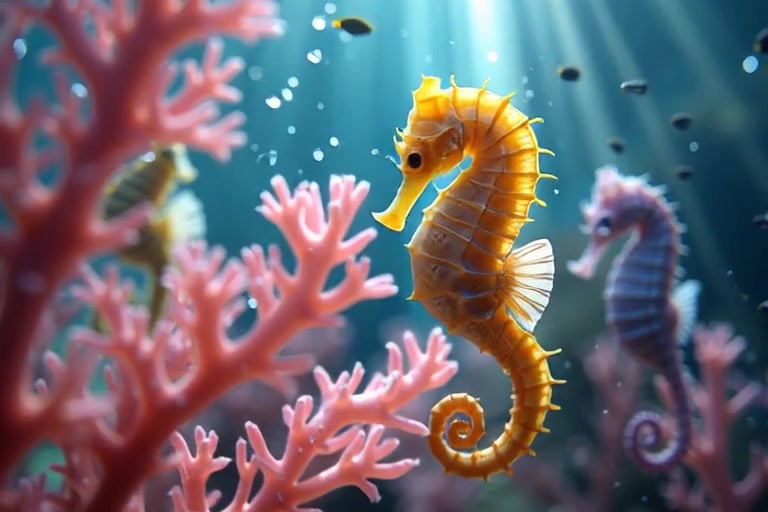} & \includegraphics[width=0.245\textwidth]{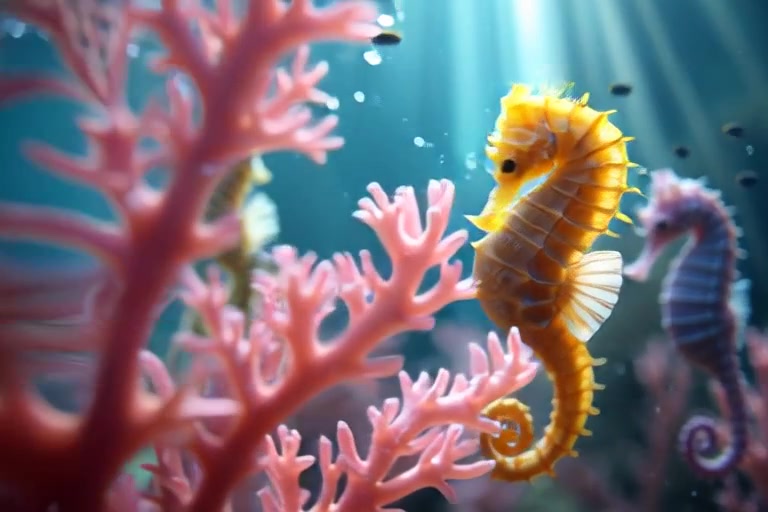} & \includegraphics[width=0.245\textwidth]{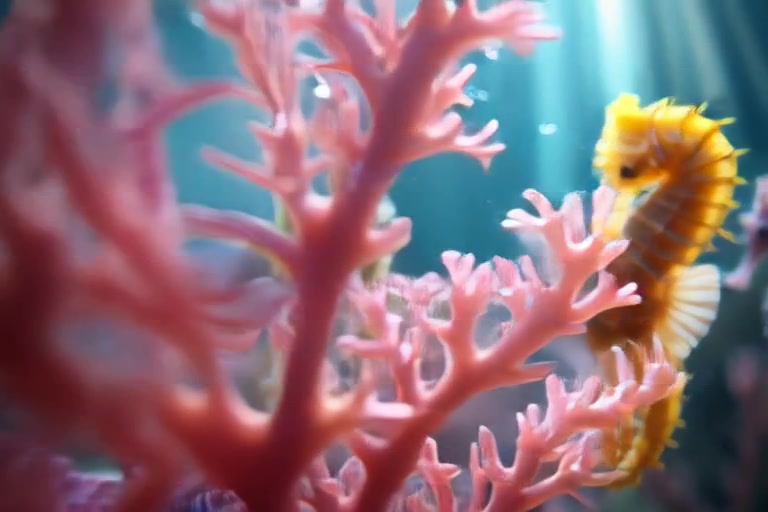} \\

\includegraphics[width=0.245\textwidth]{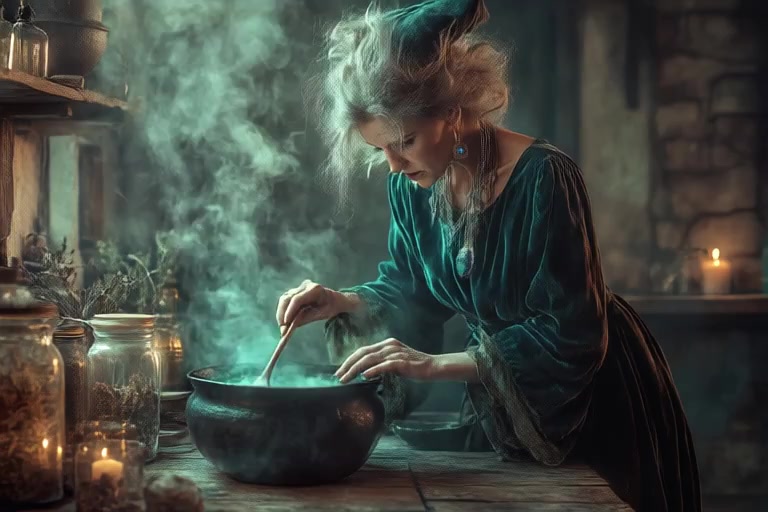} & \includegraphics[width=0.245\textwidth]{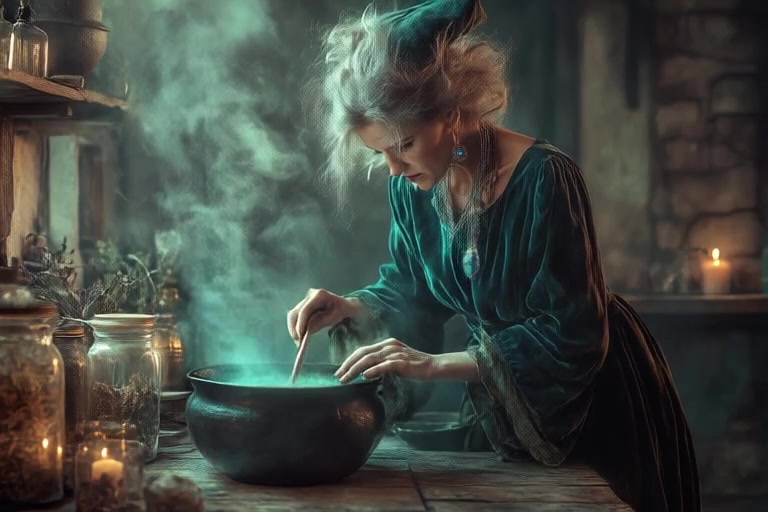} & \includegraphics[width=0.245\textwidth]{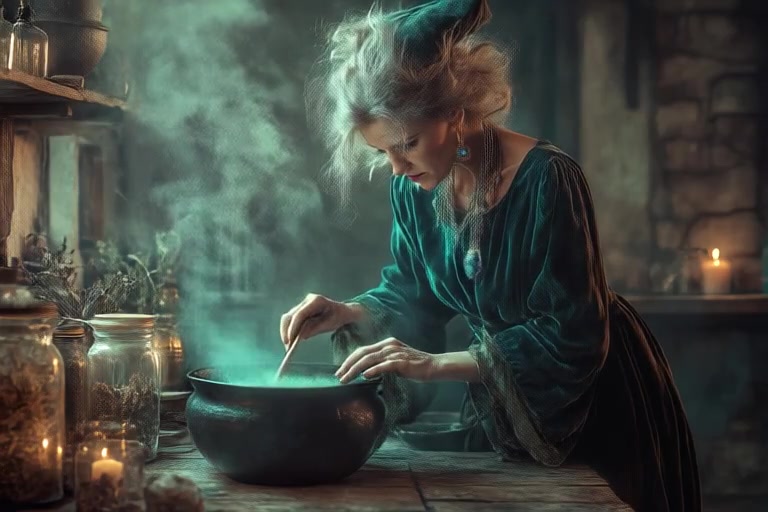} & \includegraphics[width=0.245\textwidth]{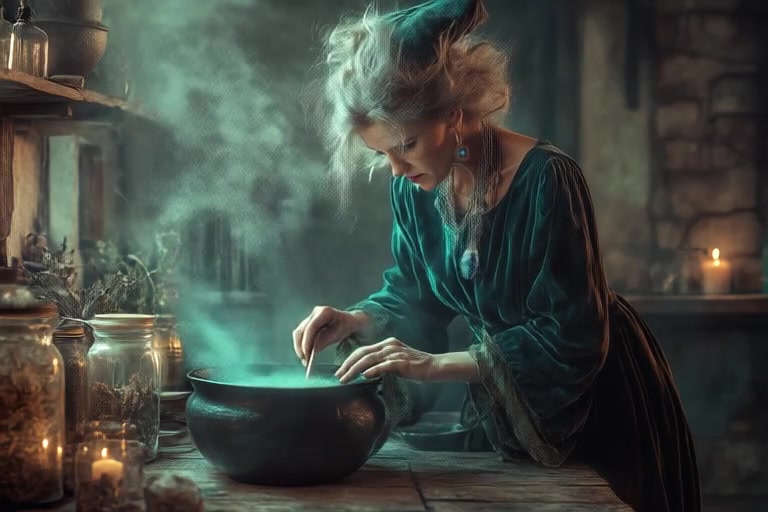} \\

\end{tabular}
\caption{Additional image-to-video results. Video generation is conditioned on the first frame (left column) and on the text prompt (not shown).}
\label{fig:more_results_img_to_video}
\end{figure}

\clearpage
\bibliographystyle{unsrt}
\bibliography{references}

\end{document}